\documentclass[10pt,twocolumn,letterpaper]{article}

\pdfoutput=1

\usepackage{iccv}
\usepackage{times}
\usepackage{epsfig}
\usepackage{graphicx}
\usepackage{amsmath}
\usepackage{amssymb}
\usepackage[export]{adjustbox}

\usepackage{graphicx}
\usepackage{amsmath}
\usepackage{amssymb}
\usepackage{booktabs}
\usepackage{multirow}
\usepackage{soul}
\usepackage{algorithm,algpseudocode}
\usepackage{color}
\usepackage{bbm}

\usepackage{adjustbox}
\usepackage{array}
\newcolumntype{R}[2]{%
    >{\adjustbox{angle=#1,lap=\width-(#2)}\bgroup}%
    l%
    <{\egroup}%
}

\newcommand{\parsection}[1]{\noindent\textbf{#1} }

\newcommand{\modelname}{R3D3}
\newcommand{\im}{\mathbf{I}}
\newcommand{\mask}{\mathbf{M}}
\newcommand{\pose}{\mathbf{G}}
\newcommand{\egopose}{\mathbf{P}}
\newcommand{\extrinsics}{\mathbf{T}}
\newcommand{\rota}{\mathbf{R}}
\newcommand{\trans}{\mathbf{t}}
\newcommand{\intrinsic}{\mathbf{K}}
\newcommand{\depth}{\mathbf{d}}
\newcommand{\corr}{\mathbf{C}}
\newcommand{\flow}{\mathbf{f}}
\newcommand{\conf}{\mathbf{w}}
\newcommand{\residual}{\mathbf{r}}
\newcommand{\covisgraph}{\mathcal{G}}
\newcommand{\nodes}{\mathcal{V}}
\newcommand{\edges}{\mathcal{E}}
\newcommand{\pe}{\operatorname{pe}}

\newcommand{\jacob}{\mathbf{J}}

\newcommand{\schurrS}{\mathbf{S}}
\newcommand{\schurrB}{\mathbf{B}}
\newcommand{\schurrE}{\mathbf{E}}
\newcommand{\schurrC}{\mathbf{C}}
\newcommand{\covar}{\mathbf{\Sigma}}
\newcommand{\identity}{\mathbf{I}}
\newcommand{\damping}{\mathbf{\lambda}}
\newcommand{\logpose}{\mathbf{\xi}}
\newcommand{\coords}{\mathbf{p}}
\newcommand{\point}{\mathbf{X}}

\newcommand{\generator}{\mathfrak{G}}
\newcommand{\expose}{\mathbf{A}}
\newcommand{\hidden}{\mathbf{h}}
\newcommand{\hthresh}{\epsilon}
\newcommand{\matmap}{\Phi}

\newcommand{\real}{\mathbb{R}\xspace}
\newcommand{\se}[1]{\mathfrak{se}(#1)}

\newcommand{\ifpapersuppl}[2]{
\ifcsname papersuppl\endcsname
#1
\else
#2
\fi
}

\definecolor{codeblue}{rgb}{0.25,0.5,0.5}

\algnewcommand\algorithmicinput{\textbf{Input:}}
\algnewcommand\INPUT{\item[\algorithmicinput]}
\algnewcommand\algorithmicinputt{\textbf{Track:}}
\algnewcommand\TRACK{\item[\algorithmicinputt]}
\newcommand{\LineComment}[1]{\hfill \textcolor{codeblue}{\# #1}}

\usepackage[breaklinks=true,bookmarks=false]{hyperref}

\iccvfinalcopy %

\def\iccvPaperID{6411} %

\ificcvfinal\fi

\begin{document}

\title{\modelname: Dense 3D Reconstruction of Dynamic Scenes from Multiple Cameras}

\author{
 Aron Schmied$^{1}$\footnotemark[1] \qquad
 Tobias Fischer$^{1}$\thanks{Equal contribution.} \qquad
 Martin Danelljan$^{1}$ \qquad
 Marc Pollefeys$^{1, 2}$ \qquad 
 Fisher Yu$^{1}$ \vspace{0.05cm}\\
{ $^{1}$ ETH Z{\"u}rich \quad $^{2}$ Microsoft}\vspace{0.05cm} \\
   \url{https://www.vis.xyz/pub/r3d3/} 
}

\maketitle

\ificcvfinal\thispagestyle{empty}\fi

\begin{abstract}

Dense 3D reconstruction and ego-motion estimation are key challenges in autonomous driving and robotics. 
Compared to the complex, multi-modal systems deployed today, multi-camera systems provide a simpler, low-cost alternative.
However, camera-based 3D reconstruction of complex dynamic scenes has proven extremely difficult, as existing solutions often produce incomplete or incoherent results.
We propose \modelname, a multi-camera system for dense 3D reconstruction and ego-motion estimation.
Our approach iterates between geometric estimation that exploits spatial-temporal information from multiple cameras, and monocular depth refinement.
We integrate multi-camera feature correlation and dense bundle adjustment operators that yield robust geometric depth and pose estimates.
To improve reconstruction where geometric depth is unreliable, \eg for moving objects or low-textured regions, we introduce learnable scene priors via a depth refinement network.
We show that this design enables a dense, consistent 3D reconstruction of challenging, dynamic outdoor environments.
Consequently, we achieve state-of-the-art dense depth prediction on the DDAD and NuScenes benchmarks.

\end{abstract}

\vspace{-2mm}
\section{Introduction}

\def \inpidth {0.3\linewidth}
\def \methodwidth {0.45\linewidth}
\def \cloudwidth {0.17\linewidth}

\def \cropcloudl {20px}
\def \cropcloudb {80px}
\def \cropcloudr {50px}
\def \cropcloudt {60px}

\begin{figure}
    \small
    \centering
    \setlength\tabcolsep{0.2 mm}
\begin{tabular}{ccc}

& Single-Camera & \multirow{2}{*}{
\begin{tabular}{l}\adjincludegraphics[width=\cloudwidth,trim={{\cropcloudl} {\cropcloudb} {\cropcloudr} {\cropcloudt}},clip]{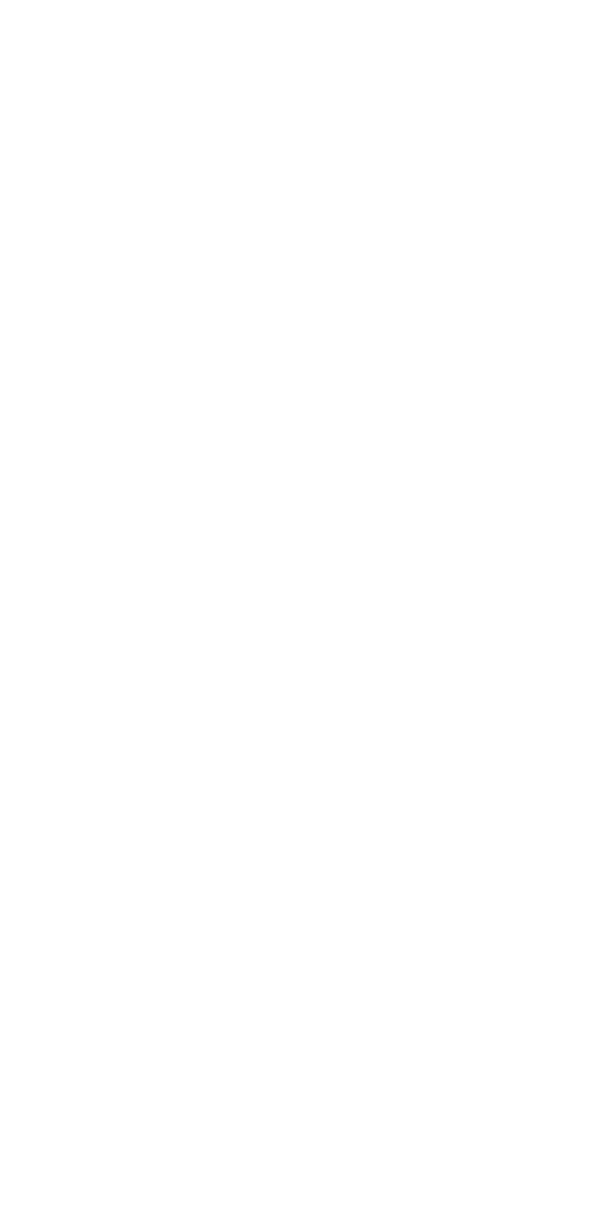}\end{tabular}
} \\
\begin{tabular}{l}\adjincludegraphics[width=\inpidth,trim={10 6 25 6},clip]{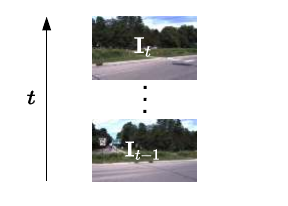}\end{tabular} &
\begin{tabular}{l}\adjincludegraphics[width=\methodwidth,trim={10 6 7 6},clip]{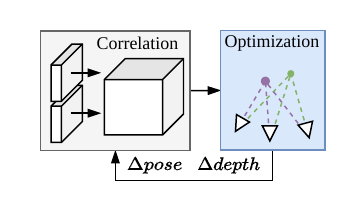}\end{tabular} & \\ \midrule

& Multi-Camera & \multirow{2}{*}{
\begin{tabular}{l}\adjincludegraphics[width=\cloudwidth,trim={{\cropcloudl} {\cropcloudb} {\cropcloudr} {150px}},clip]{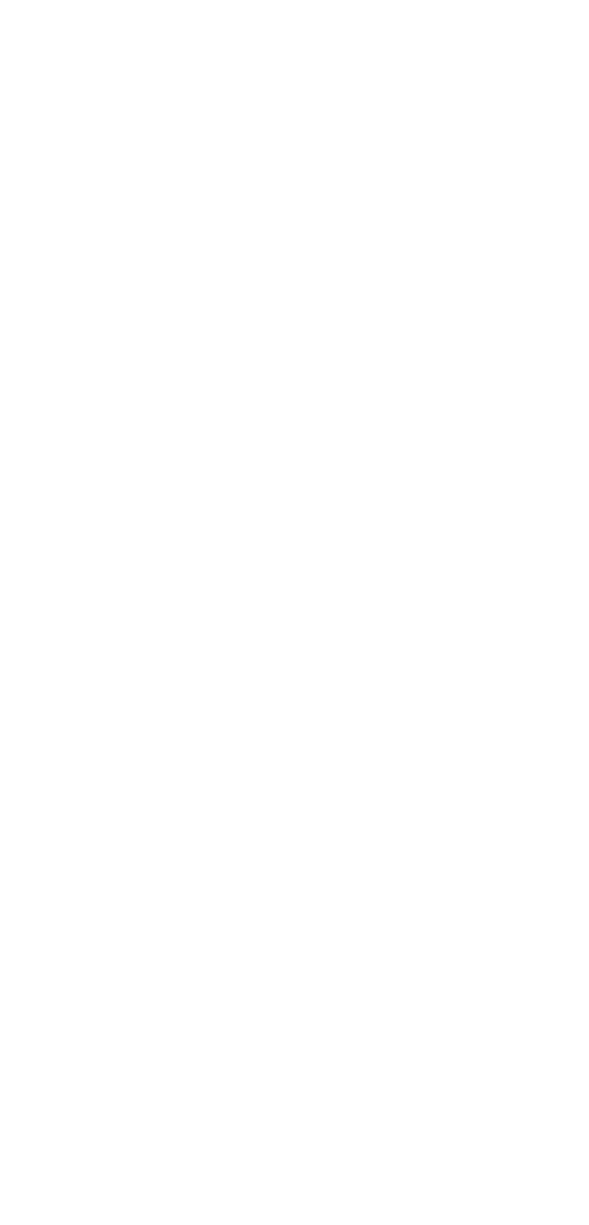}\end{tabular}
} \\
\begin{tabular}{l}\adjincludegraphics[width=\inpidth,trim={15 10 35 10},clip]{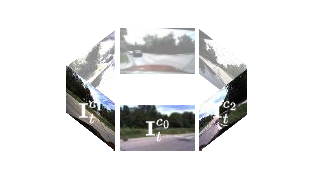}\end{tabular} &
\begin{tabular}{l}\adjincludegraphics[width=\methodwidth,trim={20 15 20 10},clip]{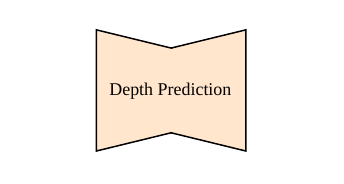}\end{tabular} & \\ \midrule

& \textbf{\modelname} (Ours) & \multirow{2}{*}{
\begin{tabular}{l}\adjincludegraphics[width=\cloudwidth,trim={{\cropcloudl} {20 px} {\cropcloudr} {\cropcloudt}},clip]{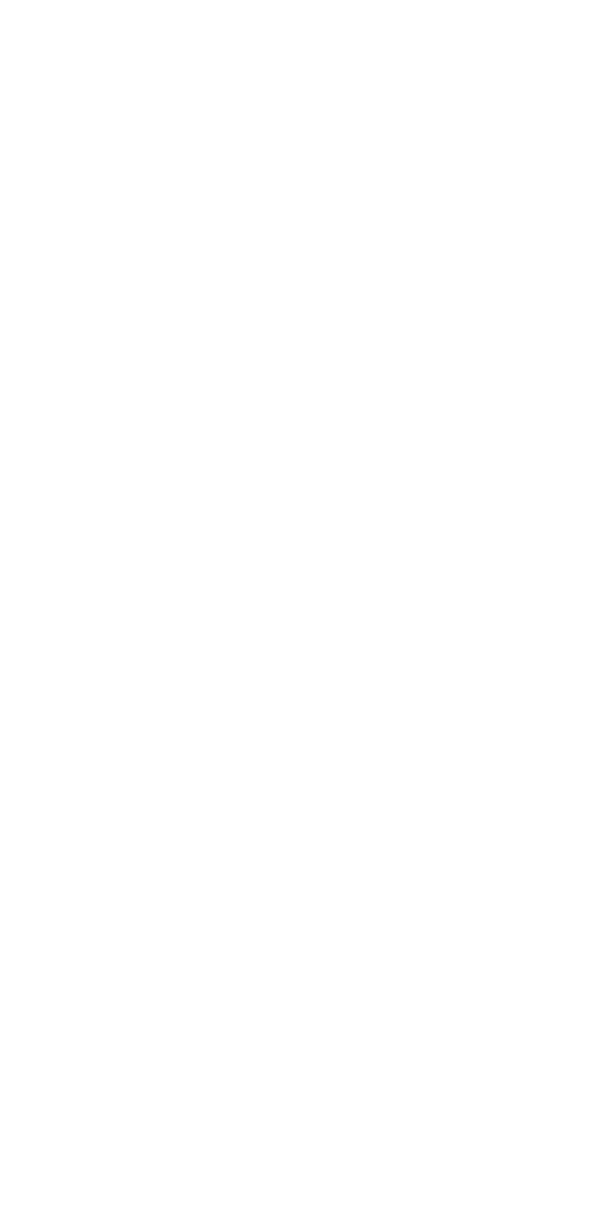}\end{tabular}
}\\
\begin{tabular}{l}\adjincludegraphics[width=\inpidth,trim={15 8 35 12},clip]{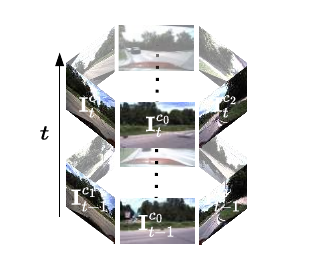}\end{tabular} & %
\begin{tabular}{l}\adjincludegraphics[width=\methodwidth,trim={6},clip]{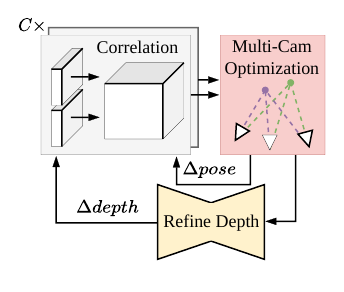}\end{tabular} & \\

\end{tabular}
    
   \caption{Many methods use temporal context but neglect inter-camera information, leading to incomplete results (\textbf{top}). Other works focus on exploiting inter-camera context but neglect temporal information, yielding incoherent predictions (\textbf{middle}). In contrast, our method achieves a consistent, dense 3D reconstruction by iteratively integrating geometric depth estimation from multiple cameras with monocular depth refinement (\textbf{bottom}).}
\label{fig:teaser}\vspace{-2mm}
\end{figure}

Translating sensory inputs into a dense 3D reconstruction of the environment and tracking the position of the observer is a cornerstone of robotics and fundamental to the development of autonomous vehicles (AVs).
Contemporary systems rely on fusing many sensor modalities like camera, LiDAR, RADAR, IMU and more, making hardware and software stacks complex and expensive.
In contrast, multi-camera systems provide a simpler, low-cost alternative already widely available in modern consumer vehicles. 
However, image-based dense 3D reconstruction and ego-motion estimation of large-scale, dynamic scenes is an open research problem as moving objects, uniform and repetitive textures, and optical degradations pose significant algorithmic challenges.

Existing works approaching the aforementioned task can be divided into two lines of research.
Many methods have focused on recovering 3D scene structure via structure-from-motion (SfM).
In particular, simultaneous localization and mapping (SLAM) methods focus on accurate ego-motion estimation and usually only recover sparse 3D structure~\cite{davison2003real, klein2007parallel, mur2015orb, engel2014lsd, engel2017direct}. They typically treat dynamic objects or uniformly textured regions as outliers yielding an incomplete 3D reconstruction result, which makes them less suitable for AVs and robotics.
In addition, only a few works have focused on multi-camera setups~\cite{harmat2012parallel, das2015entropy, liu2018towards, kuo2020redesigning, heng2019project}.
In contrast, multi-view stereo (MVS) methods~\cite{kutulakos1999theory, lhuillier2005quasi, schoenberger2016mvs, kar2017learning, yao2018mvsnet, huang2018deepmvs, gu2020cascade} aim to estimate dense 3D geometry but focus on static scenes and highly overlapping sets of images with known poses.

The second line of research focuses on dense depth prediction from monocular cues, such as perspective object appearance and scene context~\cite{SfmLearner, UnsupLRCons, monodepth2, yin2018geonet, wang2018learning, zhan2018unsupervised, PackNet}. However, due to the injective property of projecting 3D structures onto a 2D plane, reconstructing depth from a single image is an ill-posed problem, which limits the accuracy and generalization of these methods.
Recent methods~\cite{ManyDepth, DepthFormer, DRAFT}, inspired by MVS literature, combine monocular cues with temporal context but are focused on front-facing, single-camera setups.
A handful of recent works extend monocular depth estimation to multi-camera setups~\cite{FSM, SurroundDepth, MCDP}. These methods utilize the spatial context to improve accuracy and realize absolute scale depth learning.
However, those works neglect the temporal domain which provides useful cues for depth estimation.

Motivated by this observation, we introduce \modelname, a system for dense 3D reconstruction and ego-motion estimation from multiple cameras of dynamic outdoor environments. 
Our approach combines monocular cues with geometric depth estimates from both spatial inter-camera context as well as inter- and intra-camera temporal context.
We compute accurate geometric depth and pose estimates via iterative dense correspondence on frames in a co-visibility graph.
For this, we extend the dense bundle adjustment (DBA) operator in~\cite{DroidSLAM} to multi-camera setups, increasing robustness and recovering absolute scene scale.
To determine co-visible frames across cameras, we propose a simple yet effective multi-camera algorithm that balances performance and efficiency.
A depth refinement network takes geometric depth and uncertainty as input and produces a refined depth that improves the reconstruction of, \eg, moving objects and uniformly textured areas. 
We train this network on real-world driving data without requiring any LiDAR ground-truth.
Finally, the refined depth estimates serve as the basis for the next iterations of geometric estimation, thus closing the loop between incremental geometric reconstruction and monocular depth estimation.

We summarize our \textbf{contributions} as follows.
\textbf{1)} we propose \modelname, a system for dense 3D reconstruction and ego-motion estimation in dynamic scenes, \textbf{2)} we estimate geometric depth and poses with a novel multi-camera DBA formulation and a multi-camera co-visibility graph,
\textbf{3)} we integrate prior geometric depth and uncertainty with monocular cues via a depth refinement network.

As a result, we achieve state-of-the-art performance across two widely used multi-camera depth estimation benchmarks, namely DDAD~\cite{PackNet} and NuScenes~\cite{nuscenes}.
Further, we show that our system exhibits superior accuracy and robustness compared to monocular SLAM methods~\cite{DroidSLAM, campos2021orb}.
    \section{Related Work}

\parsection{Multi-view stereo.}
MVS methods aim to recover dense 3D scene structure from a set of images with known poses.
While earlier works focused on classical optimization~\cite{kutulakos1999theory, schoenberger2016mvs, campbell2008using, furukawa2009accurate, galliani2015massively, zheng2014patchmatch, lhuillier2005quasi}, more recent works capitalize on the success of convolutional neural networks (CNNs). 
They use CNNs to estimate features that are matched across multiple depth-hypothesis planes in a 3D cost volume~\cite{kar2017learning, yao2018mvsnet, zhou2018deeptam, huang2018deepmvs, gu2020cascade, yao2019recurrent, yu2020fast}. While early approaches adopt multiple cost volumes across image pairs~\cite{zhou2018deeptam}, recent approaches use a single cost volume across the whole image set~\cite{yao2018mvsnet}.
These works assume a controlled setting with many, highly overlapping images and known poses to create a 3D cost volume. Instead, we aim to achieve robust, dense 3D reconstruction from an arbitrary multi-camera setup on a moving platform with an unknown trajectory.

\parsection{Visual SLAM.}
Visual SLAM approaches focus on jointly mapping the environment and tracking the trajectory of the observer from visual inputs, \ie one or multiple RGB cameras. 
Traditional SLAM systems are often split into different stages, where first images are processed into keypoint matches, which subsequently are used to estimate the 3D scene geometry and camera trajectory~\cite{davison2003real, davison2007monoslam, clemente2007mapping, mur2015orb, mur2017orb, campos2021orb, rosinol2020kimera}. Another line of work focuses on directly optimizing 3D geometry and camera trajectory based on pixel intensities~\cite{engel2014lsd, engel2017direct,yang2018deep, yang2020d3vo}.
A handful of works have focused on multi-camera SLAM systems~\cite{harmat2012parallel, das2015entropy, liu2018towards, kuo2020redesigning, heng2019project}. Recent methods integrate CNN-based depth and pose predictions~\cite{CNNSLAM, yang2018deep, yang2020d3vo} into the SLAM pipeline.
The common challenge these methods face is outliers in the pixel correspondences caused by the presence of low-textured areas, dynamic objects, or optical degradations. Hence, robust estimation techniques are used to filter these outliers, yielding an incomplete, sparse 3D reconstruction result.

In contrast, dense 3D reconstruction and ego-motion estimation has proven to be more challenging. Early works utilize active depth sensors~\cite{newcombe2011dtam, zhou2018deeptam} to alleviate the above-mentioned challenges. 
Recently, several works on dense visual SLAM from RGB input have emerged~\cite{bloesch2018codeslam, tang2018ba, czarnowski2020deepfactors, Teed2020DeepV2D, DroidSLAM}. While these methods are able to produce high-quality depth maps and camera trajectories, they inherit the limitations of their traditional counterparts, \ie their predictions are subject to noise when there are outliers in the pixel correspondences. This can lead to artifacts in the depth maps, inaccurate trajectories, or even a complete failure of the system.
On the contrary, we achieve robust, dense 3D reconstruction and ego-motion estimates by jointly leveraging multi-camera constraints as well as monocular depth cues.

\parsection{Self-supervised depth estimation.}
The pioneering work of Zhou~\etal~\cite{SfmLearner} learns depth estimation as a proxy task while minimizing a view synthesis loss that uses geometric constraints to warp color information from a reference to a target view.
Subsequent research has focused on improving network architectures, loss regularization, and training schemes~\cite{UnsupLRCons, monodepth2, TwoStageTraining, yin2018geonet, wang2018learning, zhan2018unsupervised, PackNet}.
Recent methods draw inspiration from MVS and propose to use 3D cost volumes in order to incorporate temporal information~\cite{ManyDepth, DepthFormer, wimbauer2021monorec, DRAFT}. While these methods achieve promising results, they still focus on single-camera, front-facing scenarios that do not reflect the real-world sensor setups of AVs~\cite{nuscenes, sun2020scalability, PackNet}.
Another recent line of work focuses on exploiting spatial information across overlapping cameras in a multi-camera setup~\cite{FSM, SurroundDepth, MCDP}. These works propose to use static feature matches across cameras at a given time to guide depth learning and estimation, and to recover absolute scale.
Our work aims to exploit both spatial and temporal cues in order to achieve a robust, dense 3D reconstruction from multiple cameras in dynamic outdoor environments.

\section{Method}

\begin{figure*}
\centering
\includegraphics[width=\linewidth]{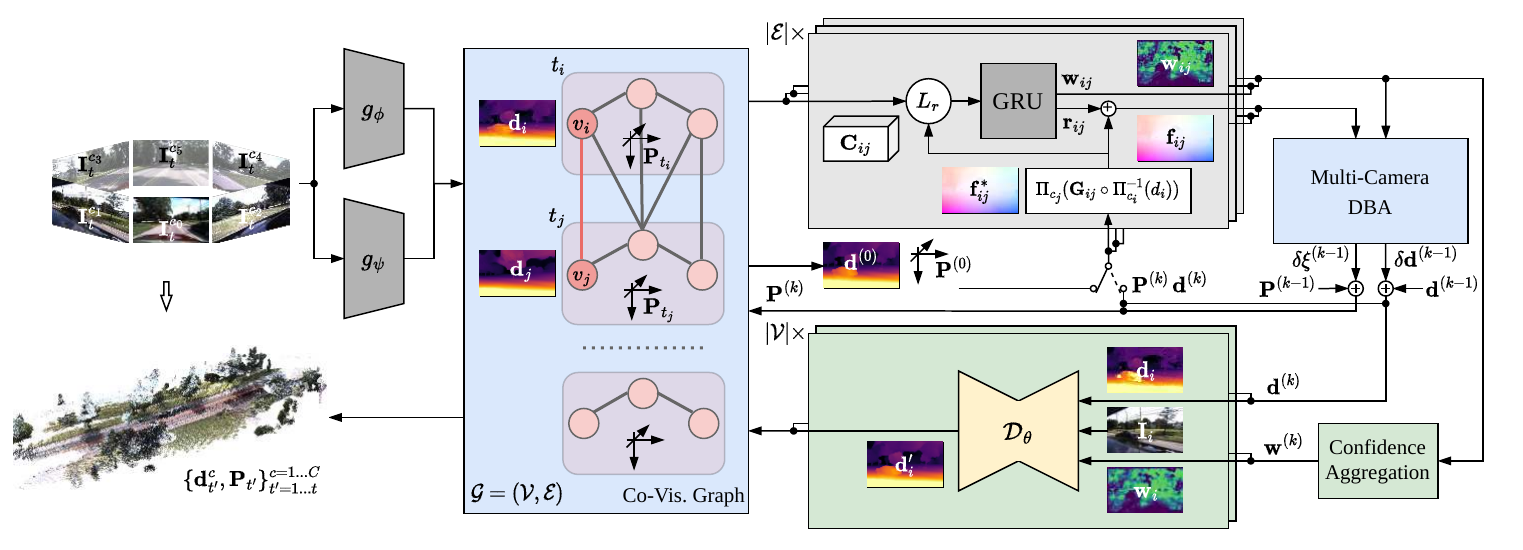}

\vspace{-2mm}
   \caption{\textbf{Method overview.} First, frames $\{\im_t^c\}_{c = 1}^C$ from $C$ cameras at time $t$ are encoded and integrated into the co-visibility graph $\covisgraph = (\nodes, \edges)$ with initial guesses of depth maps $\depth_t^c$ and ego-pose $\egopose_{t}$ (Sec.~\ref{sec:covis_graph}). Second, for each edge $(i, j) \in \edges$, we compute the induced flow $\flow^*_{ij}$ from $\depth_i$ and relative camera pose $\pose_{ij}$ derived from ego-poses $\egopose$ and camera extrinsics $\extrinsics$. Given $\flow^*_{ij}$, we pool feature correlations from $\corr_{ij}$ with operator $L_r$ as input to a GRU that estimates flow update $\residual_{ij}$ and confidence $\conf_{ij}$. We globally align depths $\depth$ and poses $\egopose$ with the new flow estimates $\flow$ via our multi-camera DBA operator in $k$ iterations (Sec.~\ref{sec:DBA}). Finally, for each node $i \in \nodes$, we refine the initial geometric estimate $\depth_i$ given the aggregated confidence $\conf_i$ via $\mathcal{D}_\theta$ (Sec.~\ref{sec:depth_refinement}). We highlight our contributions in color.}
\vspace{-2mm}
\label{fig:architecture}
\end{figure*}

\parsection{Problem formulation.}
At each timestep $t \in T$ we are given a set of images $\{\im_t^c\}_{c = 1}^C$ from $C$ cameras that are mounted on a moving platform with known camera intrinsics $\intrinsic_c$ and extrinsics $\extrinsics_{c}$ with respect to a common reference frame.
We aim to estimate the depth maps $\depth_t^c \in \real_+^{H \times W}$ and ego-pose $\egopose_{t} \in SE(3)$ at the current time step.

\parsection{Overview.}
Our system is composed of three stages. 
First, given a new set of images $\{\im_t^c\}_{c=1}^C$, we extract deep features from each $\im_t^c$. 
We maintain a co-visibility graph $\covisgraph = (\nodes, \edges)$ where a frame $\im_t^c$ represents a node $v \in \nodes$ and co-visible frames are connected with edges.
For each edge $(i, j) \in \edges$ in this graph, we compute the feature correlation $\corr_{ij}$ of the two adjacent frames.
Second, given initial estimates of the depth $\depth_i$ and the relative pose $\pose_{ij} = \left ( \egopose_{t_j} \extrinsics_{c_j} \right )^{-1} \egopose_{t_i} \extrinsics_{c_i}$ between the two frames, we compute the induced flow $\flow^*_{ij}$. We correct the induced flow with a recurrent update operator using the feature correlations $\corr_{ij}$. We then globally align the updated flow estimates at each edge $(i, j)$ with the current estimates of $\pose_{ij}$ and $\depth_i$ across the co-visibility graph $\covisgraph$ with our proposed multi-camera DBA.
Third, we introduce a depth refinement network that refines the geometric depth estimates with monocular depth cues to better handle scenarios that are challenging for geometric estimation methods.
We illustrate our architecture in Fig.~\ref{fig:architecture}.

\subsection{Feature Extraction and Correlation}
\label{sec:covis_graph}
Given a new observation $\{\im_t^c\}_{c =1}^C$, we first extract deep image features, update the co-visibility graph, and compute feature correlations. We describe these steps next.

\parsection{Feature extraction.}
We follow~\cite{RAFT, DroidSLAM} and extract both correlation and context features from each image individually via deep correlation and context encoders $g_{\phi}$ and $g_{\psi}$.

\parsection{Co-visibility graph.}
\begin{figure}
\centering
\includegraphics[width=.75\linewidth]{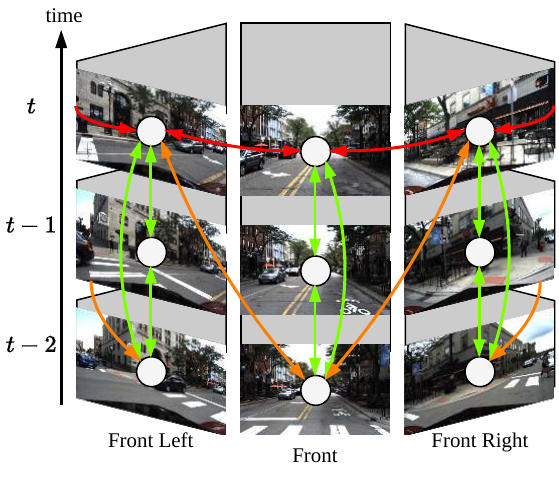}
\vspace{-2mm}
   \caption{\textbf{Co-visibility graph.} We illustrate an example of the connectivity pattern of the co-visibility graph using our multi-camera graph construction algorithm. We use $\Delta t_\text{intra} = 3$, $\Delta t_\text{inter} = 2$, $r_\text{intra} = 2$, $r_\text{inter} = 2$ with six cameras. We depict \textcolor{green}{temporal}, \textcolor{red}{spatial}, and \textcolor{orange}{spatial-temporal} edges.}
\vspace{-2mm}
\label{fig:covis_graph}
\end{figure}
We store correlation and context features in a graph $\covisgraph = (\nodes, \edges)$. Each node corresponds to an image $\im_t^c$ and an edge in the graph indicates that two images are considered a pair in the feature correlation and bundle adjustment steps. 
Contrary to~\cite{DroidSLAM}, we construct $\covisgraph$ with three kinds of edges: temporal, spatial, and spatial-temporal.
Since the number of possible edges is $|\nodes|^2$ and the system runtime scales linearly with $|\edges|$, the degree of sparsity in the co-visibility graph is crucial.
We thus carefully design a simple yet effective co-visibility graph construction algorithm in multi-camera setups (see Fig.~\ref{fig:covis_graph}).

For temporal edges (Fig.~\ref{fig:covis_graph} in \textcolor{green}{green}), we examine the time window $t-\Delta t_\text{intra}$ and connect all frames of the same camera that are less than $r_\text{intra}$ time steps apart.
For spatial and spatial-temporal edges we exploit three priors. 1) we know camera adjacency from the given calibration, 2) we assume motion in the direction of the forward-facing camera and 3) the observer undergoes limited motion within the local time window.
We establish spatial edges (Fig.~\ref{fig:covis_graph} in \textcolor{red}{red}) between adjacent cameras within time step $t$.
We further establish spatial-temporal edges (Fig.~\ref{fig:covis_graph} in \textcolor{orange}{orange}) given the static camera adjacency within $t-\Delta t_\text{inter}$. In particular, we connect camera $c_i$ and $c_j$ from any $t'$ to $t'-r_\text{inter}$, if $c_i$ and $c_j$ connect within $t$ and if $c_j$ is closer to the forward-facing camera than $c_i$ under forward motion assumption.
Finally, we remove an edge $(i, j)$ if node $i$ or $j$ is outside the local time windows $t-\Delta t_\text{inter}$ or $t-\Delta t_\text{intra}$. We remove unconnected nodes from the graph. We provide pseudo-code and additional discussion in the supplemental material.

\parsection{Feature correlation.}
For each edge $(i, j) \in \edges$ we compute the 4D feature correlation volume via the dot-product
\begin{equation}
    \corr_{ij} = \langle g_\phi(\im_i), g_\phi(\im_j) \rangle \in \real^{H \times W \times H \times W}.
\end{equation}
As in~\cite{RAFT, DroidSLAM}, we constrain the correlation search region to a radius $r$ with a lookup operator
 \begin{equation}
 \label{eq:lookup}
     L_r: \real^{H \times W \times H \times W} \times \real^{H \times W \times 2} \rightarrow \real^{H \times W \times (r + 1)^2}  \, ,
\end{equation}
that samples a $H \times W$ grid of coordinates from the correlation volume using bilinear interpolation.

\subsection{Depth and Pose Estimation}
\label{sec:DBA}
Given the correlation volume $\mathbf{C}_{ij}$, we here describe how to estimate the relative pose $\pose_{ij}$ and depth $\depth_i$ for each edge $(i, j) \in \edges$ in the co-visibility graph.

\parsection{Flow correction.}
Given an initial estimate of $\pose_{ij}$ and $\depth_i$, we first compute the induced flow $\flow_{ij}^*$ to sample the correlation volume $\corr_{ij}$ using \eqref{eq:lookup}.
We then feed the sampled correlation features, the context features $g_{\psi}(\im_i)$, and the induced flow $\flow_{ij}^*$ into a convolutional GRU. The GRU predicts a flow residual $\residual_{ij}$ and confidence weights $\mathbf{w}_{ij} \in \mathbb{R}^{H \times W \times 2}$ as in~\cite{DroidSLAM}.
A challenge of correspondence estimation in a multi-camera setup is that the flow magnitude varies greatly among edges in the co-visibility graph due to inter-camera edges with large relative rotation.
To this end, we propose to subtract the rotational part of $\flow_{ij}^*$ in those edges via
\begin{equation}
    \flow'_{ij} = (\mathbf{x}_i + \flow_{ij}^*) - \Pi_{c_j} (\rota_{c_j}^{-1} \rota_{c_i} \circ \Pi_{c_i}^{-1}(\mathbf{1})) \,,
\end{equation}
where  $\mathbf{x}_i$ are pixel coordinates of frame $i$, $\Pi_{c}$ denotes the projection operator under camera $c$ and $\rota$ is the rotation of transformation $\extrinsics = (\rota \, \trans)$. 
This aligns the flow magnitude with temporal, intra-camera edges. Note that, while we input $\flow'_{ij}$ instead of $\flow_{ij}^*$ to the GRU, we still sample the correlation volume $\corr_{ij}$ based on $\flow_{ij}^*$.

\parsection{Pose and depth correction.}
We update our pose and depth estimates given the updated flow $\flow_{ij} = \flow_{ij}^* + \residual_{ij}$ by minimizing the re-projection error, defined as
\begin{equation}
\label{eq:ba_energy}
    E = \sum_{(i, j) \in \edges} \left\| (\mathbf{x}_i + \flow_{ij}) - \Pi_{c_j}(\pose_{ij} \circ \Pi_{c_i}^{-1}(\depth_i)) \right\|_{\Sigma_{ij}}^2 \, ,
\end{equation}
where $\|.\|_{\Sigma_{ij}}$ is the Mahalanobis norm and $\Sigma_{ij} = \operatorname{diag}(\conf_{ij})$. Thus, only matched pixels where the confidence $\conf_{ij}$ is greater than zero contribute to the total cost.
We extend the dense bundle adjustment (DBA) proposed in~\cite{DroidSLAM} to include the known extrinsics of the multi-camera setup. 
In particular, we decompose the relative poses between two frames $\pose_{ij}$ into the unknown, time-varying poses $\egopose$ in the reference frame and the known static relative transformations $\extrinsics$ between the cameras and the reference frame via
\begin{equation}
\label{eq:pose_decomposition}
    \pose_{ij} = \left ( \egopose_{t_j} \extrinsics_{c_j} \right )^{-1} \egopose_{t_i} \extrinsics_{c_i}.
\end{equation}
We linearize the residual of the energy function in Eq.~\ref{eq:ba_energy} with a first-order Taylor expansion. 
We use Eq.~\ref{eq:pose_decomposition} and treat $\extrinsics_{c_j}$ and $\extrinsics_{c_i}$ as constants when calculating the Jacobians, thus computing updates only for ego-pose $\egopose$.
We apply the Gauss-Newton step to compute the updated $\egopose^{(k)} = \exp(\delta \xi^{(k-1)}) \egopose^{(k-1)}$ and depth $\depth^{(k)} = \delta \depth^{(k-1)} + \depth^{(k-1)}$ with $\delta \xi^{(k-1)} \in \se{3}$.  We provide a detailed derivation in the supplementary material.

Our formulation has two advantages over DBA~\cite{DroidSLAM}. First, the optimization is more robust since $C$ frames are connected through a single ego-pose $\egopose$. If one or multiple frames have a high outlier ratio in feature matching, \eg through lens-flare, low textured regions, or dynamic objects, we can still reliably estimate $\egopose$ with at least one undisturbed camera. Second, if there are not only pixels matched across the temporal context but also pixels matched across the static, spatial context, we can recover the absolute scale of the scene.
This is because for static, spatial matches the relative transformation $\pose_{ij} = \extrinsics_{c_j}^{-1} \extrinsics_{c_i}$ is known to scale so that $E$ is minimized if and only if $\depth$ is in absolute scale. 

\parsection{Training.}
We train the networks $g_\phi$ and $g_\psi$ as well as the flow correction GRU on dynamic scenes following the procedure in~\cite{DroidSLAM}. 
As shown in~\cite{DroidSLAM}, the geometric features learned from synthetic data can generalize to real-world scenes. We can therefore leverage existing synthetic driving datasets without relying on real-world ground-truth measurements from sensors like LiDAR or IMU to adjust our method to dynamic, multi-camera scenarios.

\subsection{Depth Refinement}
\label{sec:depth_refinement}
SfM relies on three assumptions: accurate pixel matches, sufficient camera movement, and a static scene. These assumptions do not always hold in the real world due to, \eg, low-textured areas, a static ego vehicle, or many dynamic agents. Still, we would like the system to produce reasonable scene geometry to ensure safe operation.

On the contrary, monocular depth cues are inherently not affected by these issues. However, they usually lack the accurate geometric detail of SfM methods. 
Hence, for each node $i \in \nodes$, we complement the accurate, but sparse geometric depth estimates with monocular cues.

\parsection{Network design.} 
We use a CNN $\mathcal{D}_\theta$ parameterized by $\theta$. We input the depth $\depth_i$, confidence $\conf_i$, and the corresponding image $\im_i$. The network predicts an improved, dense depth $\depth'_i = \mathcal{D}_\theta(\im_i, \depth_i, \conf_i)$.
We obtain the per-frame depth confidence $\conf_i$ for frame $i$ by using the maximum across the per-edge confidence weights $\conf_{ij}$. We compute $\conf_i = \max_j {\frac{1}{2}(\conf_{ij}^x + \conf_{ij}^y)}$ where $x$ and $y$ are the flow directions.
We use $\operatorname{max}(\cdot)$ because we observe that depth triangulation will be accurate if at least one pixel is matched with high confidence. 
We sparsify input depth and confidence weights by setting regions with confidence $\conf_i < \beta$ to zero. We concatenate these with the image $\im_i$.
We further concatenate depth and confidence with features at 1/8\textsuperscript{th} scale. As in~\cite{monodepth2}, the output depth is predicted at four scales.
To accommodate different focal lengths among cameras in the sensor setup, we do focal length scaling of the output~\cite{CNNSLAM}.

\parsection{Training.} 
Contrary to geometric approaches, monocular depth estimators infer depth from semantic cues, which makes it hard for them to generalize across domains~\cite{guo2018learning}.
Hence, instead of relying on synthetic data, we train $\mathcal{D}_\theta$ on raw, real-world video in a self-supervised manner by minimizing a view synthesis loss~\cite{SfmLearner}.
We minimize the photometric error $\pe(\im_t^c, \im_{t' \rightarrow t}^{c' \rightarrow c})$ between a target image $\im_t^c$ and a reference image $\im_{t'}^{c'}$ warped to the target viewpoint via
\begin{equation}
    \im_{t' \rightarrow t}^{c' \rightarrow c} = \Phi(\im_{t'}^{c'}; \Pi_{c'}(\pose_{(t, c) (t', c')} \circ \Pi_{c}^{-1}(\depth_t^c)) ) \, ,
\end{equation}
with $\Phi$ the bi-linear sampling function, $\depth_t^c$ the predicted depth, $c' = c\pm1$ and $t' =t\pm1$. We compute photometric similarity with SSIM~\cite{wang2004image} and $L_1$ distances.
We use $\pose_{(t, c) (t', c')} = (\egopose_{t'} \extrinsics_{c'})^{-1} \egopose_t \extrinsics_c $ generated by the first two stages of our system (see Sec.~\ref{sec:DBA}). 
Further, self-supervised depth estimation is well-studied, and we follow the common practice of applying regularization techniques to filter the photometric error~\cite{monodepth2, Zhong2019, TwoStageTraining}.
First, we mask areas where $\mask^\text{st} = \big[\pe(\im^c_t, \im_{t' \rightarrow t}^{c' \rightarrow c}) < \pe(\im_t^c, \im_{t'}^{c'})\big]$, \ie we filter regions where assuming a stationary scene would induce a lower loss than re-projection.
Second, we compute an induced flow consistency mask from $\flow^*$,
\begin{equation}
    \mask^\text{fc} \!=\! \left[\left\|\flow^*_{(t, c) (t', c')} + \Phi(\flow^*_{(t', c') (t, c)}; \mathbf{x} + \flow^*_{(t, c) (t', c')}) \right\|_2 \!<\! \gamma \right] \!.
\end{equation}
This term warps pixel coordinates $\mathbf{x}$ from target to reference view, looks up their value in $\flow^*_{(t', c') (t, c)}$, and compares them to $\flow^*_{(t, c) (t', c')}$. Therefore, pixels not following the epipolar constraint like dynamic objects are masked.
Third, to handle the self-occlusion of the ego-vehicle, we manually draw a static mask $\mask^\text{oc}$. We can use a single mask for all frames since the cameras are mounted rigidly.
Further, to handle occlusions due to change in perspective, we take the minimum loss $\min_{t', c'} \pe(\im_t^c, \im_{t' \rightarrow t}^{c' \rightarrow c})$ across all reference views.
The overall loss for $\mathcal{D}_\theta$ is thus
\begin{equation}
\label{eq:masked_loss}
    \mathcal{L} = \min_{t', c'} \left(\mask^\text{st} \mask^\text{fc} \mask^\text{oc} \cdot \pe(\im_t^c, \im_{t' \rightarrow t}^{c' \rightarrow c})\right) + \lambda\mathcal{L}_\text{smooth}  \, ,
\end{equation}
with $\mathcal{L}_\text{smooth}$ a spatial smoothness regularization term~\cite{UnsupLRCons}.

\subsection{Inference Procedure}
\label{sec:inference}
Given an input stream of multi-camera image sets, we perform incremental 3D reconstruction in three phases.

\parsection{Warmup phase.} First, we add frames to the co-visibility graph if a single GRU step of a reference camera yields a large enough mean optical flow until we reach $n_\text{warmup}$ frames. At this stage, we infer depth with our refinement network $\depth_t^c = \mathcal{D}_\theta(\im_t^c, \mathbf{0}, \mathbf{0})$. We write the depth into the co-visibility graph so that it can be used during initialization.

\parsection{Initialization phase.} We initialize the co-visibility graph with the available frames and run $n_\text{itr-wm}$ GRU and bundle adjustment iterations. 
For the first $n_\text{itr-wm} / 2$ iterations, we fix depths in the bundle adjustment and only optimize poses, which helps with inferring the absolute scale of the scene.

\parsection{Active phase.} We now add each incoming frame to the co-visibility graph. 
We initialize the new frames with $\depth_{t}^c = \operatorname{mean}(\depth_{t-4 : t-1}^c)$ and $\egopose_{t}=\egopose_{t-1}$.
Then, we perform $n_\text{iter1}$ GRU and bundle adjustment updates, and apply the refinement network to all updated frames.
If the mean optical flow of the reference camera between $t$ and $t-1$ is low, we remove all frames at $t-1$ from the co-visibility graph. This helps to handle situations with little to no ego-motion.
In case no frame is removed, we do another $n_\text{iter2}$ iterations. Finally, we write the updated depths and poses into the co-visibility graph.

\section{Experiments}

\begin{table*}[t]
\centering
\caption{\textbf{Method ablation.} We ablate our method components on the DDAD~\cite{PackNet} dataset. We examine the influence of each component given the geometric estimation as in~\cite{DroidSLAM} with naive DBA as the baseline. Further, we show the influence of VKITTI~\cite{VKITTI2} fine-tuning on the geometric estimation. We enumerate each variant for better reference. * median scaled depth, ** scaled trajectory.}
\resizebox{0.9\linewidth}{!}{%
\begin{tabular}{c|cccc|cccccc}
\toprule
No. & Geom. Est. & VKITTI & Multi-Cam DBA & Refinement & Abs Rel $\downarrow$ & Sq Rel $\downarrow$ & RMSE $\downarrow$ & $\delta_{1.25}$ $\uparrow$ & ATE [m] $\downarrow$ & ATE** [m] $\downarrow$ \\ \midrule
1a\textbf{*}& \checkmark & \checkmark &   &   & 0.438 & 26.97 & 19.109 & 0.636  & - & 2.134 \\
1b & \checkmark  &   & \checkmark &   & 0.442 & 36.39 & 18.664 & 0.736 & 1.672 & 0.435 \\ %
1c & \checkmark & \checkmark & \checkmark &   & 0.320 & 15.45 & 16.303 & 0.727 & 2.356 & 0.922 \\ \midrule %
2a &  &   &   & \checkmark & 0.211 & 3.806 & 12.668 & 0.715 & -     & -   \\ \midrule %
3a & \checkmark & \checkmark & \checkmark & \checkmark & \textbf{0.162} & \textbf{3.019} & \textbf{11.408} & \textbf{0.811} & \textbf{1.235} &  \textbf{0.433} \\ \bottomrule %
\end{tabular}}
\label{tab:ablation_main}\vspace{-2mm}
\end{table*}
\begin{table}[t]
\centering%
\caption{\textbf{Co-visibility graph ablation.} We compare our co-visibility graph construction algorithm to the original algorithm in~\cite{DroidSLAM} on the DDAD~\cite{PackNet} dataset. We observe that our algorithm improves the overall runtime by an order of magnitude while maintaining the same level of performance.}%
\resizebox{0.95\linewidth}{!}{%
\begin{tabular}{@{}l|@{~~}c@{~~}c@{~~}c@{~~}c@{~~}c@{~~}c@{}}
\toprule
Graph & Abs Rel $\downarrow$ & Sq Rel $\downarrow$ & RMSE $\downarrow$ & $\delta_{1.25}$ $\uparrow$ & ATE [m] $\downarrow$ & $t_{inf}$ [s] $\downarrow$\\ \midrule
Original & \textbf{0.162} & \textbf{2.968} & \textbf{11.156} & \textbf{0.816} & \textbf{0.982} & 3.23 \\ 
\textbf{Ours} & \textbf{0.162} & 3.019 & 11.408 & 0.811 & 1.235 & \textcolor{red}{\textbf{0.35}} \\ \bottomrule
\end{tabular}}%
\label{tab:ablation_covis}\vspace{-2mm}
\end{table}

\begin{table}[t]
\centering
\caption{\textbf{Refinement network ablation.} We show on the DDAD~\cite{PackNet} dataset that both adding confidence weights $\conf_i$ and sparsifying the depth input via $\conf_i < \beta$ to filter outliers in the input of depth refinement network $\mathcal{D}_\theta$ is vital to depth accuracy.}
\resizebox{0.85\linewidth}{!}{%
\begin{tabular}{cc|cccc}
\toprule
$\conf_i$ & $\conf_i < \beta$ & Abs Rel $\downarrow$ & Sq Rel $\downarrow$ & RMSE $\downarrow$ & $\delta_{1.25}$ $\uparrow$ \\ \midrule
      &    & 0.181 &	4.078 &	12.108 & 0.789 \\ 
\checkmark &    & 0.173 & 3.497 & 11.924 & 0.792 \\
\checkmark & \checkmark   & \textbf{0.162} & \textbf{3.019} & \textbf{11.408} & \textbf{0.811} \\ \bottomrule
\end{tabular}}
\label{tab:ablation_depth}\vspace{-2mm}
\end{table}

\subsection{Experimental Setup}
We perform extensive experiments on large-scale driving datasets that contain recordings from vehicles equipped with multiple cameras, LiDARs, and other sensors. 
Note that we only use camera data for training and inference, and use LiDAR data only as ground-truth for evaluation. We evaluate our method in inference mode (cf. Sec.~\ref{sec:inference}), \ie we obtain predictions in a fully online (causal) manner to mimic real-world deployment.

\parsection{DDAD.} This dataset consists of 150 training and 50 validation scenes from urban areas. Each sequence has a length of 50 or 100 time steps at a frame rate of 10Hz. The sensor setup includes six cameras in a surround-view setup with up to 20\% overlap.
We follow~\cite{PackNet} and evaluate depth up to 200m averaged across all cameras. We use self-occlusion masks from~\cite{SurroundDepth} to filter the ego-vehicle in the images. Images have a resolution of $1216 \times 1936$ and are downsampled to $384 \times 640$ in our experiments. 

\parsection{NuScenes.} This dataset contains 700 training, 150 validation, and 150 testing sequences of urban scenes with challenging conditions such as nighttime and rain. Each scene is composed of 40 keyframes at 2Hz synchronized across sensors. The six cameras have a sampling rate of 12Hz and are arranged in a surround-view setup with up to 10\% overlap. 
We follow~\cite{FSM} and evaluate depth averaged across all cameras on the validation set. We use a single self-occlusion mask for all scenes. While the raw images have a resolution of $900 \times 1600$, we use $768 \times 448$.

\def \imwidth {0.2 \linewidth}

\begin{figure}
\footnotesize
\centering
\setlength\tabcolsep{0.5 pt}
\begin{tabular}{ccc}
Input & Geom. Depth & Refined \\
\begin{tabular}{l}\includegraphics[height=\imwidth]{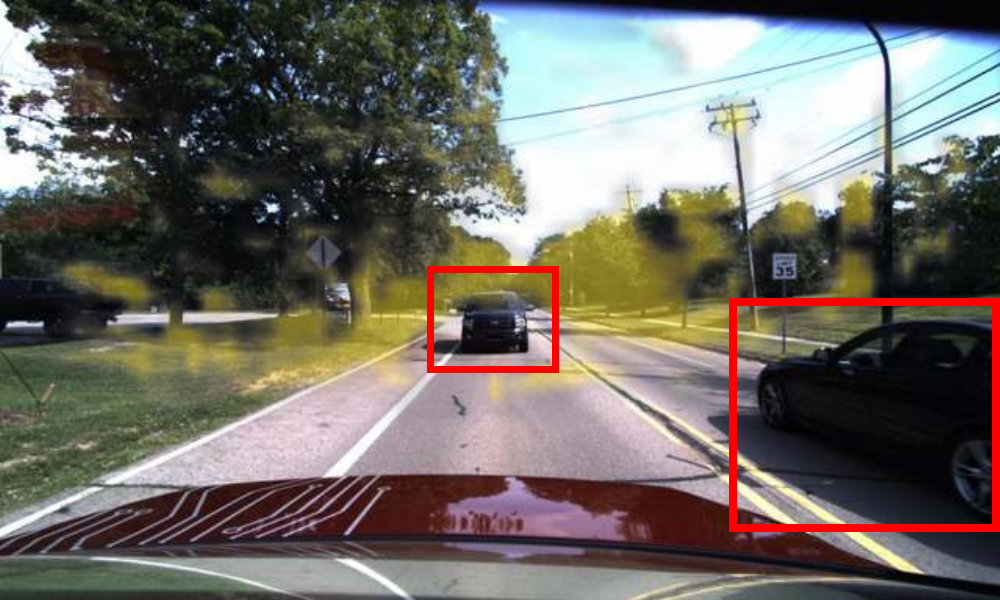}\end{tabular} &
\begin{tabular}{l}\includegraphics[height=\imwidth]{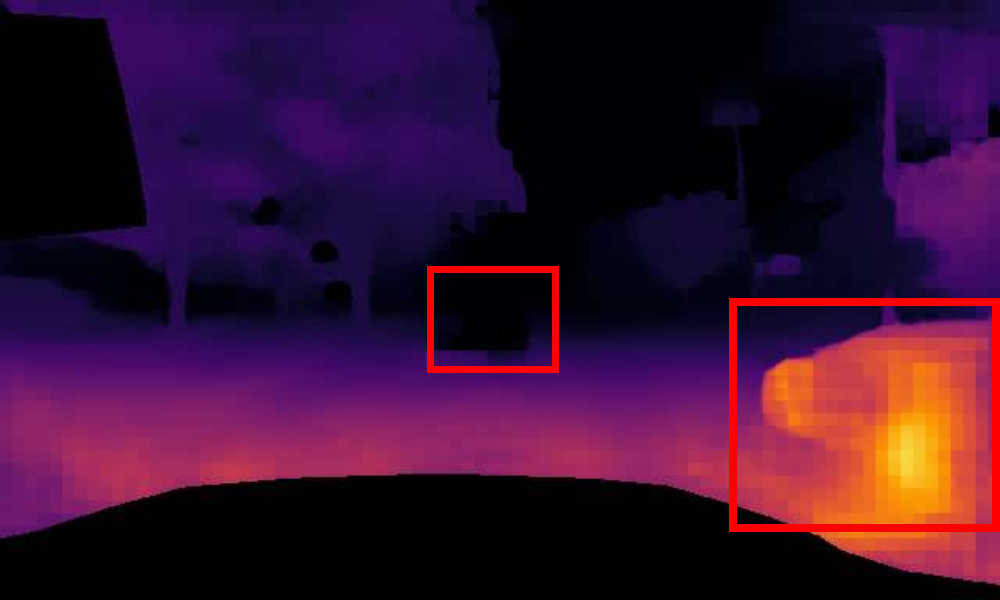}\end{tabular} &
\begin{tabular}{l}\includegraphics[height=\imwidth]{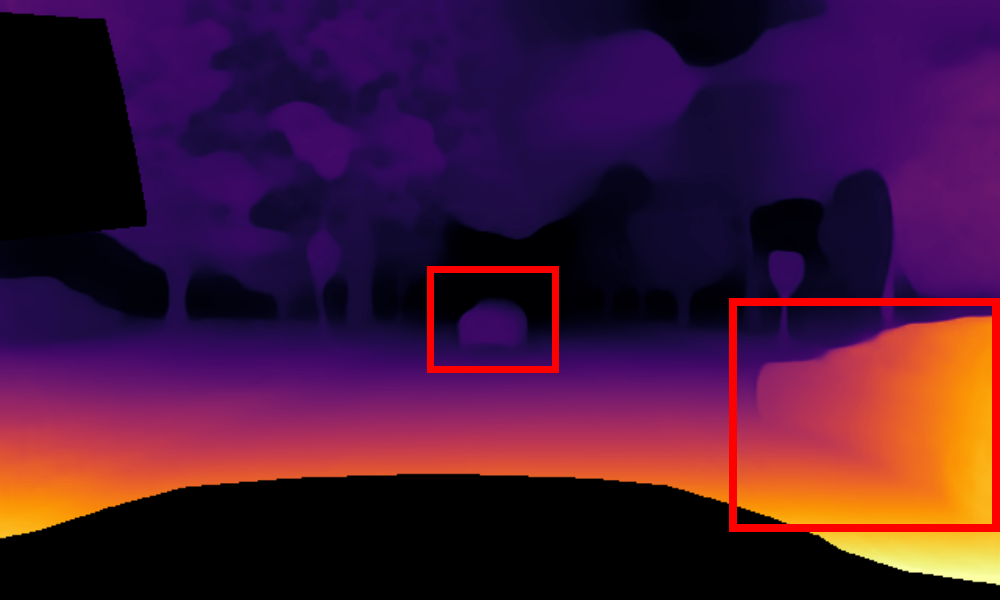}\end{tabular} \\

\begin{tabular}{l}\includegraphics[height=\imwidth]{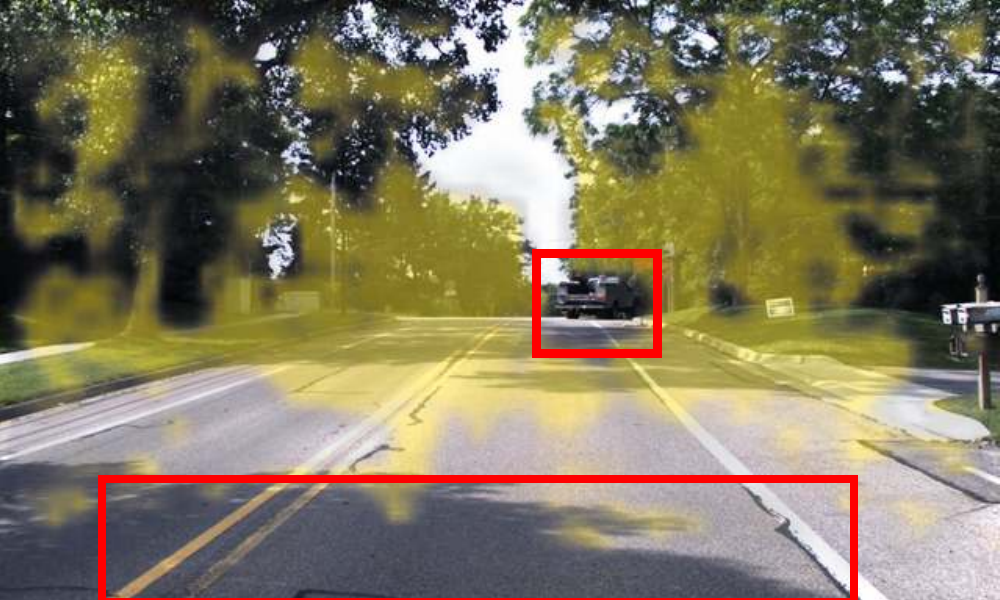}\end{tabular} &
\begin{tabular}{l}\includegraphics[height=\imwidth]{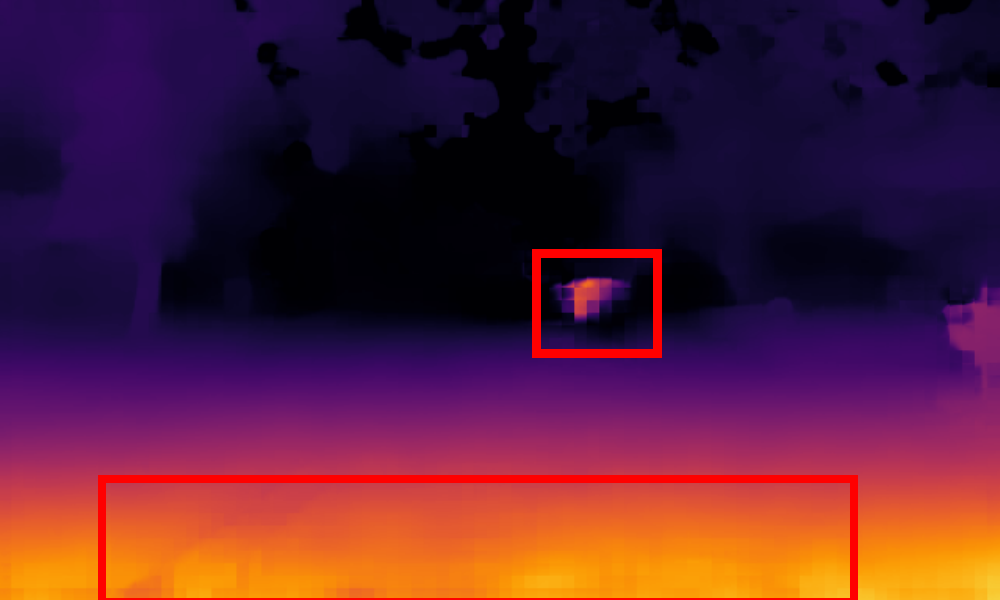}\end{tabular} &
\begin{tabular}{l}\includegraphics[height=\imwidth]{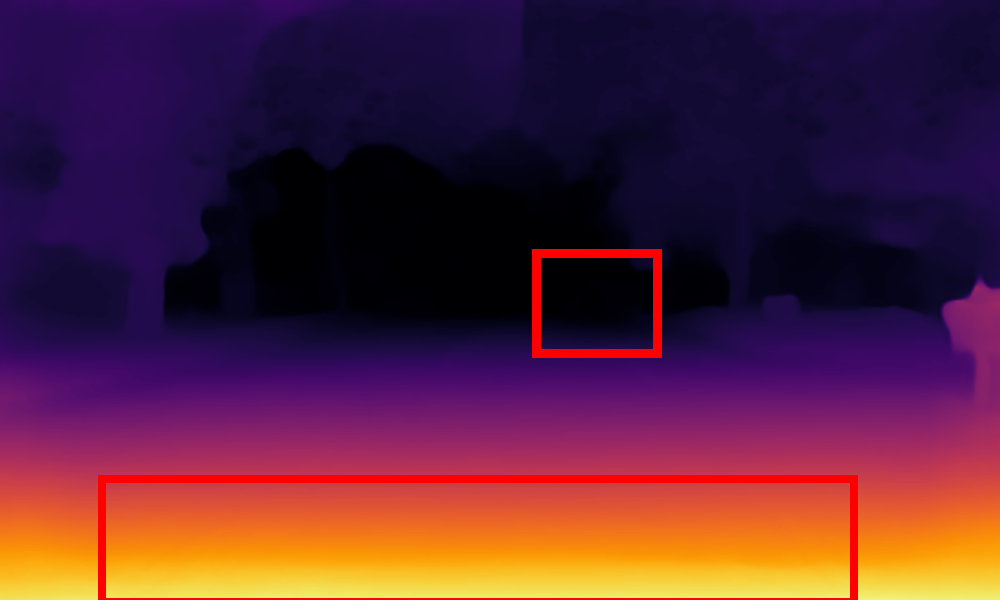}\end{tabular} \\

\begin{tabular}{l}\includegraphics[height=\imwidth]{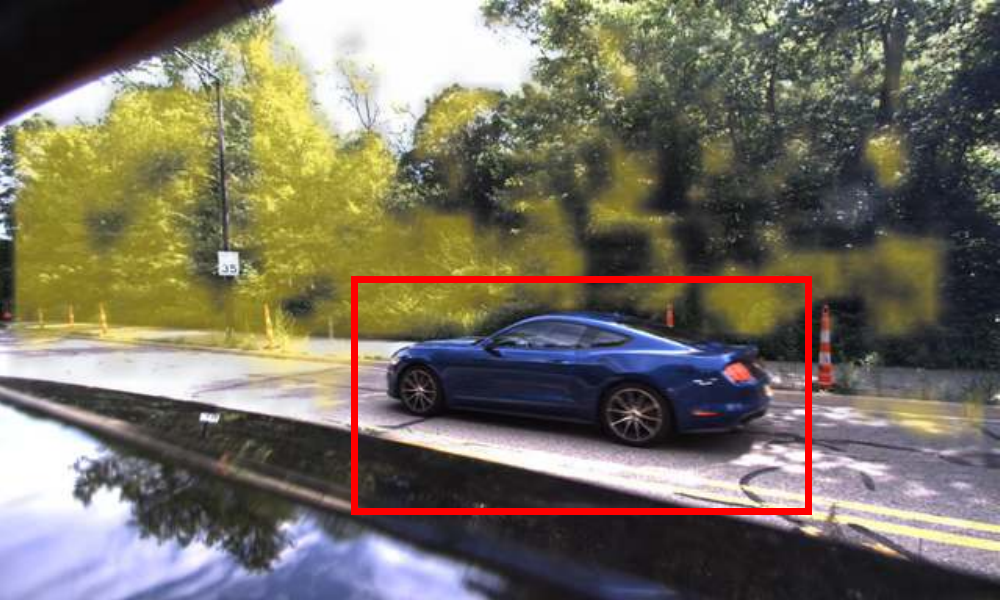}\end{tabular} &
\begin{tabular}{l}\includegraphics[height=\imwidth]{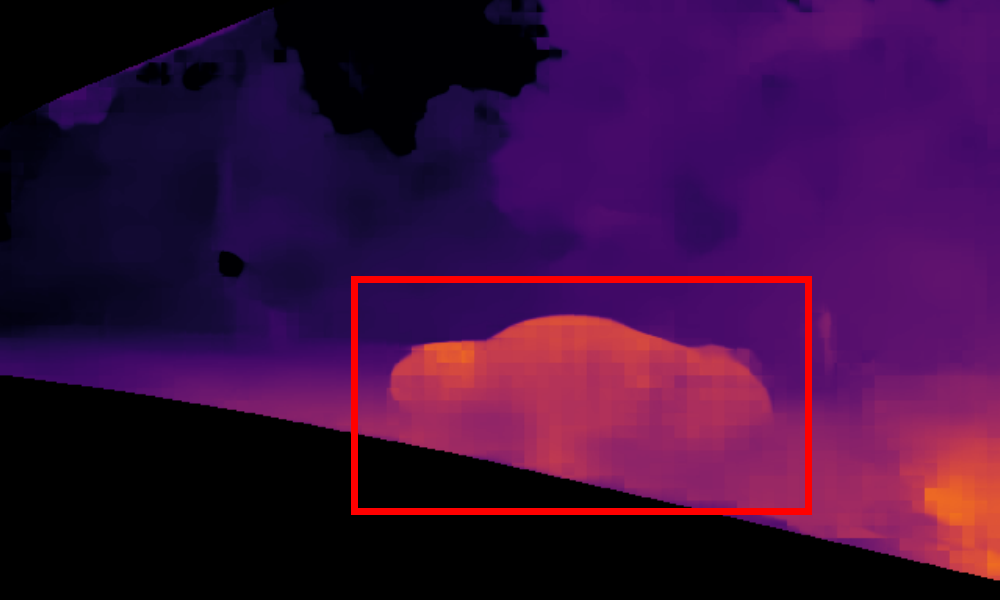}\end{tabular} &
\begin{tabular}{l}\includegraphics[height=\imwidth]{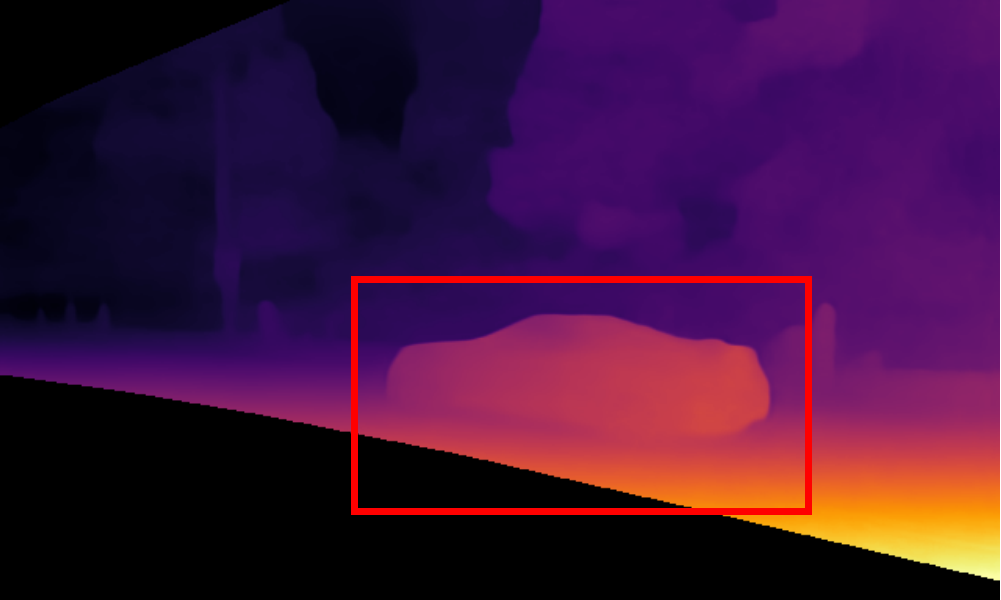}\end{tabular} \\

\end{tabular}

\caption{\textbf{Depth refinement.} We plot images overlayed with confidence $\conf_i$, the geometric depth estimate $\depth$, and the refined depth $\depth'$ from $\mathcal{D}_\theta$. We observe a smoother ground plane, corrected dynamic objects, and filtered outliers.}
\label{fig:refinement}\vspace{-2mm}
\end{figure}

\parsection{Implementation details.}
We implement our system in PyTorch~\cite{paszke2019pytorch}. Our multi-camera DBA is implemented in CUDA.
We fine-tune $g_\phi$, $g_\psi$ and the flow correction GRU on VIKITTI2~\cite{VKITTI2} for 10 epochs with a batch size of 1, sequence length of 7, and learning rate $10^{-4}$ with the pre-trained model in~\cite{DroidSLAM}. 
For $\mathcal{D}_\theta$, we use a U-Net~\cite{ronneberger2015unet} with ResNet18~\cite{ResNet} encoder and initialize it with ImageNet~\cite{ImageNet} pre-trained weights. We use $\beta = 0.5$, $\gamma = 3$ and $\lambda = 10^{-3}$.
We train for 20 epochs with batch size of 6 and the Adam optimizer with learning rate $10^{-4}$, $\beta_1=0.9$, and $\beta_2=0.999$. We reduce the learning rate by a factor of 10 after 15 epochs. 
For inference, the runtime is measured for a set of six images on a single RTX3090 GPU. Each sequence is initialized with 3 warmup frames filtered at a mean flow threshold of 1.75. We build the co-visibility graph with $\Delta t_\text{intra} = 3$, $r_\text{intra} = 2$, $\Delta t_\text{inter} = 2$ and $r_\text{inter} = 2$.

\subsection{Ablation Studies}
\label{sec:ablations}
\parsection{Method ablation.} We ablate our method components on the DDAD dataset in Tab.~\ref{tab:ablation_main}.
The geometric estimation baseline (1a) consists of~\cite{DroidSLAM} with the naive DBA formulation applied to all cameras and the co-visibility graph as described in Sec.~\ref{sec:covis_graph}. It performs poorly in depth and pose estimation, with an Abs.~Rel.\ score of 0.438 and an absolute trajectory error (ATE)~\cite{sturm2012benchmark} of 2.134m, even when adjusting for scale.
Next, we add our multi-camera DBA (1b). We observe that, even without VKITTI fine-tuning, the pose estimation accuracy improves dramatically over the naive DBA baseline. 
The relative scale ATE drops from 2.134m to only 0.435m. The absolute scale ATE is only about 1.2m higher, indicating the scene scale is recovered accurately.

\begin{table}[t]
\centering
\caption{\textbf{Comparison to state-of-the-art on DDAD}. We compare favorably to existing methods on both scale-aware and scale-invariant depth prediction. * median scaled depth. }
\resizebox{0.85\linewidth}{!}{%
\begin{tabular}{@{}l|cccc@{}}
\toprule
\textbf{Method} & Abs Rel $\downarrow$ & Sq Rel  $\downarrow$ & RMSE  $\downarrow$  & $\delta_{1.25}$  $\uparrow$   \\ \midrule
FSM \cite{FSM}                     & \underline{0.201}   & -  & - & - \\
SurroundDepth \cite{SurroundDepth}  & 0.208   & 3.371  & \underline{12.977} & 0.693 \\
\textbf{Ours}   & \textbf{0.162}   & \textbf{3.019}  & \textbf{11.408} & \textbf{0.811} \\ \midrule
FSM* \cite{FSM}                     & 0.202   & -  & - & - \\
SurroundDepth* \cite{SurroundDepth} & 0.200   & 3.392  & 12.270 & 0.740 \\
MCDP* \cite{MCDP}                   & \underline{0.193}   & \underline{3.111}  & \underline{12.264} & \textbf{0.811} \\
\textbf{Ours}*  & \textbf{0.169} & \textbf{3.041} & \textbf{11.372} & \underline{0.809} \\ \bottomrule
\end{tabular}}
\label{tab:comparison_ddad_main}\vspace{-2mm}
\end{table}

\begin{table}[t]
\centering
\caption{\textbf{Per-camera evaluation on DDAD.} We show a per-camera breakdown of previous works, our $\mathcal{D}_\theta$ only (cf. 2a in Tab.~\ref{tab:ablation_main}), and our full method. Our full method performs the best with a particularly significant improvement on the side-view cameras while $\mathcal{D}_\theta$ performs similarly to previous works.}
\resizebox{1.0\linewidth}{!}{%
\begin{tabular}{@{}l|cccccc@{}}
\toprule
              & \multicolumn{6}{c}{Abs Rel  $\downarrow$}                          \\ \cmidrule(l){2-7} 
\textbf{Method} & \textit{Front} & \textit{F.Left} & \textit{F.Right} & \textit{B.Left} & \textit{B.Right} & \textit{Back} \\ \midrule
FSM \cite{FSM}  & \underline{0.130} & \underline{0.201} &\underline{0.224} & 0.229 & 0.240 & \underline{0.186} \\
SurroundDepth \cite{SurroundDepth}  & 0.152 & 0.207 & 0.230 & \underline{0.220} & 0.239 & 0.200 \\ \midrule
$\mathcal{D}_\theta$ only & 0.154 &	0.213 & 0.237 & 0.231 & \underline{0.237} & 0.194  \\
\textbf{Ours} & \textbf{0.128} & \textbf{0.160} & \textbf{0.168} & \textbf{0.172} & \textbf{0.174} & \textbf{0.169} \\ \bottomrule
\end{tabular}}
\label{tab:comparision_ddad_per_cam}\vspace{-2mm}
\end{table}

\begin{table}[t]
\centering
\caption{\textbf{Comparison to multi-frame methods on DDAD.} We compare to methods that exploit temporal context from a single camera~\cite{ManyDepth, DepthFormer}. We evaluate only the front camera since these methods were trained only on this camera.}
\resizebox{0.85\linewidth}{!}{%
\begin{tabular}{@{}l|cccc@{}}
\toprule
& \multicolumn{4}{c}{Front Camera} \\ \cmidrule(l){2-5} 
\textbf{Method} & Abs Rel $\downarrow$ & Sq Rel  $\downarrow$ & RMSE  $\downarrow$  & $\delta_{1.25}$  $\uparrow$   \\ \midrule
ManyDepth \cite{ManyDepth}     & 0.146   & 3.258  & 14.098 & 0.822 \\
DepthFormer \cite{DepthFormer} & \underline{0.135}   & \textbf{2.953}  & \textbf{12.477} & \underline{0.836} \\
\textbf{Ours}   & \textbf{0.128}   & \underline{3.168}  & \underline{13.214} & \textbf{0.868} \\ \bottomrule
\end{tabular}}
\label{tab:comparison_ddad_multiframe}\vspace{-2mm}
\end{table}

\begin{table}[t]
\centering
\caption{\textbf{Pose evaluation on DDAD.} We compare our method to state-of-the-art monocular SLAM methods for which we report the average ATE across the cameras that successfully tracked the camera path. ** scaled trajectory. }
\resizebox{0.85\linewidth}{!}{%
\begin{tabular}{@{}l|ccc@{}}
\toprule
        & DROID-SLAM~\cite{DroidSLAM} & ORB-SLAMv3~\cite{campos2021orb} & \textbf{Ours} \\ \midrule
ATE** {[}m{]} $\downarrow$  & 7.500    & 6.179   & \textbf{0.433}   \\ 
\bottomrule
\end{tabular}
}
\label{tab:pose_evaluation}\vspace{-2mm}
\end{table}

Using VKITTI fine-tuning (1c), we achieve significant gains in depth accuracy, where Abs.\ Rel.\ drops to 0.32 and $\delta_{1.25}$ increases to 72.7\%. However, note that VKITTI fine-tuning does not affect the $\delta_{1.25}$ score, \ie it helps mostly with depth outliers. We attribute this to the absence of dynamic objects in the training data of~\cite{DroidSLAM} so that fine-tuning helps to adjust the confidence prediction to handle such outliers (cf. Sec.~\ref{sec:DBA}).
We further test the depth refinement network (Sec.~\ref{sec:depth_refinement}) without geometric estimation prior (2a). The depth obtained from $\mathcal{D}_\theta$ is less prone to outliers, as evidenced by its low Abs.\ Rel., Sq.\ Rel.\ and RMSE scores. However, note that its $\delta_{1.25}$ is only 71.5\%, \ie its depth accuracy compared to the geometric depth estimates is actually lower when disregarding outliers.
Finally, we test our full system (3a).
Compared to geometric estimation (1d), we observe a significant improvement in outlier sensitive metrics like RMSE, but also in outlier robust metrics like $\delta_{1.25}$. Additionally, the pose estimation accuracy is increased and the difference between scaled and non-scaled ATE is smaller.
Compared to depth obtained from $\mathcal{D}_\theta$ (2a), the increase in outlier sensitive metrics is smaller but still significant, \eg 1.26m in RMSE. Further, we observe a sizable gain in $\delta_{1.25}$ of 9.6 percentage points.

\begin{figure}
\centering
\includegraphics[width=0.9\linewidth]{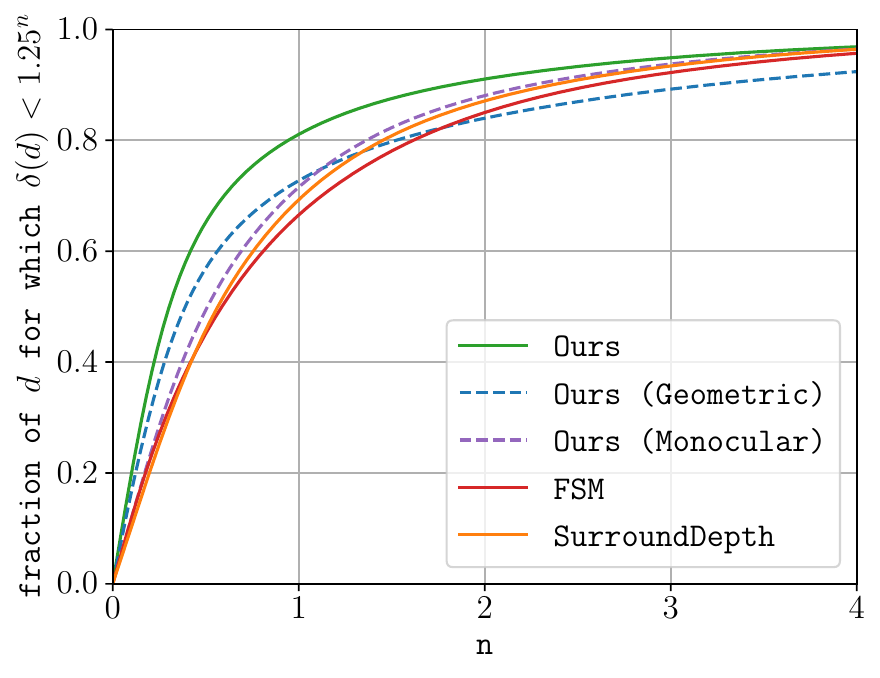}\vspace{-2mm}
\caption{\textbf{Depth in $\delta_{1.25^n}$ range for different thresholds on DDAD.} We depict the ratio of depth estimates $d$ for which $\delta(d) < 1.25^n$ where $\delta(d) = \max{( \frac{d}{d^*}, \frac{d^*}{d} )}$ and $d^*$  is ground-truth depth. Note that values at $n=1, 2, 3$ represent $\delta_{1.25}, \delta_{1.25^2}, \delta_{1.25^3}$.}\vspace{-2mm}
\label{fig:acc_a1_plot}
\end{figure}

Further, in Fig.~\ref{fig:acc_a1_plot}, we compare the $\delta_{1.25^n}$ scores at different accuracy thresholds $n$ of our geometric depth estimation without refinement (`Ours (Geometric)'), our refinement network only (`Ours (Monocular)') and our full method (`Ours'). The results confirm our hypothesis that while geometric estimates are more accurate than monocular depth estimates, they are also noisier. Importantly, the results further illustrate that our full method can effectively combine the strengths of both approaches while not suffering from their weaknesses.

\begin{table}[t]
\centering
\caption{\textbf{Comparison to state-of-the-art on NuScenes.} Evaluated up to 80m for scale-aware depth as in~\cite{FSM, SurroundDepth} and 60m for median-scaled depth (denoted with *) as in~\cite{MCDP}.}
\resizebox{0.85\linewidth}{!}{%
\begin{tabular}{@{}l|cccc@{}}
\toprule
\textbf{Method} & Abs Rel $\downarrow$  & Sq Rel  $\downarrow$ & RMSE  $\downarrow$  & $\delta_{1.25}$  $\uparrow$      \\ \midrule
FSM \cite{FSM}                          & 0.297             & -                 & -                 & -                 \\ 
SurroundDepth \cite{SurroundDepth}      & \underline{0.280} & \textbf{4.401}    & \underline{7.467} & 0.661             \\
\textbf{Ours}                     & \textbf{0.253} 	        & 4.759 	        & \textbf{7.150} 	        & \textbf{0.729}             \\ \midrule
PackNet*\cite{PackNet}              & 0.303             & 3.154             & 7.014             & 0.655             \\
FSM* \cite{FSM}                     & 0.270             & 3.185             & 6.832             & 0.689             \\
SurroundDepth* \cite{SurroundDepth} & 0.245             & \underline{3.067} & 6.835             & \underline{0.719} \\
MCDP* \cite{MCDP}                   & \underline{0.237}    & \textbf{3.030}    & \underline{6.822} & \underline{0.719} \\
\textbf{Ours*}                & \textbf{0.235}             & 3.332             & \textbf{6.021}             & \textbf{0.749}             \\ \bottomrule
\end{tabular}}
\label{tab:comparison_nuscenes_main}\vspace{-2mm}
\end{table}

\begin{table}[t]
\centering
\caption{\textbf{Per-camera evaluation on NuScenes}. We compare our method with previous works on scale-aware depth estimation. Evaluated up to 80m as in Tab.~\ref{tab:comparison_nuscenes_main}. We observe the same trend as on DDAD, \ie our method performs best across all cameras, with a particularly high improvement on the side views.}

\resizebox{0.95\linewidth}{!}{%
\begin{tabular}{@{}l|cccccc @{}}
\toprule
              & \multicolumn{6}{c}{Abs Rel $\downarrow$}                          \\ \cmidrule(l){2-7} 
\textbf{Method} & \textit{Front} & \textit{F.Left} & \textit{F.Right} & \textit{B.Left} & \textit{B.Right} & \textit{Back} \\ \midrule
FSM \cite{FSM}  & 0.186 & 0.287 & 0.375 & 0.296 & 0.418 & 0.221 \\
SurroundDepth \cite{SurroundDepth}  & \underline{0.179} & \underline{0.260} & \underline{0.340} & \underline{0.282} & \underline{0.403} & \underline{0.212} \\ \midrule
\textbf{Ours} & \textbf{0.174}             &	\textbf{0.230}          & \textbf{0.302}          & \textbf{0.249}          & \textbf{0.360}          & \textbf{0.201} \\ \bottomrule %
\end{tabular}}
\label{tab:nuscenes_per_cam}
\end{table}
\begin{table}[t]
\centering
\caption{\textbf{Cross-dataset transfer.} We use models trained on DDAD to evaluate scale-aware depth prediction on the NuScenes dataset. We compare to state-of-the-art methods~\cite{FSM, SurroundDepth} and observe that our method exhibits better generalization ability (cf. Tab.~\ref{tab:comparison_nuscenes_main}).}
\resizebox{0.85\linewidth}{!}{%
\begin{tabular}{@{}l|cccc@{}}
\toprule
\textbf{Method}                     & Abs Rel $\downarrow$ & Sq Rel  $\downarrow$ & RMSE  $\downarrow$  & $\delta_{1.25}$  $\uparrow$   \\ \midrule
FSM \cite{FSM}                      & \underline{0.349} & \underline{5.064} & 8.785 & 0.499 \\
SurroundDepth \cite{SurroundDepth}  & 0.364 & 5.476 & \underline{8.447} & \underline{0.525} \\
\textbf{Ours}                       & \textbf{0.292} & \textbf{4.800} & \textbf{7.677} & \textbf{0.660} \\ \bottomrule %
\end{tabular}}
\label{tab:nuscenes_transfer}\vspace{-4mm}
\end{table}

\parsection{Co-visibility graph construction.} In Tab.~\ref{tab:ablation_covis}, we compare our co-visibility graph construction algorithm (Sec.~\ref{sec:covis_graph}) to a multi-camera adaption of the algorithm in~\cite{DroidSLAM}. Ours yields similar performance in depth estimation, with the exact same Abs. Rel. score and only minor difference in other metrics. Further, the absolute scale ATE is only slightly (25cm) higher. In contrast, using our algorithm, we achieve a runtime reduction of nearly $10\times$ on the full system.

\def \imheight {60pt}
\def \pointcloudwidth {1.0}

\begin{figure*}
\footnotesize
\centering
\setlength\tabcolsep{0.5 pt}

\begin{tabular}{lcccc}
& B.Right & Back & B.Left & 3D Reconstruction \\

\rotatebox[origin=c]{90}{Input / GT} &
\begin{tabular}{l}\includegraphics[height=\imheight]{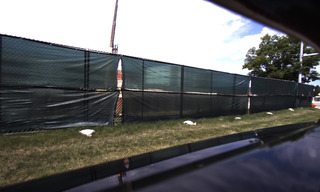}\end{tabular} &
\begin{tabular}{l}\includegraphics[height=\imheight]{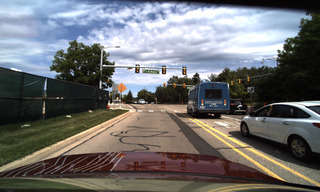}\end{tabular} &
\begin{tabular}{l}\includegraphics[height=\imheight]{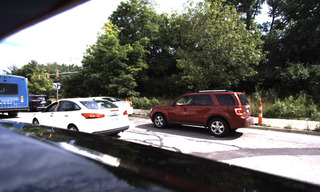}\end{tabular} &
\begin{tabular}{l}\adjincludegraphics[height=\imheight,trim={0 {.15\height} 0 {.25\height}},clip]{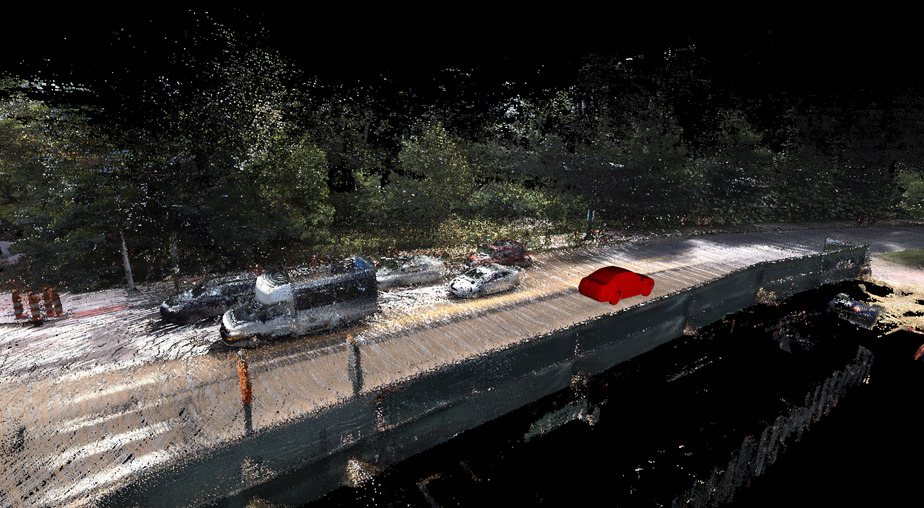}\end{tabular} \\

\rotatebox[origin=c]{90}{FSM~\cite{FSM}} &
\begin{tabular}{l}\includegraphics[height=\imheight]{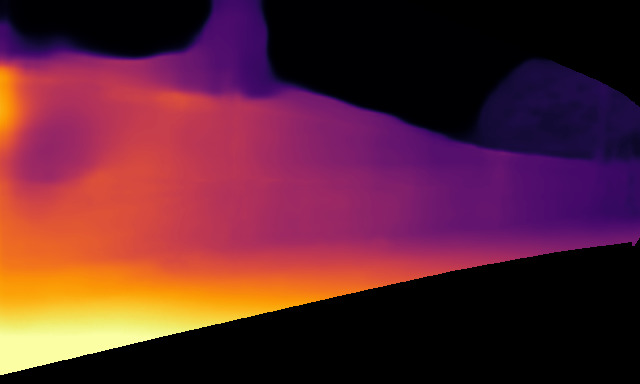}\end{tabular} &
\begin{tabular}{l}\includegraphics[height=\imheight]{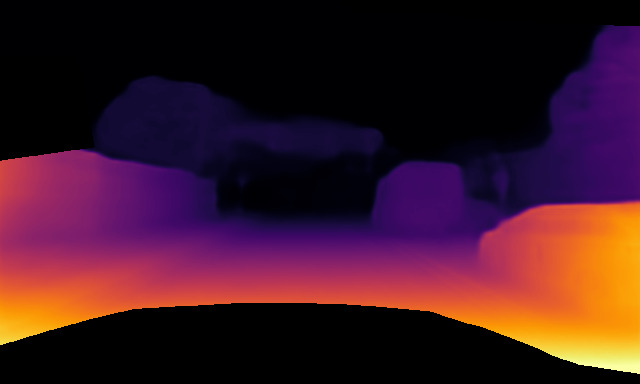}\end{tabular} &
\begin{tabular}{l}\includegraphics[height=\imheight]{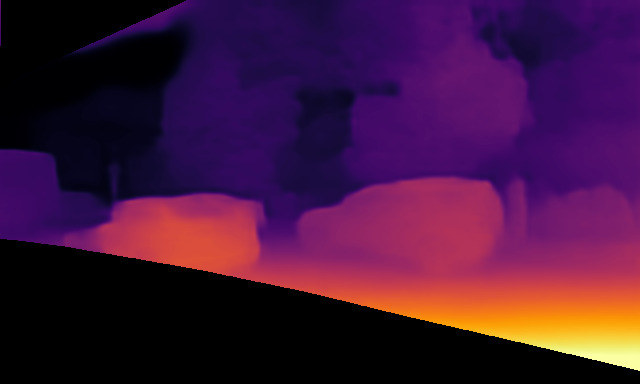}\end{tabular} &
\begin{tabular}{l}\adjincludegraphics[height=\imheight,trim={0 {.15\height} 0 {.25\height}},clip]{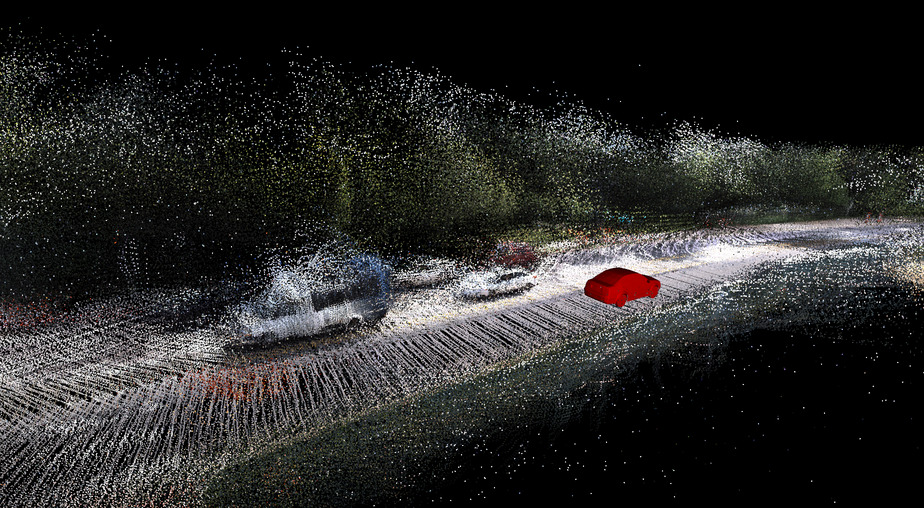}\end{tabular} \\

\rotatebox[origin=c]{90}{SD~\cite{SurroundDepth}} &
\begin{tabular}{l}\includegraphics[height=\imheight]{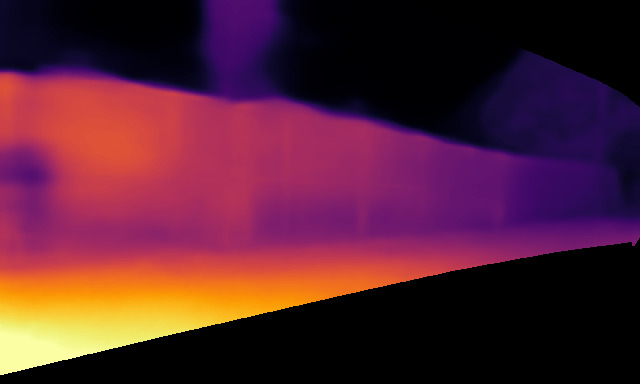}\end{tabular} &
\begin{tabular}{l}\includegraphics[height=\imheight]{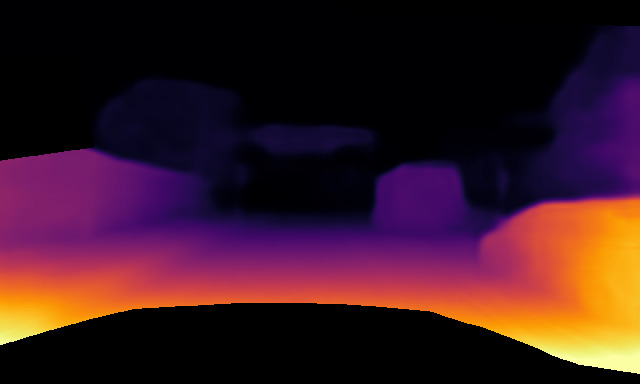}\end{tabular} &
\begin{tabular}{l}\includegraphics[height=\imheight]{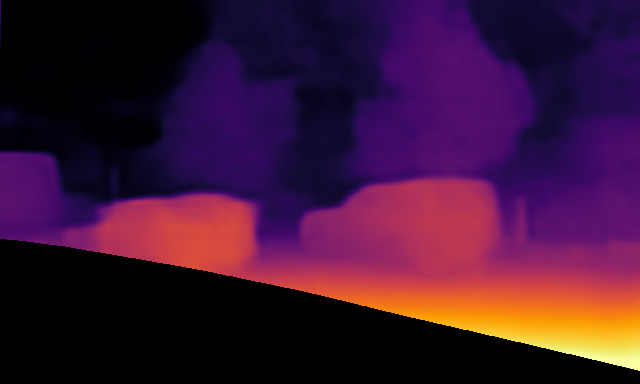}\end{tabular} &
\begin{tabular}{l}\adjincludegraphics[height=\imheight,trim={0 {.15\height} 0 {.25\height}},clip]{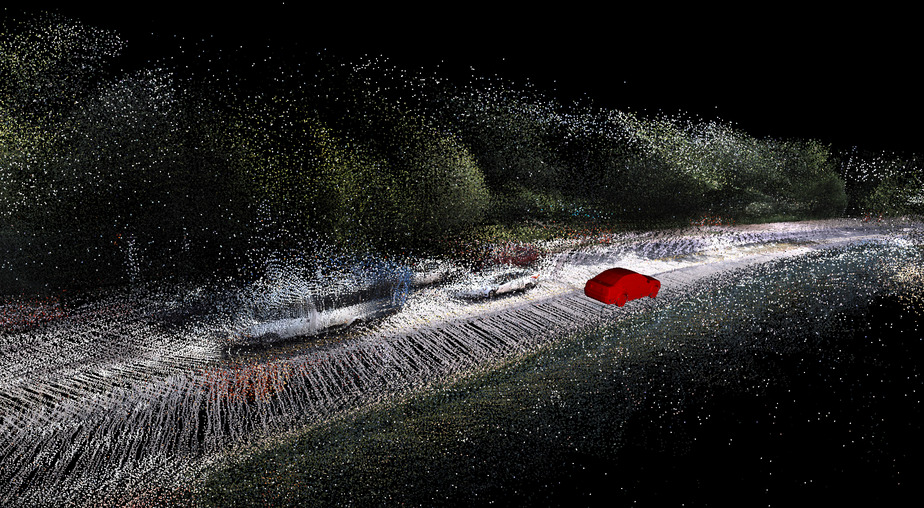}\end{tabular} \\

\rotatebox[origin=c]{90}{Ours} &
\begin{tabular}{l}\includegraphics[height=\imheight]{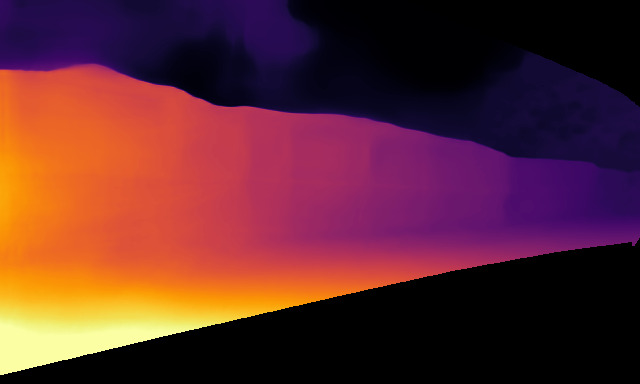}\end{tabular} &
\begin{tabular}{l}\includegraphics[height=\imheight]{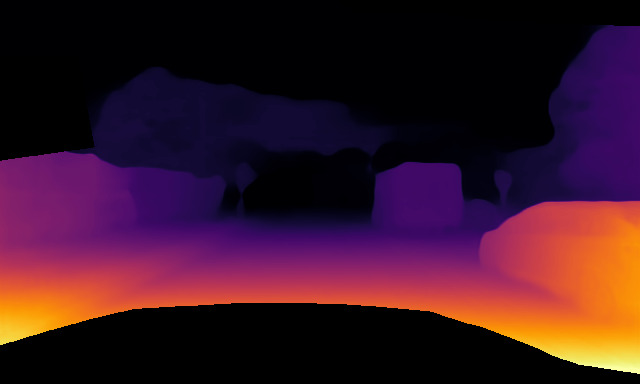}\end{tabular} &
\begin{tabular}{l}\includegraphics[height=\imheight]{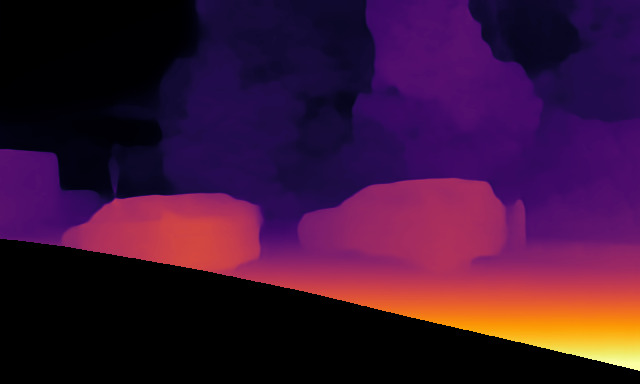}\end{tabular} &
\begin{tabular}{l}\adjincludegraphics[height=\imheight,trim={0 {.15\height} 0 {.25\height}},clip]{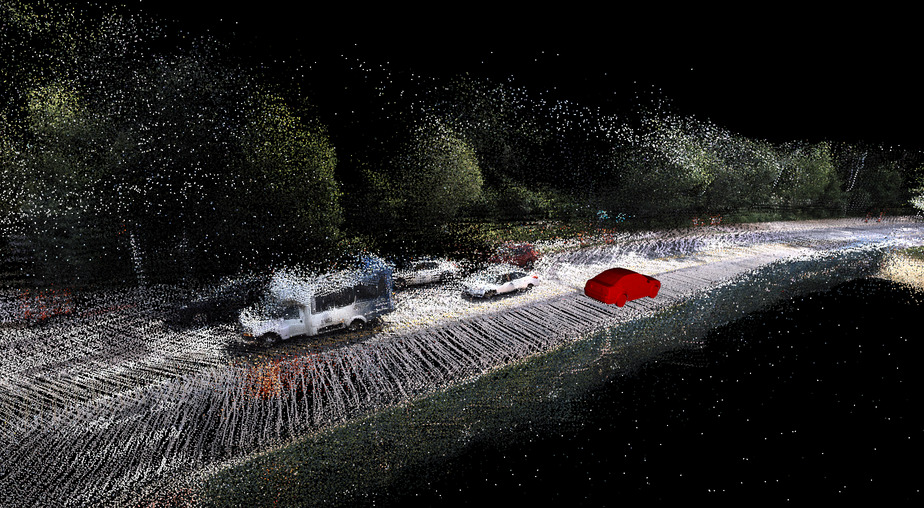}\end{tabular} \\
\end{tabular}

\caption{\textbf{Qualitative comparison on DDAD.} We show depth maps and point clouds accumulated over time along with input images, ground-truth LiDAR 3D Reconstruction, and the ego-vehicle in red. Our depth maps are sharper and more accurate. Our accumulated point clouds yield a more consistent 3D reconstruction. Note that we use our pose predictions for~\cite{FSM, SurroundDepth} since these do not predict pose.}
\label{fig:comparison_ddad}\vspace{-2mm}
\end{figure*}

\parsection{Depth refinement network.} We verify the choice of our inputs to $\mathcal{D}_\theta$ in Tab.~\ref{tab:ablation_depth}.
If we do not input the confidence $\conf_i$ and do not sparsify the depth via $\conf_i < \beta$ before feeding it to $\mathcal{D}_\theta$, we observe the lowest performance. If we input the confidence $\conf_i$ but do not sparsify the depth, we see an improvement over not using them.  We achieve the best performance with the inputs described in Sec.~\ref{sec:depth_refinement}.

Moreover, we show qualitative examples of our depth refinement in Fig.~\ref{fig:refinement}. The confidence $\conf_i$ (overlaid with the image) segments regions where geometric estimation is ambiguous, \ie dynamic objects and uniformly textured areas like the sky and the road surface. The geometric depth maps exhibit severe artifacts in such regions. After refinement, these artifacts are corrected.

\subsection{Comparison to State-of-the-Art}

\parsection{DDAD.} In Tab.~\ref{tab:comparison_ddad_main}, we compare our method to existing self-supervised multi-camera depth estimation methods.
We achieve significant improvements in scale-aware and scale-invariant depth prediction with only minor differences in absolute and relative scale results. In both settings, we achieve state-of-the-art results with the lowest Abs. Rel. scores of 0.162 and 0.169, respectively.

In Fig.~\ref{fig:acc_a1_plot} we compare the $\delta_{1.25^n}$ scores of our method to the state-of-the-art approaches on DDAD at different $n$. For small values of $n$, we observe a particularly pronounced performance gain over existing methods. This shows the advantage of combining highly accurate geometric estimates with monocular cues. Further, our method consistently outperforms the baselines over all values of $n$ for $\delta < 1.25^n$.

In Fig.~\ref{fig:comparison_ddad}, we show a qualitative comparison to existing works. On the left side, we illustrate depth maps across three cameras. Compared to FSM~\cite{FSM} and SurroundDepth~\cite{SurroundDepth}, we produce sharper depth boundaries. On the right side, we illustrate point clouds accumulated over time. We produce a more consistent reconstruction, as can be seen when focusing on the vehicles or the fences beside the road. 

Tab.~\ref{tab:comparision_ddad_per_cam} shows the per-camera breakdown of the scale-aware depth prediction comparison. The advantage of our method is particularly pronounced on the side cameras. Other methods struggle with the side views because the static spatial context is limited and they do not exploit any temporal or spatial-temporal context. The temporal and spatial-temporal contexts are conducive in the side views since sideward motion leads to better triangulation angles than forward motion. This is also evidenced by the results of $\mathcal{D}_\theta$ without geometric estimation, which performs similarly to the baseline methods, struggling with the side views. In contrast, our full system performs well across all cameras.

We further compare to self-supervised single-camera methods that leverage temporal information in Tab.~\ref{tab:comparison_ddad_multiframe}. For a fair comparison, we evaluate only the front camera. We compare favorably to DepthFormer~\cite{DepthFormer} and ManyDepth~\cite{ManyDepth}. In particular, we substantially improve in the outlier robust metric $\delta_{1.25}$ and perform competitively in all others. Our Abs. Rel. score is the lowest at 0.128.

To evaluate ego-motion estimation, we compare our system to state-of-the-art monocular SLAM methods~\cite{DroidSLAM, campos2021orb} in Tab.~\ref{tab:pose_evaluation}. We run the baselines on each camera and report the average ATE across the cameras that successfully tracked the camera path. Note that, contrary to our approach, the competing methods frequently failed to produce a trajectory.
We observe that state-of-the-art monocular SLAM methods exhibit a large error (ATE), while our method recovers accurate trajectories. 

\parsection{NuScenes.} In Tab.~\ref{tab:comparison_nuscenes_main}, we compare our system to previous self-supervised multi-camera depth estimation methods.
We outperform previous methods by a significant margin, corroborating our findings on DDAD.
Notably, we observe substantial improvements in scale-aware depth estimation, while previous works usually struggle with recovering absolute scale on this dataset since the overlap between cameras is smaller than in DDAD~\cite{FSM}. 
Therefore, previous works like MCDP~\cite{MCDP} report results on scale-invariant depth prediction. We outperform existing methods also in this setting, although the performance gap is smaller. Note that the gap between scale-aware and scale-invariant evaluation is only 2 percentage points in $\delta_{1.25}$ for our method, while for SurroundDepth~\cite{SurroundDepth} the gap is much larger with 5.8 percentage points.
We show a per-camera breakdown of the comparison to the state-of-the-art in scale-aware depth estimation in Tab.~\ref{tab:nuscenes_per_cam}. The results corroborate our findings on DDAD, namely that our method outperforms previous works across all cameras with a particularly pronounced improvement on the side views.

\parsection{Cross-dataset transfer.}
We finally test the cross-dataset generalization of our method versus FSM~\cite{FSM} and SurroundDepth~\cite{SurroundDepth}. 
In Tab.~\ref{tab:nuscenes_transfer}, we show \textit{scale-aware} depth estimation results of models trained on DDAD, evaluated on NuScenes. 
We observe that the gap in Abs. Rel. widens compared to Tab.~\ref{tab:comparison_nuscenes_main}, with our method outperforming SurroundDepth by 0.072 in Abs. Rel. and FSM by 0.057.
Further, our method maintains much higher levels of $\delta_{1.25}$ and Sq. Rel., while SurroundDepth drops significantly in both metrics.
Note that we apply focal length scaling~\cite{CNNSLAM} to all methods to accommodate differences in camera intrinsics.

\section{Conclusion}
We introduced \modelname, a multi-camera system for dense 3D reconstruction of dynamic outdoor environments.
The key ideas we presented are a multi-camera DBA operator that greatly improves geometric depth and pose estimation, a multi-camera co-visibility graph construction algorithm that reduces the runtime of our system by nearly $10\times$ without significant performance drop, and a depth refinement network that effectively fuses geometric depth estimates with monocular cues.
We observe that our design choices enable dense 3D mapping in challenging scenes presenting many dynamic objects, uniform and repetitive textures, and complicated camera phenomena like lens flare and auto-exposure.
We achieve state-of-the-art performance in dense depth prediction across two multi-camera benchmarks.

{\small
\bibliographystyle{ieee_fullname}
\bibliography{egbib}
}

\newpage
\appendix

{%
\centerline{\large\bf Appendix}%
\vspace*{12pt}%
\it%
}

This supplementary material provides further details on our method, our experimental setup, and more quantitative and qualitative results and comparisons.
In Sec.~\ref{sec:method}, we provide further details and discussion on the geometric pose and depth estimation, the refinement network, our co-visibility graph, and training and inference procedures.
In Sec.~\ref{sec:experiments}, we provide additional ablation studies and more quantitative and qualitative comparisons and results.
In Sec.~\ref{sec:dba_derivation}, we provide a detailed derivation of our multi-camera DBA.
Note that we group large qualitative comparisons at the end of this document for better readability.

\section{Implementation Details}
\label{sec:method}
\subsection{Geometric Depth and Pose Estimation}
We follow~\cite{DroidSLAM} for the context feature encoder $g_\psi$ and correlation feature encoder $g_\phi$ and GRU architecture. In addition to the context features, $g_\psi$ also outputs an initialization to the GRU hidden state $\hidden_{ij}^{(0)}$ from input $\im_i$.
The hidden state after the last iteration is pooled among all outgoing edges from node $i$ to predict a damping factor $\damping_{i}$ that serves as a parameter to the multi-camera DBA, which improves convergence when the current depth estimate is inaccurate.
We perform feature matching and geometric depth updates at 1/8\textsuperscript{th} of the original image resolution. We predict upsampling masks from the pooled hidden state to upsample depth to the original resolution. The confidence maps $\conf_i$ are linearly interpolated. Furthermore, following RAFT~\cite{RAFT}, the correlation volume is built as a 4-level pyramid, and sampled features from all layers are concatenated. Further, each GRU update is followed by a multi-camera DBA with two Gauss-Newton steps.

\subsection{Depth Refinement}
We show the architecture of the depth refinement network in Tab.~\ref{tab:refinement_network}. Note that we input full-resolution frames and up-sampled depth and confidence. Further, we concatenate the depth and confidence predicted at 1/8\textsuperscript{th} scale with features after the third skip connection.
We use the inverse of the input depth. Further, we obtain the refined depth prediction from the sigmoid output $\mathbf{o}_t^c \in [0, 1]^{H \times W}$ via

\begin{equation}
    \frac{1}{\depth_t^c} = \frac{f_{c}}{f_\text{norm}} \cdot \Bigl( \frac{1}{d_\text{max}} + (\frac{1}{d_\text{min}} - \frac{1}{d_\text{max}}) \cdot \mathbf{o}_t^c \Bigr) \, ,
\end{equation}

where $f_{c}$ is the focal length, $f_\text{norm}$ a constant normalization factor and $d_\text{min}$ and $d_\text{max}$ are pre-defined minimum and maximum depth. For experiments on the DDAD dataset we set $d_\text{min}=1$, $d_\text{max}=200$, $f_\text{norm} = 715$ aligned with the focal length of the front-facing camera, for NuScenes we choose $d_\text{min}=1$, $d_\text{max}=80$, $f_\text{norm} = 500$.

\parsection{Dataset generation.}
To train the refinement network, we first generate a dataset of samples that contain the prior geometric depth, confidence maps, and poses with the first two stages of our system. 
We filter scenes with inaccurate scene scale by measuring how many reliable feature matches there are across both temporal and spatial edges with the given confidences $\conf_{ij}$. The fewer reliable matches, the weaker the constraint on the metric scale. Furthermore, based on the generated poses, we remove static scenes. This allows us to train an absolute scale monocular depth estimation model from the raw video data.

\parsection{Training details.}
During the training of refinement network $\mathcal{D}_\theta$, we randomly set input depth and confidence weights to zero to learn depth prediction with and without prior geometric estimates as input. We use color-jitter and random horizontal flipping as augmentations.
Further, we follow a two-stage training paradigm as in~\cite{TwoStageTraining}. After training $\mathcal{D}_\theta$ in the first stage, we remove training samples with outliers in the depth estimates of the current $\mathcal{D}_\theta$. In particular, we apply RANSAC to determine the ground plane in the front view and calculate the height of each pixel in all views. We omit training samples with $\frac{1}{H \cdot W} \sum_{u, v}{[h_{u, v} < -0.5m]}>\hthresh$ where $h_{u, v}$ is the height of pixel $(u, v)$ w.r.t. the ground plane and $\hthresh$ is set to 0.005 for the front and backward-facing cameras and 0.02 for the side views. This filters frames where a significant amount of pixels are below the ground plane. In the second stage, we re-train the network from scratch on the filtered dataset for 20 epochs with the same settings. On NuScenes, we train our refinement network with all available camera images (12Hz) instead of only keyframes (2Hz).

\begin{table*}[]
\caption{\textbf{Depth refinement network architecture.} $K$ describes the kernel size, $S$ the stride. ResidualBlock consists of two convolutional layers and a skip-connection as proposed in~\cite{ResNet}. We generate output at four scales in $[0, 1]$ which is normalized by focal length of the respective camera and scaled to $[1/d_{max}, 1/d_{min}]$.}
\centering
\footnotesize
\begin{tabular}{@{}lclllc@{}}
\toprule
No.  & Input       & Layer Description                                & K & S & Output Size                   \\ \midrule
\multicolumn{6}{c}{\textbf{(\#A, \#B) UpBlock}}                                                      \\ \midrule
\#i  & (\#A)       & Conv2d$\rightarrow$ ELU $\rightarrow$ Up          & 3 & 1 &                               \\
\#ii & (\#i, \#B)  & Concatenate $\rightarrow$ Conv $\rightarrow$ ELU & 3 & 1 &                               \\ \midrule
\multicolumn{6}{c}{\textbf{Encoder}}                                                                          \\ \midrule
\#0  &             & \textbf{Input}: Image + Inv. Geometric Depth + Confidence &   &   & $5 \times H \times W$         \\
\#1 & (\#0) & Conv $\rightarrow$ BN $\rightarrow$ ReLU                & 7 & 2 & $64 \times H/2 \times W/2$ \\
\#2  & (\#1)       & MaxPool          $\rightarrow$ \textbf{Skip}     & 3 & 2 & $64 \times H/4 \times W/4$    \\
\#3  & (\#2)       & 2xResidualBlock  $\rightarrow$ \textbf{Skip}     & 3 & 1 & $64 \times H/4 \times W/4$    \\
\#4  & (\#3)       & 2xResidualBlock  $\rightarrow$ \textbf{Skip}     & 3 & 2 & $128 \times H/8 \times W/8$   \\
\#5  &             & \textbf{Input}: Inv. Geometric Depth + Confidence         &   &   & $2 \times H/8 \times W/8$     \\
\#6  & (\#3, \#5)  & Concatenate                                      &   &   & $130 \times H/8 \times W/8$   \\
\#7  & (\#6)       & 2xResidualBlock  $\rightarrow$ \textbf{Skip}     & 3 & 2 & $256 \times H/16 \times W/16$ \\
\#8  & (\#7)       & 2xResidualBlock                                  & 3 & 2 & $512 \times H/32 \times W/32$ \\ \midrule
\multicolumn{6}{c}{\textbf{Decoder}}                                                                          \\ \midrule
\#9  & (\#8, \#7)  & UpBlock                                          & 3 & 1 & $256 \times H/16 \times W/16$ \\
\#10 & (\#9, \#4)  & UpBlock                                          & 3 & 1 & $128 \times H/8 \times W/8$   \\
\#11 & (\#10)      & Conv2d $\rightarrow$ Sigmoid $\rightarrow$ \textbf{Output}& 3 & 1 & $1 \times H/8 \times W/8$     \\
\#12 & (\#10, \#3) & UpBlock                                          & 3 & 1 & $64 \times H/4 \times W/4$    \\
\#13 & (\#12)      & Conv2d $\rightarrow$ Sigmoid $\rightarrow$ \textbf{Output}& 3 & 1 & $1 \times H/4 \times W/4$     \\
\#14 & (\#12, \#2) & UpBlock                                          & 3 & 1 & $32 \times H/2 \times W/2$    \\
\#15 & (\#14)      & Conv2d $\rightarrow$ Sigmoid $\rightarrow$ \textbf{Output}& 3 & 1 & $1 \times H/2 \times W/2$     \\
\#16 & (\#15)      & UpBlock                                          & 3 & 1 & $16 \times H \times W$        \\
\#17 & (\#16)      & Conv2d $\rightarrow$ Sigmoid $\rightarrow$ \textbf{Output}& 3 & 1 & $1 \times H \times W$         \\ \bottomrule
\end{tabular}
\label{tab:refinement_network}
\end{table*}

\subsection{Co-visibility Graph}
We detail our multi-camera co-visibility graph construction described in \ifpapersuppl{Sec.~\ref{sec:covis_graph}}{Sec.~3.1} of the main paper in Algorithm~\ref{alg:covis_graph}.
Note that \texttt{GetAdjacentNodes} returns a different adjacency pattern than the spatial edges in $A$, as described in \ifpapersuppl{Sec.~\ref{sec:covis_graph}}{Sec.~3.1} of the main paper.
In particular, we leverage the forward motion assumption in order to connect two frames $(i, j)$ if camera $c_j$ is closer to the forward-facing camera than camera $c_i$. For further clarification, please refer to \ifpapersuppl{Fig.~\ref{fig:covis_graph}}{Fig.~3} of the main paper.
\begin{algorithm}
\caption{Co-visibility graph construction}
\begin{algorithmic}[1]
    \INPUT $\intrinsic, \extrinsics, \covisgraph, \{\im_t^c\}_{c=0}^{C}$
    \State $A$ = \texttt{ComputeStaticAdjacency}($\intrinsic$, $\extrinsics$)
    
    \State $N$  = \texttt{GetNodes}($\covisgraph$)
    \State $M$ = \texttt{AddNodes}($\covisgraph$, $\im_t^c$)

    \For{$i \in M$}\LineComment{add temporal edges}
        \For{$j \in N$}
            \If{\texttt{Radius}($i$, $j$) $< r_\text{intra}$} 
                \State \texttt{AddEdge}($\covisgraph$, $i$, $j$)
            \EndIf
        \EndFor
    \EndFor
    
    \For{$(i, j) \in A$}\LineComment{add spatial edges}
        \State \texttt{AddEdge}($\covisgraph$, $i$, $j$)
    \EndFor\LineComment{add spatial-temporal edges}
    \State $O$ = \texttt{GetNodesAtTime}($\covisgraph$, $t-r_\text{intra}$)
    \State $O'$ = \texttt{GetAdjacentNodes}($\covisgraph$, $A$, $M$, $O$)
    \For{$(i, j) \in O'$}
        \State \texttt{AddEdge}($\covisgraph$, $i$, $j$)
    \EndFor
    \State \LineComment{Remove out-of-context edges and nodes}
    \State \texttt{RemoveTemporalEdges}($\covisgraph$, $t-\Delta_{t_\text{intra}}$)
    \State \texttt{RemoveSpatialTemporalEdges}($\covisgraph$, $t-\Delta_{t_\text{inter}}$)    
    \State \texttt{RemoveUnconnectedNodes}($\covisgraph$)
\end{algorithmic}
\label{alg:covis_graph}
\end{algorithm}

\parsection{Dynamic alternative.} We further implement a dynamic algorithm without the assumptions stated in \ifpapersuppl{Sec.~\ref{sec:covis_graph}}{Sec.~3.1} of the main paper. 
The dynamic algorithm establishes edges based on camera frustum overlap, \ie it measures the intersection over union (IoU) of the camera frustums of each frame across a local time window in world space given the current camera pose estimates $\pose$. We order frame pairs by their IoU in descending order and choose the $N$ highest overlapping pairs as co-visible frames. These establish the temporal and spatial-temporal edges. 

We found empirically that the static algorithm performs similarly to the dynamic alternative while being simpler and more efficient, so we use it in our experiments. However, for applications where the assumptions in \ifpapersuppl{Sec.~\ref{sec:covis_graph}}{Sec.~3.1} of the main paper do not hold, this algorithm provides a suitable alternative.

\subsection{Inference Details}

For both datasets, we observe that camera shutters are not well synchronized in both datasets. This poses a problem, especially at high speeds. Thus, instead of using constant camera extrinsics, we compute time-dependent relative camera extrinsics.
For inference, we set $n_\text{itr-wm} = 16$, $n_\text{iter1} = 4$ and $n_\text{iter2} = 2$. For Nuscenes, use a different threshold $\beta = 0.8$.

\section{Experiments}
\label{sec:experiments}

\parsection{Evaluation metrics.}
We evaluate the proposed method in terms of depth accuracy and trajectory accuracy.

\parsection{Depth.}
Given the estimated depths $\depth_t^c$ and ground truth depth ${\depth_t^*}^c$ we use compare the following depth metrics

\ifpapersuppl{
\begin{equation}
    \text{Abs Rel:}\; \frac{1}{T \cdot C} \sum_{d, c}{\frac{|\depth_t^c - {\depth_t^*}^c|}{{\depth_t^*}^c}}
\end{equation}
\begin{equation}
    \text{Sqr Rel:}\; \frac{1}{T \cdot C} \sum_{d, c}{\frac{\|\depth_t^c - {\depth_t^*}^c\|^2}{{\depth_t^*}^c}}
\end{equation}
\begin{equation}
    \text{RMSE:}\; \frac{1}{T \cdot C} \sqrt{\sum_{d, c}{\|\depth_t^c - {\depth_t^*}^c\|^2}}
\end{equation}
\begin{equation}
    \text{$\delta_{1.25}$:}\; \text{fraction of} \; d \in \depth \; \text{for which} \; \max{\left(\frac{d}{d^*}, \frac{d^*}{d}\right) < 1.25}
\end{equation}
}{
\begin{equation}
\begin{split}
    \text{Abs Rel:}\; &\frac{1}{T \cdot C} \sum_{d, c}{\frac{|\depth_t^c - {\depth_t^*}^c|}{{\depth_t^*}^c}} \\
    \text{Sqr Rel:}\; &\frac{1}{T \cdot C} \sum_{d, c}{\frac{\|\depth_t^c - {\depth_t^*}^c\|^2}{{\depth_t^*}^c}} \\
    \text{RMSE:}\; &\frac{1}{T \cdot C} \sqrt{\sum_{d, c}{\|\depth_t^c - {\depth_t^*}^c\|^2}} \\
    \text{$\delta_{1.25}$:}\; & \text{fraction of} \; d \in \depth \; \text{for which} \; \max{\left(\frac{d}{d^*}, \frac{d^*}{d}\right) < 1.25} \\
\end{split}
\end{equation}
}

For up-to-scale evaluation, we resort to the camera-wise metric scaling as described in~\cite{FSM} which results in scaling factor $s$ described as

\begin{equation}
    s = \frac{1}{C}\cdot \sum_{c}{\frac{median({d^*}^c)}{median(d^c)}}
\end{equation}

We then scale the predicted depth as $s \cdot \depth$.

\parsection{Trajectory.} For trajectory evaluation we use the absolute trajectory error (ATE) score. Given an estimated trajectory $\egopose_1, ..., \egopose_T \in SE(3)$ and ground truth trajectory $\textbf{Q}_1, ..., \textbf{Q}_T$ the ATE is defined as

\begin{equation}
\begin{split}
    \textbf{F}_i &= \textbf{Q}_i^{-1} \textbf{S} \egopose_i \\
    ATE &= RMSE(\textbf{F}_{1:T}) = \sqrt{\frac{1}{T}\sum_{i}{\|trans(\textbf{F}_i)\|^2}}
\end{split}
\end{equation}

where $trans$ defines the translational part of $\textbf{F}$ and $\textbf{S}$ is identity $\textbf{I}$ for unscaled evaluation or $s_{traj} \cdot \textbf{I}$ where $s_{traj}$ is determined via least squares for scaled evaluation.

\parsection{Ablation studies.}
In Tab.~\ref{tab:ablation_vkitti}, we show a comparison of our depth refinement network trained with synthetic data only, with both synthetic data and real-world data, and real-world data only. As stated in \ifpapersuppl{Sec.~\ref{sec:depth_refinement}}{Sec.~3.3} of the main paper, the results show that synthetic data cannot help the depth refinement network performance, even when fine-tuning on real-world data afterward. Instead, we observe the best performance when starting training from real-world data directly. This underlines the importance of our self-supervised training scheme for the refinement network, since contrary to the geometric parts of our system, here we cannot rely on synthetic data to provide us with ground-truth supervision.

For Tab.~\ref{tab:ablation_mono_weights}, we train two versions of our refinement network described in \ifpapersuppl{Sec.~\ref{sec:depth_refinement}}{Sec.~3.3} of the main paper. We train $\mathcal{D}_{\theta}$ as described in the main paper with sparsified input depth and train $\mathcal{D}_{\omega}$ purely for monocular depth estimation without refining geometric estimates. We evaluate both networks on monocular depth estimation on DDAD. We observe that the networks perform similarly, while $\mathcal{D}_{\theta}$ can both refine geometric depth estimates and estimate depth without geometric depth input. This shows that the refinement network learns strong scene priors that can estimate depth even without any additional input. Further, we can conclude that the network generalizes well to both depth prediction and depth refinement.

\begin{figure*}[t]
\centering%
\includegraphics[width=0.33\linewidth]{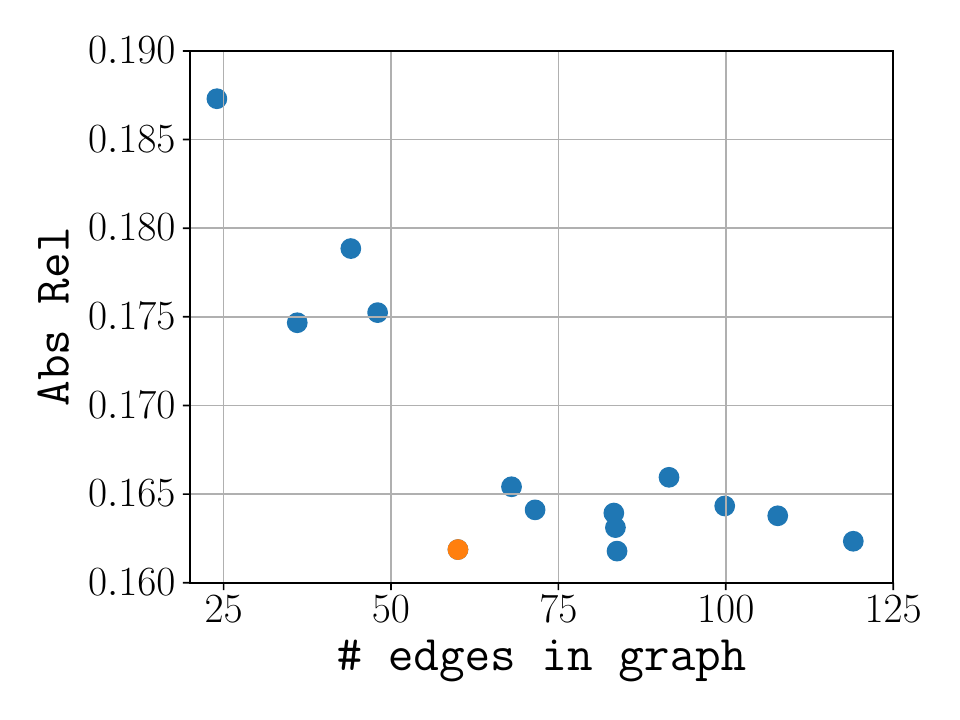}%
\includegraphics[width=0.33\linewidth]{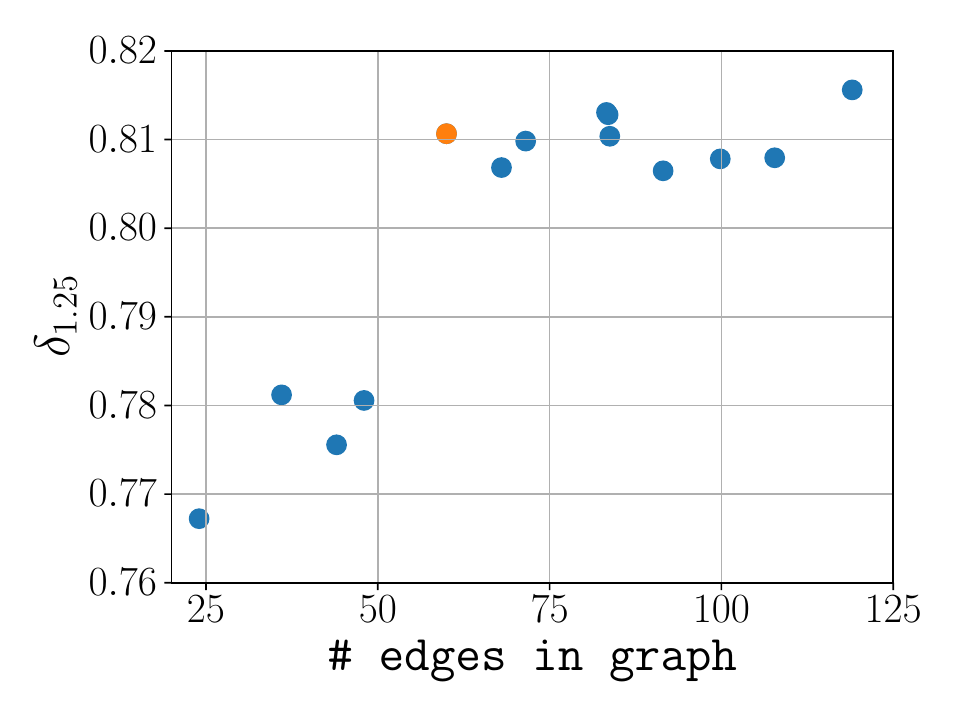}%
\includegraphics[width=0.33\linewidth]{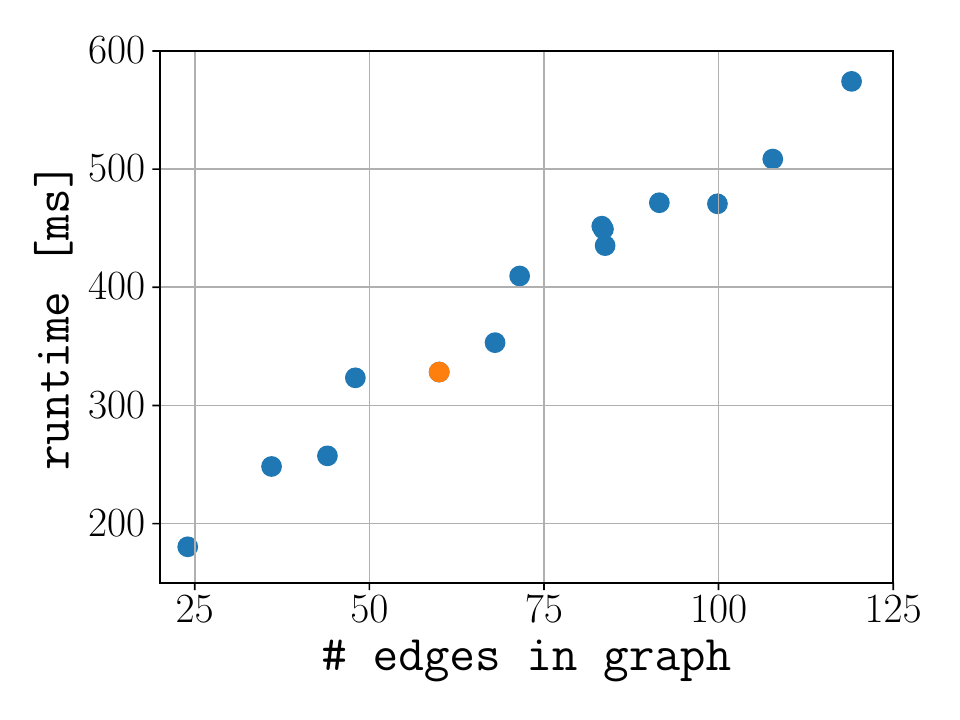}%
\caption{\textbf{Graph density analysis on DDAD.} We plot the AbsRel, $\delta_{1.25}$ scores and runtime w.r.t.\ the number of edges in the co-visibility graph. Our default setting is shown in \textcolor{orange}{orange}.} \vspace{-2mm}
\label{fig:graph_density}
\end{figure*}
In Fig.~\ref{fig:graph_density} we show an ablation of the covisibility graph density by varying the parameters described in \ifpapersuppl{Sec.~\ref{sec:covis_graph}}{Sec.~3.1} of our paper. An increase in the number of edges in the co-visibility graph only yields marginal improvement, while resulting in a linear increase in runtime.

\begin{table}[t]
\caption{\textbf{Masking scheme ablation on DDAD.} We compare the influence of the different masks used to train the refinement network (cf. Eq. 8 of the main paper) on the final performance. }
\centering
\resizebox{0.85\linewidth}{!}{%
\begin{tabular}{@{}ccc|cccc@{}}
    \toprule
    $\mask^\text{st}$ & $\mask^\text{oc}$ & $\mask^\text{fc}$ & Abs Rel $\downarrow$ & Sq Rel $\downarrow$ & RMSE $\downarrow$ & $\delta_{1.25}$ $\uparrow$ \\ \midrule
    
     & & & 0.637 & 51.086 & 18.129 & 0.779 \\ 
    \checkmark & & & 0.288 & 10.304 & 12.982 & 0.786 \\
    \checkmark & \checkmark & & \textbf{0.162} & 3.304 & 11.638 & \textbf{0.811} \\ %
    \checkmark & \checkmark &\checkmark & \textbf{0.162}   & \textbf{3.019}  & \textbf{11.408} & \textbf{0.811} \\\bottomrule 
\end{tabular}
}
\label{tab:ablation_flow_consistency_mask}
\end{table}

During training of $\mathcal{D}_\theta$, we mask regions that do not provide useful information for depth learning when minimizing the view synthesis loss in \ifpapersuppl{Eq.~\ref{eq:masked_loss}}{Eq.~8} of the paper. We provide an ablation study of the three masks used in \ifpapersuppl{Eq.~\ref{eq:masked_loss}}{Eq.~8} in Tab.~\ref{tab:ablation_flow_consistency_mask}. The self-occlusion $\mask^\text{oc}$ and static $\mask^\text{st}$ masks are essential to depth learning by removing ego-vehicle and \eg sky regions, respectively. The flow consistency mask $\mask^\text{fc}$ further reduces outliers, caused by \eg dynamic objects.

\begin{table}[t]
\centering
\caption{\textbf{Refinement network training.} We show that training and pre-training the refinement network on the synthetic VKITTI~\cite{VKITTI2} dataset does not yield improvement over self-supervised training with real-world data as proposed in Sec.~3.3 of the main paper.}
\resizebox{0.9\linewidth}{!}{%
\begin{tabular}{cc|cccc}
\toprule
VKITTI & DDAD & Abs Rel $\downarrow$ & Sq Rel $\downarrow$ & RMSE $\downarrow$ & $\delta_{1.25}$ $\uparrow$ \\ \midrule
\checkmark &             & 0.282 & 5.025 & 15.281 & 0.572 \\ 
\checkmark & \checkmark & 0.163 & 3.313 & 11.580 & 0.809 \\
           & \checkmark & 0.162 & 3.019 & 11.408 & 0.811 \\ \bottomrule
\end{tabular}}
\label{tab:ablation_vkitti}
\end{table}
\begin{table}[t]
\centering
\caption{\textbf{Refinement network training comparison.} We train $\mathcal{D}_{\theta}$ as described in the main paper with sparsified input depth and train $\mathcal{D}_{\omega}$ purely for monocular depth estimation without prior geometric depth input. We evaluate both networks on monocular depth estimation \textit{without geometric depth input} on DDAD. }
\resizebox{0.95\linewidth}{!}{%
\begin{tabular}{cc|cccc}
\toprule
$\mathcal{D}_{\omega}(\im_{t}^c)$  &  $\mathcal{D}_{\theta}(\im_{t}^c, \mathbf{0}, \mathbf{0})$  & Abs Rel $\downarrow$ & Sq Rel $\downarrow$ & RMSE $\downarrow$ & $\delta_{1.25}$ $\uparrow$ \\ \midrule
\checkmark &            & 0.204 & 3.583 & 12.652 & 0.723 \\ 
           & \checkmark & 0.211 & 3.806 & 12.668 & 0.715 \\
           \bottomrule
\end{tabular}}
\label{tab:ablation_mono_weights}
\end{table}

\parsection{Qualitative comparisons.}
In Fig.~\ref{fig:qualitative_ddad_000163}, we qualitatively compare our method to existing works in terms of scale-aware depth estimation on DDAD. Further, we show our geometric depth estimate alongside the refined depth. We notice that, especially for the side views, existing works struggle to obtain accurate depth. On the other hand, the geometric depth of our method produces many accurate depth predictions, but is at the same time very noisy, especially in low-textured areas and for dynamic objects. Our full method demonstrates the best performance.

 In Fig.~\ref{fig:nuscenes_examples}, we show a comparison of our method to the state-of-the-art approach SurroundDepth~\cite{SurroundDepth} on the NuScenes dataset. We observe that our approach produces significantly sharper and more accurate depth maps for all three of the examples.

 Finally, we show additional comparison on accumulated point clouds on DDAD in Fig.~\ref{fig:ddad_pointclouds} and on NuScenes in Fig.~\ref{fig:nuscenes_pointclouds}. Our method produces significantly more consistent 3D reconstructions than competing methods, as can be seen by the more consistent reconstruction of road markings, sidewalks, and trucks in Fig.~\ref{fig:ddad_pointclouds}. In Fig.~\ref{fig:nuscenes_pointclouds} we observe that our method approaches the 3D reconstruction accuracy of the LiDAR-based 3D reconstruction.

\begin{table}[t]
\centering
\caption{\textbf{Inference runtime analysis.} We list the runtime of each component of our system during inference.}
\resizebox{1.0\linewidth}{!}{%
\begin{tabular}{@{}lccc@{}}
\toprule
Component       & Complexity                  & Runtime {[}ms{]} & Percent {[}\%{]} \\ \midrule
Feature encoder       & $\mathcal{O}(|C|)$            & 5                         & 1.4                     \\
Context encoder       & $\mathcal{O}(|C|)$            & 5                         & 1.4                      \\
Create corr. volumes  & $\mathcal{O}(|\edges_{new}|)$          & 13                        & 3.7                     \\
Corr. volume sampling & $\mathcal{O}(n_\text{iter} \cdot |\edges|)$ & 31                        & 8.8                    \\
GRU steps             & $\mathcal{O}(n_\text{iter} \cdot |\edges|)$ & 196                       & 55.4                  \\
Multi-cam. DBA steps             & $\mathcal{O}(n_\text{iter})$            & 1                         & 0.3                    \\
Completion            & $\mathcal{O}(|\nodes|)$                & 98                        & 27.7                    \\
Others                & -                                 & 5                         & 1.4                   \\ \midrule
Total                 &                                   & 354                       &  100                      \\ \bottomrule
\end{tabular}}
\label{tab:timing}
\end{table}
 	 		
\parsection{Runtime breakdown and memory consumption.}
In Tab.~\ref{tab:timing} we show a component-wise breakdown of time-complexity and measured runtime of our approach. Compared to~\cite{DroidSLAM}, we tackle a more complex scenario with six images per timestep. This leads to many more possible edges in the co-visibility graph, increasing the computational burden. Thus, the runtimes are slower than reported in~\cite{DroidSLAM}. However, Tab.~\ref{tab:timing} shows that runtime is dominated by the GRU, which scales with the number of edges $|\edges|$. The breakdown and our observed $10\times$ runtime improvement are both at test time. The peak GPU memory consumption in inference with our parameter setting is 6.08 GB ($\sim$61 MB per edge).

\parsection{Limitations.}
The components of our system rely on deep neural networks with downsampling operations. This means a large chunk of the computation will happen at a lower resolution. While this provides a computational advantage, it also comes at the cost of losing high-frequency details that are important for thin structures like fences. In Fig.~\ref{fig:limitations}, we show an example of this phenomenon where our depth estimate results in a large error because there is an ambiguity between the background and the foreground fence. Similarly, other thin structures like poles could be missed, especially if they are far away.

\def \imgpath {supplementary/figures/qualitative_results/depth_maps_ddad/000163/img/000080}
\def \absrelpath {supplementary/figures/qualitative_results/depth_maps_ddad/000163/abs_rel/000080}
\def \depthpath {supplementary/figures/qualitative_results/depth_maps_ddad/000163/depth/000080}

\begin{figure}
\centering
\setlength\tabcolsep{0.2mm}

\resizebox{1.0\linewidth}{!}{
\begin{tabular}{lcc}
\includegraphics[width=0.34\linewidth]{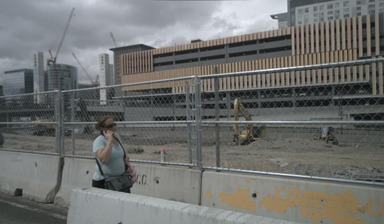} & \includegraphics[width=0.34\linewidth]{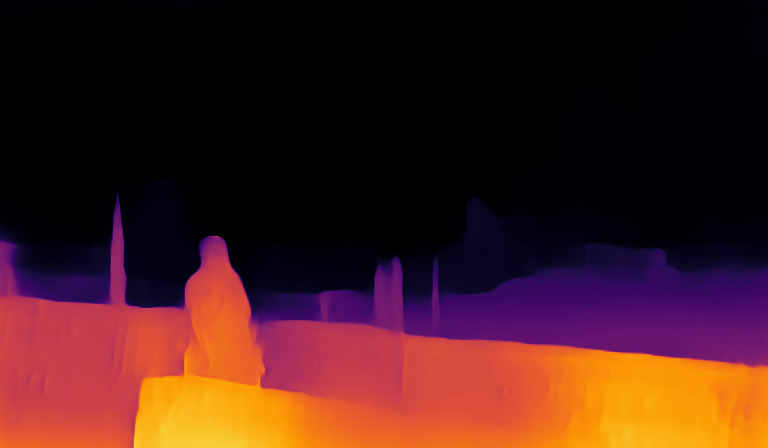} & \adjincludegraphics[width=0.34\linewidth,trim={0 0 0 0},clip]{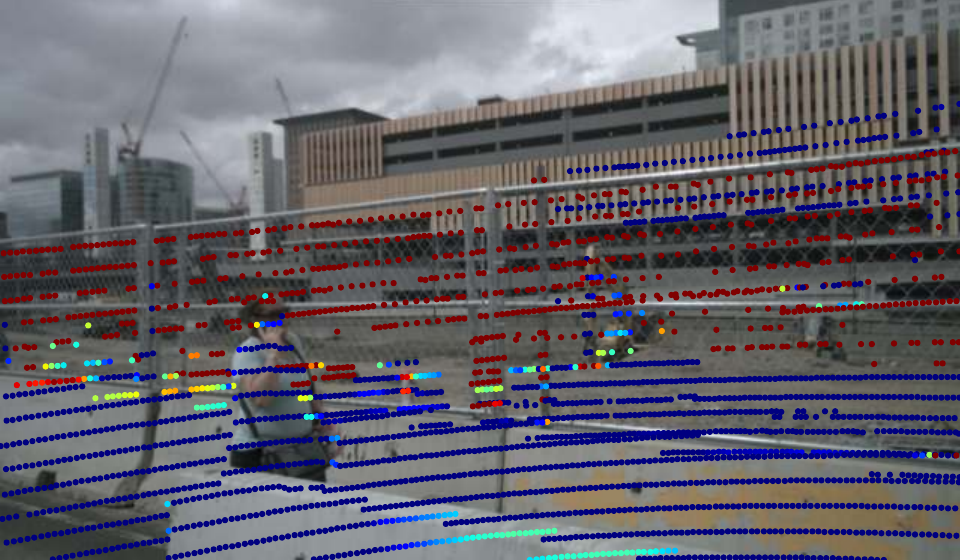} \\
\end{tabular}}
\caption{\textbf{Illustration of thin structures.} We depict an example frame, prediction, and error map from the NuScenes dataset. The fence in the picture poses a problem for our depth estimator since it is partially transparent. Further, since behind the fence, there is a large free space, we observe a large Sq. Rel. score for the areas that are predicted as background.}
\label{fig:limitations}
\end{figure}

\section{Multi-Camera DBA}
\label{sec:dba_derivation}
We provide a detailed derivation of our multi-camera DBA.
The goal of the bundle adjustment is to minimize the energy function $E$, \ie align edge-wise relative poses $\pose_{ij}$ and frame-wise depth $\depth_i$ whose reprojection minimizes the Mahalanobis distance to the estimated flow $\flow_{ij}$ over all edges $\edges$ in the co-visibility graph. In the following, let $H$ and $W$ be the depth map's height and width and $T = |\nodes| / C$ the number of timesteps we consider, where $C$ is the number of cameras.

\begin{equation}
    E = \sum_{(i, j) \in \mathcal{E}} \left\| \coords_{ij} - \Pi_{c_j}(\pose_{ij} \circ \Pi_{c_i}^{-1}(\depth_i)) \right\|_{\covar_{ij}}^2
\label{eq:energy_function}
\end{equation}

With $\covar_{ij} = \operatorname{diag}(\conf_{ij})$, $\coords_{ij} = \mathbf{x}_i + \flow_{ij}$ and

\begin{equation}
    \pose_{ij} = \left(\egopose_{t_j} \extrinsics_{c_j}\right)^{-1} \egopose_{t_i} \extrinsics_{c_i}
\end{equation}

where $\extrinsics_{c_i}$ and $\extrinsics_{c_j}$ are known constants and $\egopose_{t_i}$ and $\egopose_{t_j}$ are optimization variables. This will lead to updates

\begin{equation}
\begin{split}
    \egopose^{(k)} &= \exp{(\delta \logpose)} \egopose^{(k-1)} \\
    \depth^{(k)} &= \depth + \depth^{(k-1)}
\end{split}
\end{equation}

where $\delta \logpose$ and $\delta \depth$ are the solution to the normal equation

\begin{equation}
\begin{split}
    \jacob^\top \covar \jacob \cdot \begin{bmatrix}
        \delta \logpose \\ \delta \depth
    \end{bmatrix} = -\jacob^\top  \covar\residual
\end{split}
\label{eq:gauss_newton}
\end{equation}

where $\jacob \in \real^{|\edges| \cdot H \cdot W \cdot 2 \times (T \cdot 6 + |\nodes| \cdot H \cdot W)}$ is the Jacobian of the residuals w.r.t. all optimization variables and $\residual \in \real^{|\edges| \cdot H \cdot W \cdot 2}$ is the vector of all residuals

\subsection{ Pose - Depth Decomposition}
\label{sec:dba_decomposition}
We can make three observations. First, pose $\egopose_k$ only appears in $\residual_{ij}$ if $k$ is either $i$ or $j$ and $t_i \neq t_j$. Second, there are no loops, thus $i \neq j$. Third, depth $\depth_k$ only appears in $\residual_{ij}$ if $k = i$, therefore $\frac{\partial \mathbf{r}_{ij}}{\partial \mathbf{d}_k} = 0 \;\forall k \neq i$. \par
Let us now consider a single edge $(i, j) \in \edges$ with $\residual_{ij} \in \real^{H \cdot W \cdot 2}$ the residuals and $\jacob_{ij} \in \real^{H \cdot W \cdot 2 \times (12 + H \cdot W)}$ the Jacobian w.r.t. all optimization variables. We can decompose the Jacobian into its components, \ie  $\jacob_{ij} = \begin{bmatrix} \jacob_{\logpose_i} + \jacob_{\logpose_j} & \jacob_{\depth_i} \end{bmatrix} $ where $\jacob_{\depth_i}$ is diagonal. 
Now, when only considering the aforementioned edge, Eq.~\ref{eq:gauss_newton} can be written as

\begin{equation}
    \begin{bmatrix}
    \schurrB_{ii} & \schurrB_{ij} & \schurrE_{ii}\\
    \schurrB_{ji} & \schurrB_{jj} & \schurrE_{ji}\\
    \schurrE_{ii}^\top & \schurrE_{ji}^\top & \schurrC_i   \\
    \end{bmatrix}
    \begin{bmatrix}
    \delta \logpose_i\\
    \delta \logpose_j\\
    \delta \mathbf{d}_i\\
    \end{bmatrix} = 
    \begin{bmatrix}
    \mathbf{v}_i\\
    \mathbf{v}_j\\
    \mathbf{w}_i\\
    \end{bmatrix} 
\end{equation}

with

\begin{align}
\schurrB_{ii} &=  \jacob_{\logpose_i}^\top \covar_{\residual_{ij}} \jacob_{\logpose_i} &
\schurrE_{ii} &=  \jacob_{\logpose_i}^\top \covar_{\residual_{ij}} \jacob_{\depth_i} \nonumber \\ 
\schurrB_{ij} &=  \jacob_{\logpose_i}^\top \covar_{\residual_{ij}} \jacob_{\logpose_j} &
\schurrE_{ji} &=  \jacob_{\logpose_j}^\top \covar_{\residual_{ij}} \jacob_{\depth_i} \nonumber \\
\schurrB_{ji} &=  \jacob_{\logpose_j}^\top \covar_{\residual_{ij}} \jacob_{\logpose_i} &
\mathbf{v}_i  &=  -\jacob_{\logpose_i}^\top \covar_{\residual_{ij}} \residual_{ij} \nonumber \\%
\schurrB_{jj} &=  \jacob_{\logpose_j}^\top \covar_{\residual_{ij}} \jacob_{\logpose_j} & 
\mathbf{v}_j  &=  -\jacob_{\logpose_j}^\top \covar_{\residual_{ij}} \residual_{ij} \nonumber \\
\schurrC_{i}  &=  \jacob_{\depth_i}^\top \covar_{\residual_{ij}} \jacob_{\depth_i} &
\mathbf{w}_i  &=  -\jacob_{\depth_i}^\top \covar_{\residual_{ij}} \residual_{ij} \nonumber \\
\label{eq:block_matrices}
\end{align}

We now consider again all edges $\edges$. Because the energy function in Eq.~\ref{eq:energy_function} is the sum of the energies for all edges, we can apply the sum rule for derivatives. Thus, we can combine the components of the normal equation. Since the block matrices from Eq.~\ref{eq:block_matrices} are dependent on their respective residual $\residual_{ij}$, we can combine the blocks via a scattered sum

\begin{equation}
\begin{split}
    \schurrB =& \sum_{(i, j) \in \edges}\matmap_{t_i t_i}^{T \times T}\left[ \schurrB_{ii}(\residual_{ij}) \right] + \matmap_{t_i t_j}^{T \times T}\left[ \schurrB_{ij}(\residual_{ij}) \right] \\
    & + \matmap_{t_j t_i}^{T \times T}\left[ \schurrB_{ji}(\residual_{ij}) \right] + \matmap_{t_j t_j}^{T \times T}\left[ \schurrB_{jj}(\residual_{ij}) \right] \\
    \schurrE =& \sum_{(i, j) \in \edges}{ \matmap_{t_i i}^{T \times |\nodes|} \left[ \schurrE_{ii}(\residual_{ij}) \right] + \matmap_{t_j i}^{T \times |\nodes|} \left[ \schurrE_{ji}(\residual_{ij}) \right] } \\
    \schurrC    =& \sum_{(i, j) \in \edges}{ \matmap_{ii}^{|\nodes| \times |\nodes|} \left[ \schurrC_{i}(\residual_{ij}) \right] } \\
    \mathbf{v}  =& \sum_{(i, j) \in \edges}{ \matmap_{t_i}^{T \times 1} \left[ \mathbf{v}_{i}(\residual_{ij}) \right] + \matmap_{t_j}^{T \times 1} \left[ \mathbf{v}_{j}(\residual_{ij}) \right]} \\
    \mathbf{w}  =& \sum_{(i, j) \in \edges}{ \matmap_{i}^{|\nodes| \times 1} \left[ \mathbf{w}_{i}(\residual_{ij}) \right]} \\
\end{split}
\end{equation}

where $\matmap_{mn}^{M \times N}: \real^{U \times V} \rightarrow \real^{M \cdot U \times N \cdot V}$ is the function which maps a block matrix to row $m \in \{1, ..., M\}$ and column $n \in \{1, ..., N\}$ while the other elements are set to zero.
To improve convergence, the Levenberg-Marquardt method is used on $\schurrC$, leading to

\begin{equation}
    \begin{bmatrix}
    \schurrB& \schurrE\\
    \schurrE^\top & \schurrC  + \damping \identity \\
    \end{bmatrix}
    \begin{bmatrix}
    \delta \logpose\\
    \delta \mathbf{d}\\
    \end{bmatrix} = 
    \begin{bmatrix}
    \mathbf{v}\\
    \mathbf{w}\\
    \end{bmatrix} 
\label{eq:gauss_newton_decomposed}
\end{equation}

With $\identity$ the identity matrix and pixel-wise damping factor $\damping$ is predicted by the GRU (see Sec.~\ref{sec:method}) for each node $i$.
We observe that $\schurrC$ is diagonal.
We can use the Schur complement to solve Eq.~\ref{eq:gauss_newton_decomposed} efficiently, due to $(\schurrC + \damping \identity)^{-1}$ being very easy to invert. Thus the updates are given by

\begin{equation}
\begin{split}
    \schurrS &= \schurrB - \schurrE \schurrC^{-1} \schurrE^\top \\
    \delta \logpose &=  \schurrS^{-1} (\mathbf{v} - \schurrE \schurrC^{-1} \mathbf{w}) \\
    \delta \depth &= (\schurrC + \damping \identity)^{-1} (\mathbf{w} - \schurrE^\top \delta \logpose) \\
\end{split}
\end{equation}

Lastly, it can be shown that $\schurrS \succ 0$, thus the Cholesky-decomposition can be used to efficiently solve for $\schurrS^{-1}$. 

\subsection{Jacobians}
Given the decomposition above, we now define the Jacobians $\jacob_{\logpose_i}$, $\jacob_{\logpose_j}$, and $\jacob_{\depth_i}$. Let us consider a single depth $d$ of node $i$ at location $(u, v)$. The residual is given by

\begin{equation}
    \residual_{ij, uv} = \coords_{ij, uv} - \hat{\coords}_{ij, uv} \in \real^{2}
\end{equation}

where $\hat{\coords}_{ij, uv} = \Pi_{c_j}(\pose_{ij} \circ \Pi_{c_i}^{-1}(d))$. We further define the 3D point corresponding to pixel $(u, v)$ as
$\point = \begin{bmatrix}
    X & Y & Z & W
\end{bmatrix}^\top$ 
which is given by $\point = \Pi_{c_i}^{-1}(d)$ and the transformed point $\point' = \pose_{ij} \point$. We can thus define the projection and unprojection operators for pinhole cameras $c_i$ and $c_j$

\begin{equation}
\begin{split}
    \Pi_{c_j}(\point') = 
    \begin{bmatrix}
    f_{c_j}^x \frac{X'}{Z'} +c_{c_j}^x \\
    f_{c_j}^y \frac{Y'}{Z'} +c_{c_j}^y \\
    \end{bmatrix}
\end{split}
\end{equation}

\begin{equation}
\begin{split}
    \Pi_{c_i}^{-1}(d) = 
    \begin{bmatrix}
    \frac{u - c_{c_i}^x}{f_{c_i}^x} \\
    \frac{v - c_{c_i}^y}{f_{c_i}^y}  \\
    1 \\
    d \\
    \end{bmatrix}
\end{split}
\end{equation}

where $f_c^x, f_c^y$ are the camera $c$'s focal lengths in $x$ and $y$ direction, $c_{c}^x, c_{c}^y$ are the respective principal points.

\parsection{Depth}
The Jacobian $\jacob_d$ w.r.t. depth $d$ is defined as
\begin{equation}
\begin{split}
    \jacob_{d} &= \frac{\partial \residual_{ij}}{\partial d} = -\frac{\partial \hat{\coords}_{ij}}{\partial d} = -\frac{\partial \Pi_{c_j}(\point')}{\partial \point'} \frac{\partial \point'}{\partial d} \\
    & = -\frac{\partial \Pi_{c_j}(\point')}{\partial \point'} \pose_{ij} \frac{\partial \Pi_{c_i}^{-1}(d)}{\partial d}
\end{split}
\end{equation}

\begin{equation}
\begin{split}
    \frac{\partial \Pi_{c_j}(\point')}{\partial \point'} = 
    \begin{bmatrix}
    f_{c_j}^x \frac{1}{Z'} & 0                      & - f_{c_j}^x \frac{X'}{Z'^2} & 0 \\
    0                      & f_{c_j}^y \frac{1}{Z'} & - f_{c_j}^y \frac{Y'}{Z'^2} & 0 \\
    \end{bmatrix}
\end{split}
\label{eq:proj_derivative}
\end{equation}

\begin{equation}
\begin{split}
    \frac{\partial \Pi_{c_i}^{-1}(d)}{\partial d} = 
    \begin{bmatrix}
    0 \\
    0 \\
    0 \\
    1 \\
    \end{bmatrix}
\end{split}
\end{equation}

\parsection{Pose} The Jacobian $\jacob_{\logpose}$ w.r.t. $\logpose$ where $\logpose \in \se{3}$ is either $\logpose_i$ or $\logpose_j$.

\begin{equation}
    \jacob_{\logpose} = \frac{\partial \residual_{ij}}{\partial \logpose} = -\frac{\partial \hat{\coords}_{ij}}{\partial \logpose} = -\frac{\partial \Pi_{c_j}(\point')}{\partial \point'} \frac{\partial \point'}{\partial \logpose}
\end{equation}

The partial derivative $\frac{\partial \Pi_{c_j}(\point')}{\partial \point'}$ has been derived in Eq.~\ref{eq:proj_derivative}. For $\frac{\partial \pose_{ij}}{\partial \logpose}$, we can again decompose $\pose_{ij}$ into the static parts $\extrinsics_{c_i}$ and $\extrinsics_{c_j}$ and the unknown, to be optimized parts, $\egopose_{t_i}$ and $\egopose_{t_j}$

\begin{equation}
\point' = (\egopose_{t_j} \extrinsics_{c_j})^{-1} \egopose_{c_i} \extrinsics_{c_i} \point
\end{equation}

First, similar to Eq.~39 to~44 in~\cite{eade2013lie}, for $\expose = \expose_1 (\expose_0)^{-1}$ we can write

\begin{equation}
\begin{split}
    \frac{\partial \expose}{\partial \expose_0} &=  \frac{\partial \log{\left( \expose_1 (\exp{(\logpose)} \expose_0)^{-1} (\expose_1 \expose_0^{-1})^{-1} \right)}}{\partial \logpose}|_{\logpose = 0} \\
    &=  \frac{\partial}{\partial \logpose}|_{\logpose = 0} \left[ \log{\left( \expose_1 \expose_0^{-1} \exp{(-\logpose)} \expose_0 \expose_1^{-1} \right)} \right] \\
    &=  \frac{\partial}{\partial \logpose}|_{\logpose = 0} \left[ \log{\left(\exp{(-\operatorname{Adj}_{\expose_1 \expose_0^{-1}} \logpose)} \expose_1 \expose_0^{-1} \expose_0 \expose_1^{-1} \right)} \right] \\
    &=  \frac{\partial}{\partial \logpose}|_{\logpose = 0} \left[ \log{\left(\exp{(-\operatorname{Adj}_{\expose_1 \expose_0^{-1}} \logpose)} \right)} \right] \\
    &=  \frac{\partial}{\partial \logpose}|_{\logpose = 0} \left[ -\operatorname{Adj}_{\expose_1 \expose_0^{-1}} \logpose \right] \\
    &= -\operatorname{Adj}_{\expose_{1} \expose_{0}^{-1}}
\end{split}
\label{eq:inverse_derivative}
\end{equation}

In the following, we omit the explicit perturbation around $\logpose = 0$. Let now $\generator_1,..., \generator_6 \in \real^{4 \times 4}$ be generators as defined in Eq.~65 in~\cite{eade2013lie}. With the chain rule and Eq.~94 in~\cite{eade2013lie} and Eq.~\ref{eq:inverse_derivative} above, the derivative w.r.t. the pose of incoming node $j$ is given by

\begin{equation}
\begin{split}
    \frac{\partial \point'}{\partial \logpose_j} &= \frac{\partial}{\partial \logpose_j} \left[ (\exp{(\logpose_j)} \egopose_{t_j} \extrinsics_{c_j})^{-1} \egopose_{t_i} \extrinsics_{c_i} \point \right] \\
    &= \frac{\partial}{\partial \logpose_j} [ \underbrace{\extrinsics_{c_j}^{-1} (\exp{(\logpose_j)} \egopose_{t_j})^{-1}}_{\expose} \underbrace{\egopose_{t_i} \extrinsics_{c_i} \point}_{\point^w} ] \\
    &= \frac{\partial}{\partial \expose} \left[ \expose \point^w \right] \cdot \frac{\partial}{\partial \logpose_j}[\underbrace{\extrinsics_{c_j}^{-1}}_{\expose_1} (\underbrace{\exp{(\logpose_j)} \egopose_{t_j}}_{\expose_0})^{-1} ] \\
    &= \frac{\partial}{\partial \expose} [\expose \point^w] \cdot \frac{\partial}{\partial \expose_0} [\expose_1 \expose_0^{-1}] \\
    &= \begin{bmatrix} \generator_1 \point^w & ... & \generator_6 \point^w \end{bmatrix} \cdot (-\operatorname{Adj}_{\expose_1 \expose_0^{-1}}) \\
    &= -\begin{bmatrix} \generator_1 \point^w & ... & \generator_6 \point^w \end{bmatrix} \cdot \operatorname{Adj}_{\extrinsics_{c_j}^{-1} \egopose_{t_j}^{-1}}
\end{split}
\end{equation}

Finally, with Eq.~94 and~97 in~\cite{eade2013lie} and the derivative w.r.t. the pose of the outgoing node $i$

\begin{equation}
\begin{split}
    \frac{\partial \point'}{\partial \logpose_i} &= \frac{\partial}{\partial \logpose_i} [\underbrace{(\egopose_{t_j} \extrinsics_{c_j})^{-1} \exp{(\logpose_i)} \egopose_{t_i}}_{\expose} \underbrace{\extrinsics_{c_i} \point}_{\point^{t_i}}] \\
    &= \frac{\partial}{\partial \expose} [\expose \point^{t_i}] \cdot \frac{\partial}{\partial \logpose_i} [(\egopose_{t_j} \extrinsics_{c_j})^{-1} \exp{(\logpose_i)} \egopose_{t_i}]\\
    &= \begin{bmatrix} \generator_1 \point^{t_i} & ... & \generator_6 \point^{t_i} \end{bmatrix} \cdot \operatorname{Adj}_{\extrinsics_{c_j}^{-1} \egopose_{t_j}^{-1}}
\end{split}
\end{equation}

Finally, we compose the component-wise Jacobians above into the Jacobian $\jacob$ as stated in Sec.~\ref{sec:dba_decomposition}. We can now use $\jacob$ to compute the updates $\delta\xi$ and $\delta d$ in Eq.~\ref{eq:gauss_newton}.

\def \imgpath {fig_qualitative_ddad_000163_img_000080}
\def \absrelpath {fig_qualitative_ddad_000163_abs_rel_000080}
\def \depthpath {fig_qualitative_ddad_000163_depth_000080}

\def \imheight {47pt}

\def \cropl {0px}
\def \cropb {66px}
\def \cropr {30px}
\def \cropt {48px}

\begin{figure*}
\centering
\setlength\tabcolsep{0.5 pt}

\begin{tabular}{lcccccc}
& Front & F.Left & F.Right & B.Left & B.Right & Back \\

\rotatebox[origin=c]{90}{Input} &
\begin{tabular}{l}\includegraphics[height=\imheight]{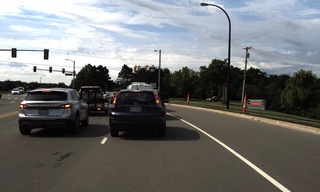}\end{tabular} &
\begin{tabular}{l}\includegraphics[height=\imheight]{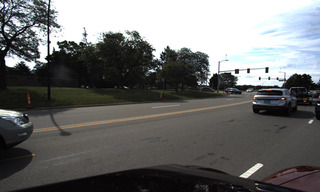}\end{tabular} &
\begin{tabular}{l}\includegraphics[height=\imheight]{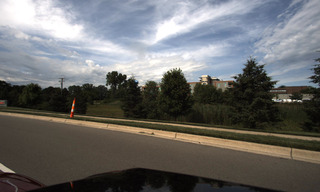}\end{tabular} &
\begin{tabular}{l}\includegraphics[height=\imheight]{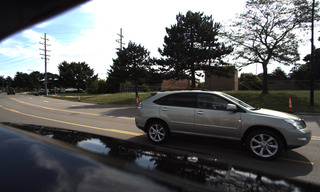}\end{tabular} &
\begin{tabular}{l}\includegraphics[height=\imheight]{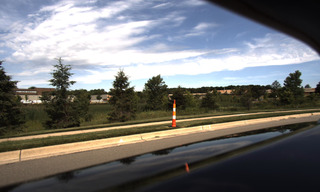}\end{tabular} &
\begin{tabular}{l}\includegraphics[height=\imheight]{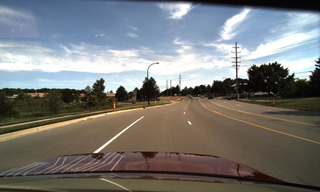}\end{tabular} \\ \midrule

\multirow{2}{*}[-1em]{\rotatebox[origin=c]{90}{FSM~\cite{FSM}}} &
\begin{tabular}{l}\includegraphics[height=\imheight]{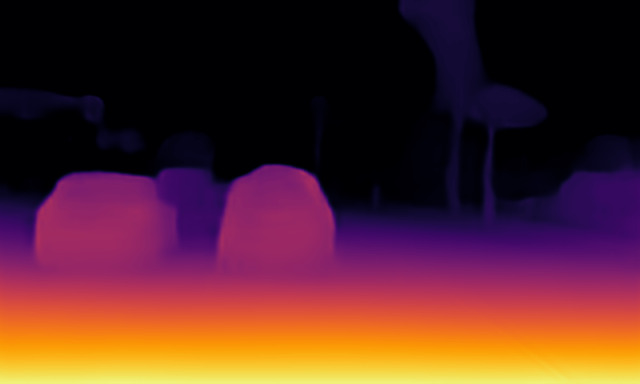}\end{tabular} &
\begin{tabular}{l}\includegraphics[height=\imheight]{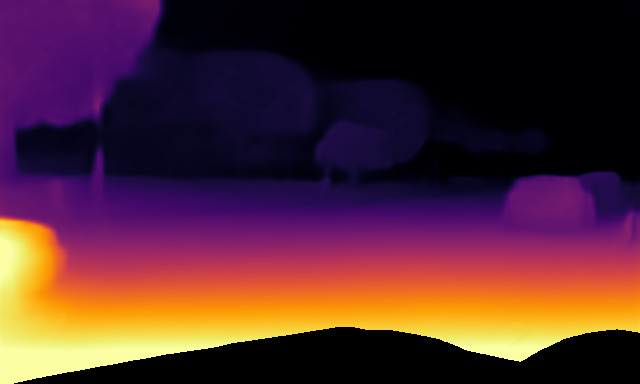}\end{tabular} &
\begin{tabular}{l}\includegraphics[height=\imheight]{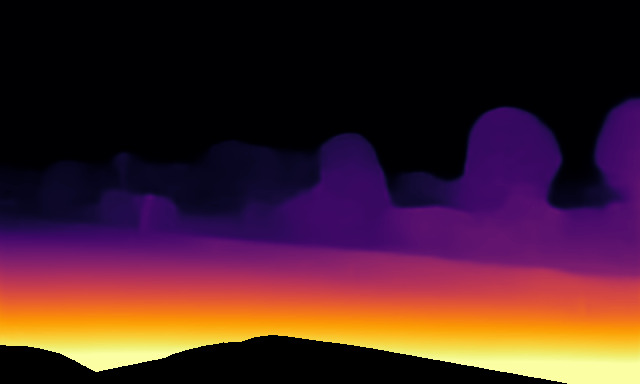}\end{tabular} &
\begin{tabular}{l}\includegraphics[height=\imheight]{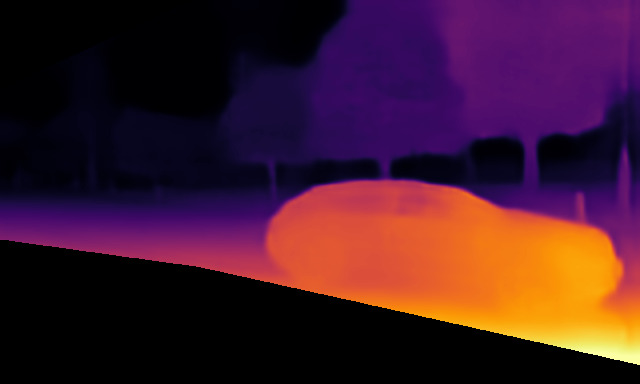}\end{tabular} &
\begin{tabular}{l}\includegraphics[height=\imheight]{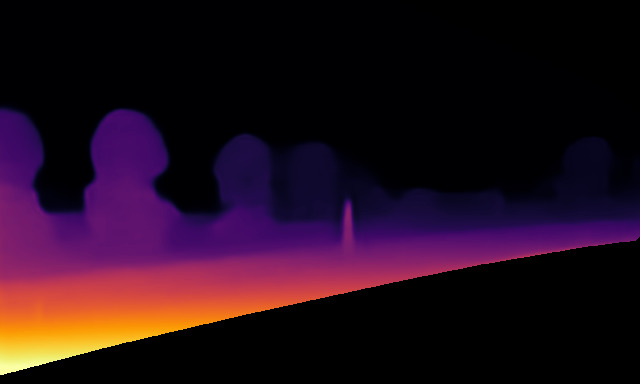}\end{tabular} &
\begin{tabular}{l}\includegraphics[height=\imheight]{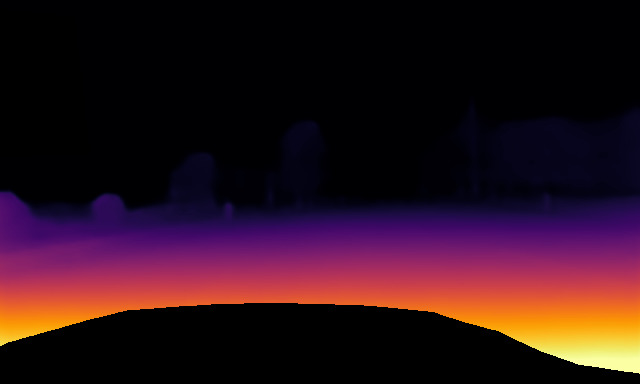}\end{tabular} \\
&
\begin{tabular}{l}\includegraphics[height=\imheight]{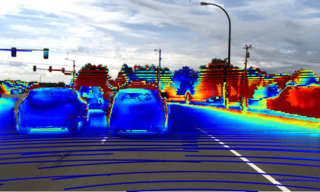}\end{tabular} &
\begin{tabular}{l}\includegraphics[height=\imheight]{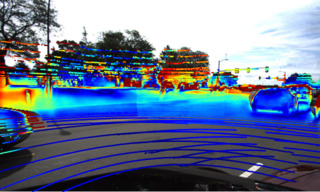}\end{tabular} &
\begin{tabular}{l}\includegraphics[height=\imheight]{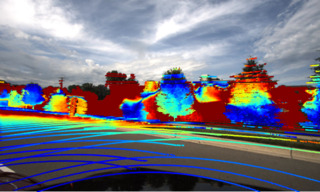}\end{tabular} &
\begin{tabular}{l}\includegraphics[height=\imheight]{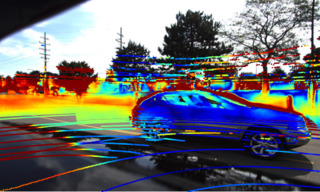}\end{tabular} &
\begin{tabular}{l}\includegraphics[height=\imheight]{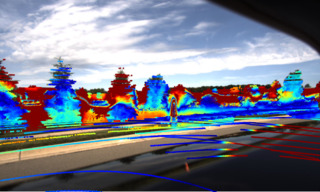}\end{tabular} &
\begin{tabular}{l}\includegraphics[height=\imheight]{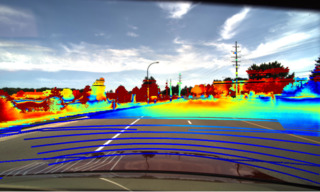}\end{tabular} \\ \midrule

\multirow{2}{*}[1em]{\rotatebox[origin=c]{90}{SurroundDepth~\cite{SurroundDepth}}} &
\begin{tabular}{l}\includegraphics[height=\imheight]{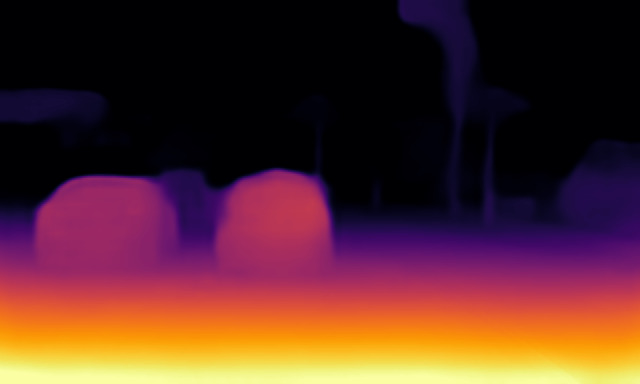}\end{tabular} &
\begin{tabular}{l}\includegraphics[height=\imheight]{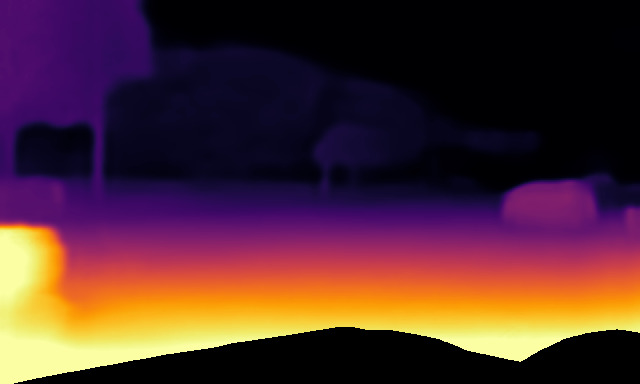}\end{tabular} &
\begin{tabular}{l}\includegraphics[height=\imheight]{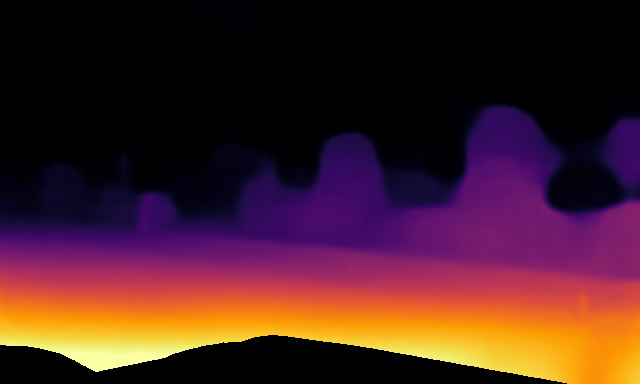}\end{tabular} &
\begin{tabular}{l}\includegraphics[height=\imheight]{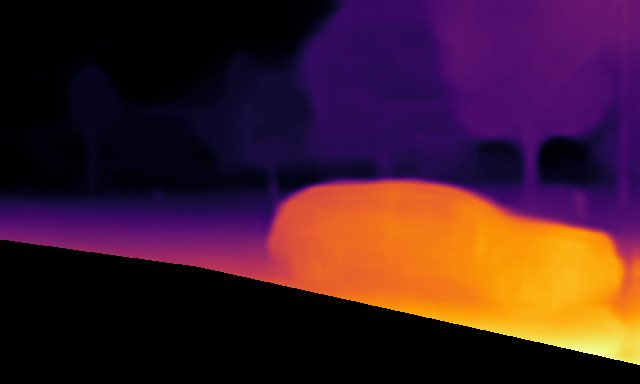}\end{tabular} &
\begin{tabular}{l}\includegraphics[height=\imheight]{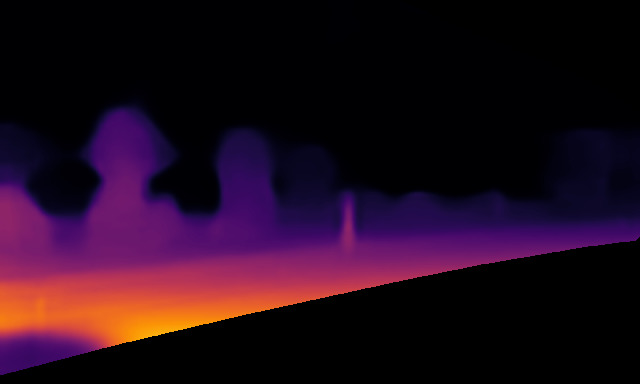}\end{tabular} &
\begin{tabular}{l}\includegraphics[height=\imheight]{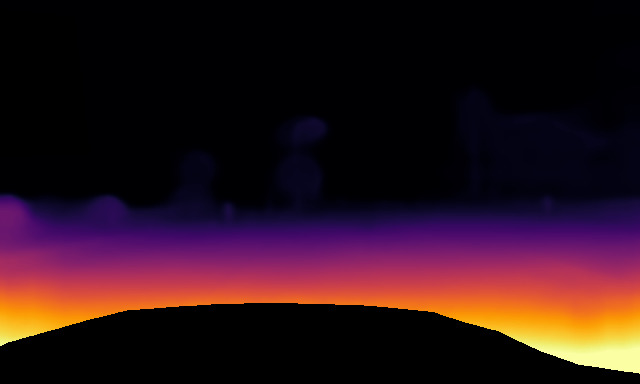}\end{tabular} \\
&
\begin{tabular}{l}\includegraphics[height=\imheight]{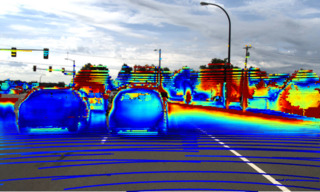}\end{tabular} &
\begin{tabular}{l}\includegraphics[height=\imheight]{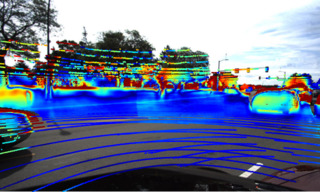}\end{tabular} &
\begin{tabular}{l}\includegraphics[height=\imheight]{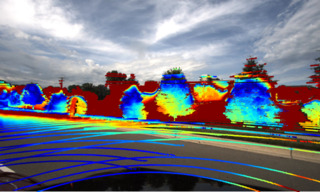}\end{tabular} &
\begin{tabular}{l}\includegraphics[height=\imheight]{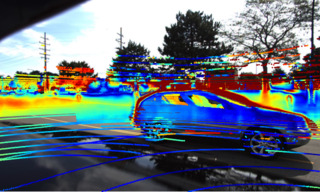}\end{tabular} &
\begin{tabular}{l}\includegraphics[height=\imheight]{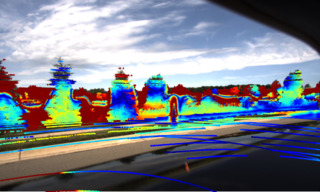}\end{tabular} &
\begin{tabular}{l}\includegraphics[height=\imheight]{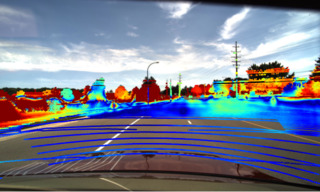}\end{tabular} \\ \midrule

\multirow{2}{*}[0.5em]{\rotatebox[origin=c]{90}{Geometric Depth}} &
\begin{tabular}{l}\includegraphics[height=\imheight]{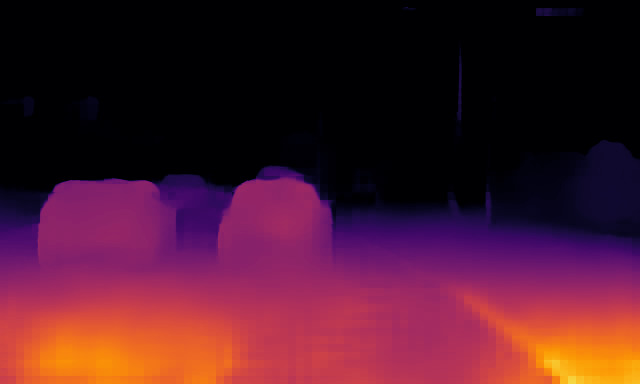}\end{tabular} &
\begin{tabular}{l}\includegraphics[height=\imheight]{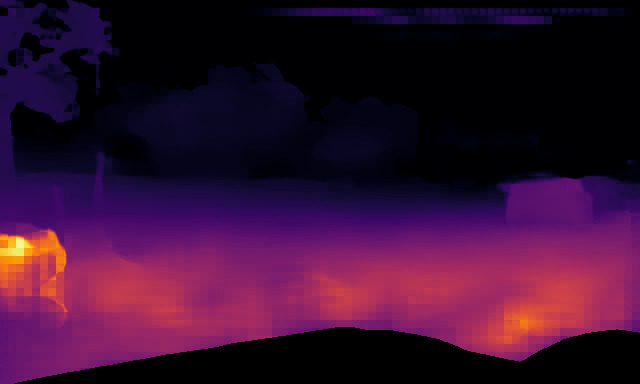}\end{tabular} &
\begin{tabular}{l}\includegraphics[height=\imheight]{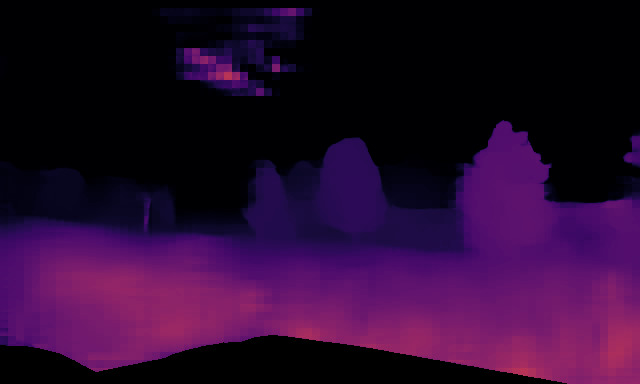}\end{tabular} &
\begin{tabular}{l}\includegraphics[height=\imheight]{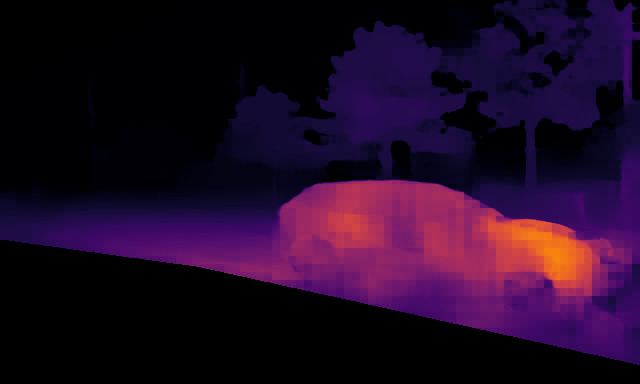}\end{tabular} &
\begin{tabular}{l}\includegraphics[height=\imheight]{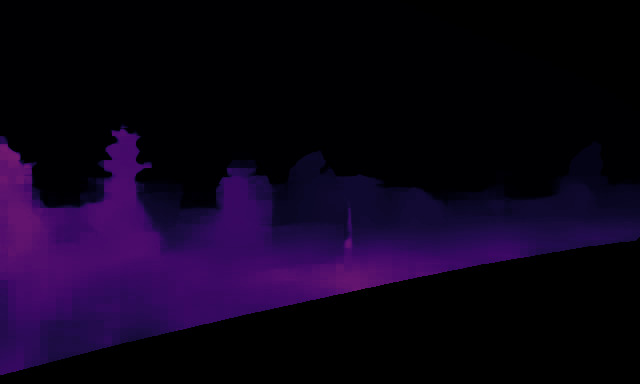}\end{tabular} &
\begin{tabular}{l}\includegraphics[height=\imheight]{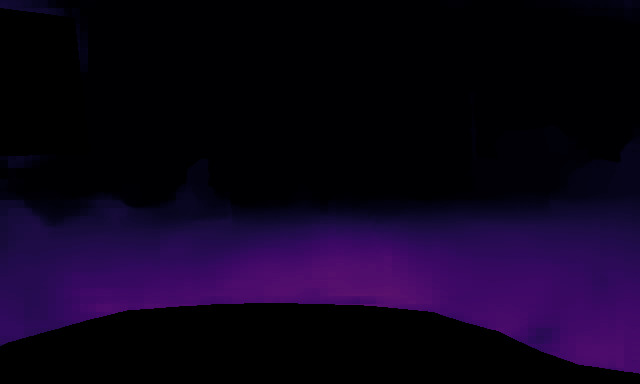}\end{tabular} \\
&
\begin{tabular}{l}\includegraphics[height=\imheight]{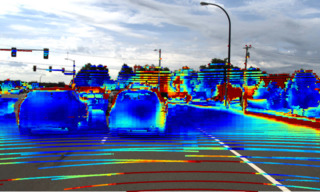}\end{tabular} &
\begin{tabular}{l}\includegraphics[height=\imheight]{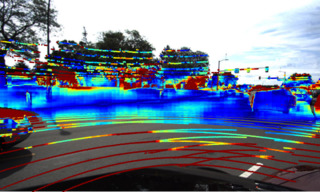}\end{tabular} &
\begin{tabular}{l}\includegraphics[height=\imheight]{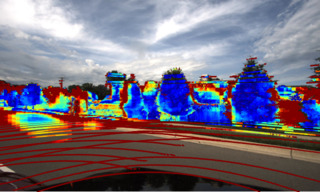}\end{tabular} &
\begin{tabular}{l}\includegraphics[height=\imheight]{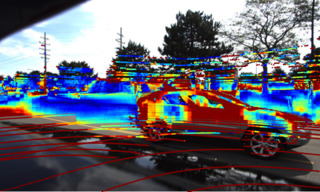}\end{tabular} &
\begin{tabular}{l}\includegraphics[height=\imheight]{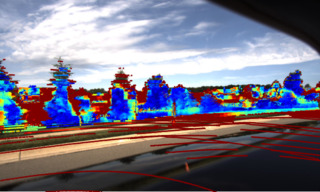}\end{tabular} &
\begin{tabular}{l}\includegraphics[height=\imheight]{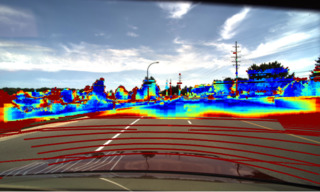}\end{tabular} \\ \midrule

\multirow{2}{*}[-1.8em]{\rotatebox[origin=c]{90}{Ours}} &
\begin{tabular}{l}\includegraphics[height=\imheight]{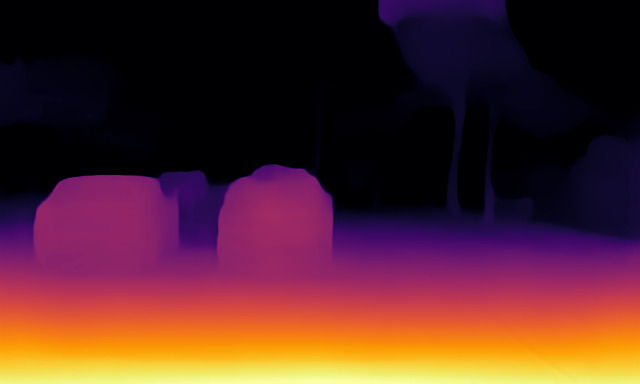}\end{tabular} &
\begin{tabular}{l}\includegraphics[height=\imheight]{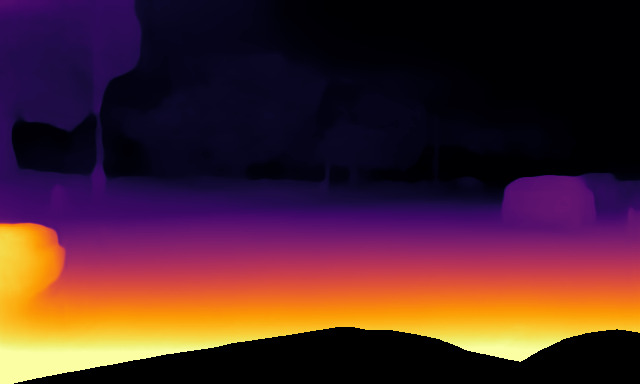}\end{tabular} &
\begin{tabular}{l}\includegraphics[height=\imheight]{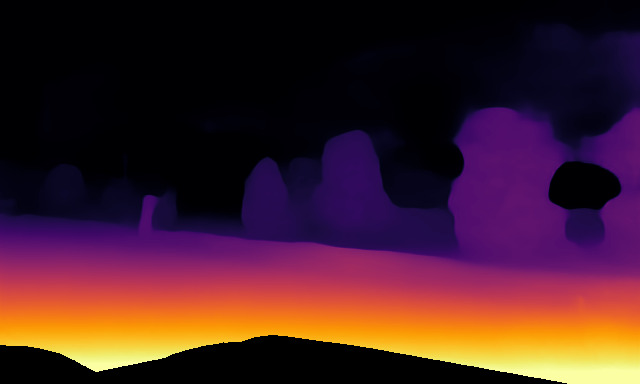}\end{tabular} &
\begin{tabular}{l}\includegraphics[height=\imheight]{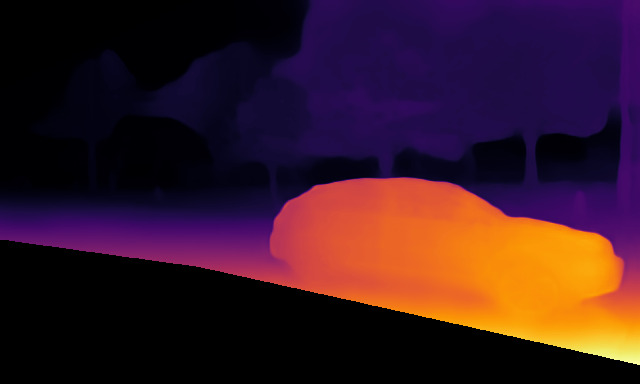}\end{tabular} &
\begin{tabular}{l}\includegraphics[height=\imheight]{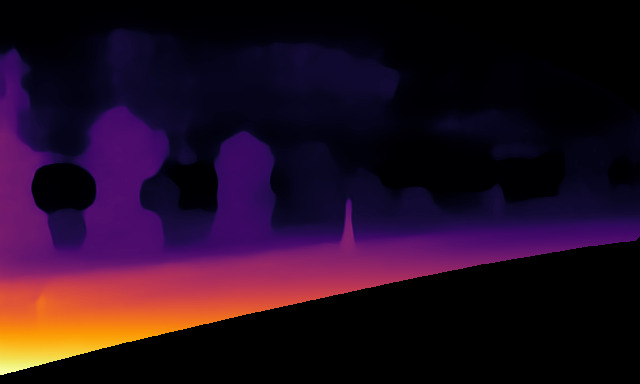}\end{tabular} &
\begin{tabular}{l}\includegraphics[height=\imheight]{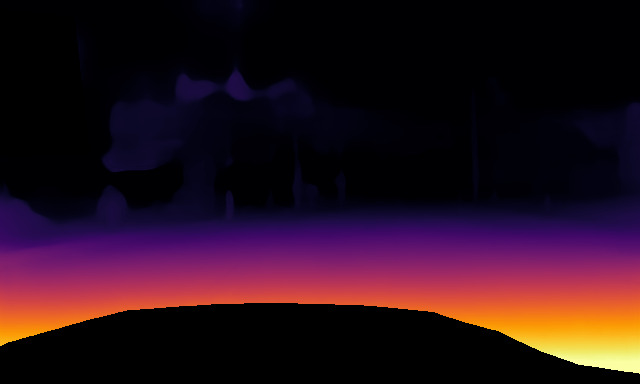}\end{tabular} \\
&
\begin{tabular}{l}\includegraphics[height=\imheight]{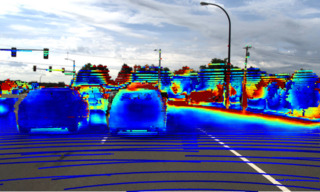}\end{tabular} &
\begin{tabular}{l}\includegraphics[height=\imheight]{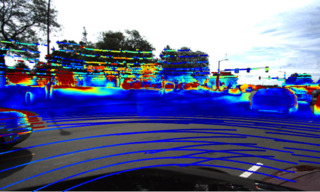}\end{tabular} &
\begin{tabular}{l}\includegraphics[height=\imheight]{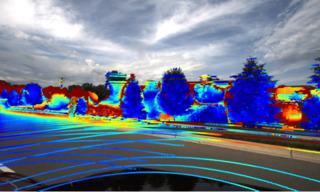}\end{tabular} &
\begin{tabular}{l}\includegraphics[height=\imheight]{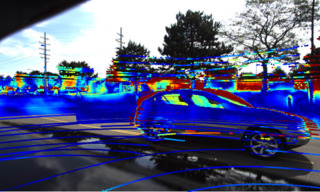}\end{tabular} &
\begin{tabular}{l}\includegraphics[height=\imheight]{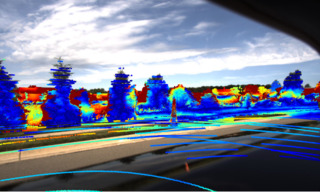}\end{tabular} &
\begin{tabular}{l}\includegraphics[height=\imheight]{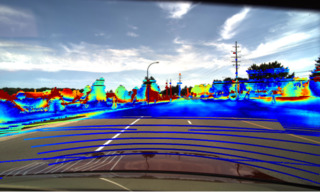}\end{tabular} \\ \midrule
&
\multicolumn{3}{c}{\begin{tabular}{l}\includegraphics[width=.41\linewidth]{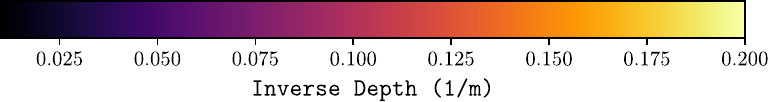}\end{tabular}} &
\multicolumn{3}{c}{\begin{tabular}{l}\includegraphics[width=.45\linewidth]{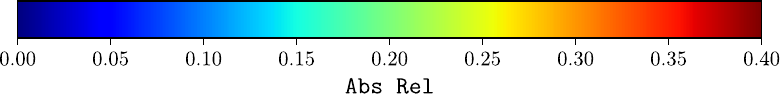}\end{tabular}} \\

\end{tabular}

\caption{\textbf{Qualitative comparison on DDAD.} We compare existing works to both our geometric and refined depth estimates. Especially for the side views, existing works struggle to obtain accurate depth. The geometric depth produces many accurate depth predictions, but contains many noisy points, especially in low-textured areas and for dynamic objects. Our full method demonstrates the best performance.}
\label{fig:qualitative_ddad_000163}
\end{figure*}
\def \imgpathi {fig_qualitative_nuscenes_0103_img_000072}
\def \depthpathi {fig_qualitative_nuscenes_0103_depth_000072}

\def \imgpathii {fig_qualitative_nuscenes_0268_img_000043}
\def \depthpathii {fig_qualitative_nuscenes_0268_depth_000043}

\def \imgpathiii {fig_qualitative_nuscenes_0274_img_000214}
\def \depthpathiii {fig_qualitative_nuscenes_0274_depth_000214}

\begin{figure*}[t]
\setlength\tabcolsep{0.5 pt}
\centering

\begin{tabular}{lcccccc}

& Front & F.Left & F.Right & B.Left & B.Right & Back \\
\rotatebox[origin=c]{90}{Input} &
\begin{tabular}{l}\includegraphics[width=0.15\linewidth]{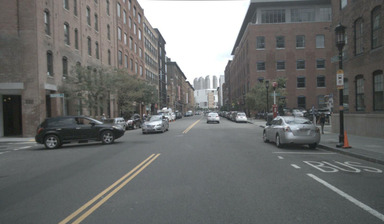}\end{tabular} &
\begin{tabular}{l}\includegraphics[width=0.15\linewidth]{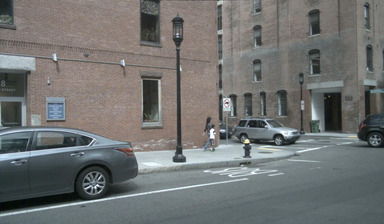}\end{tabular} &
\begin{tabular}{l}\includegraphics[width=0.15\linewidth]{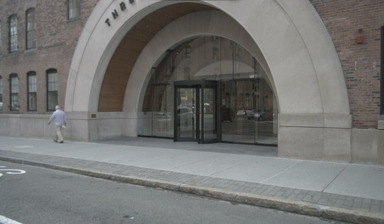}\end{tabular} &
\begin{tabular}{l}\includegraphics[width=0.15\linewidth]{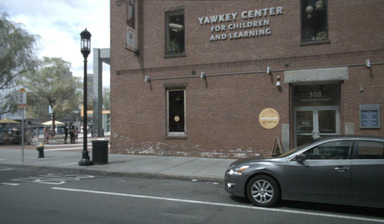}\end{tabular} &
\begin{tabular}{l}\includegraphics[width=0.15\linewidth]{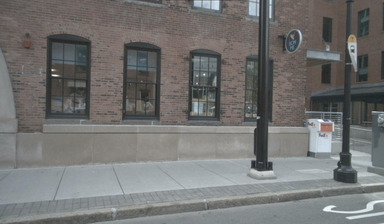}\end{tabular} &
\begin{tabular}{l}\includegraphics[width=0.15\linewidth]{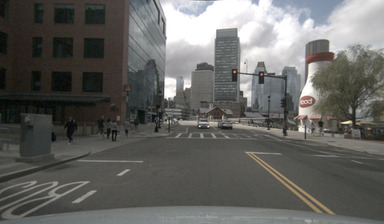}\end{tabular} \\

\rotatebox[origin=c]{90}{SD~\cite{SurroundDepth}} &
\begin{tabular}{l}\includegraphics[width=0.15\linewidth]{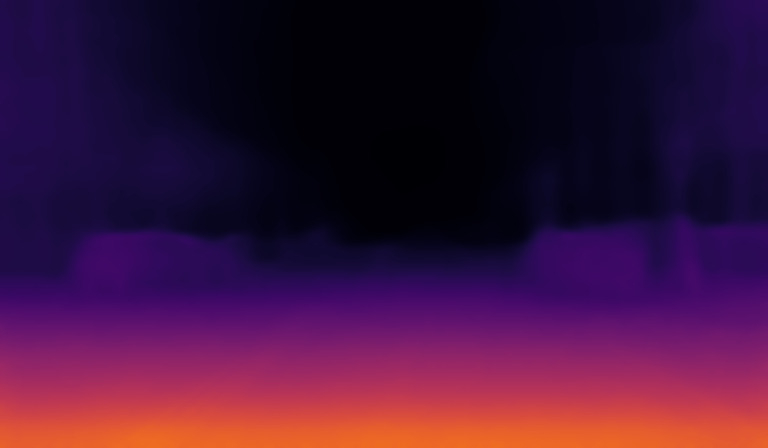}\end{tabular} &
\begin{tabular}{l}\includegraphics[width=0.15\linewidth]{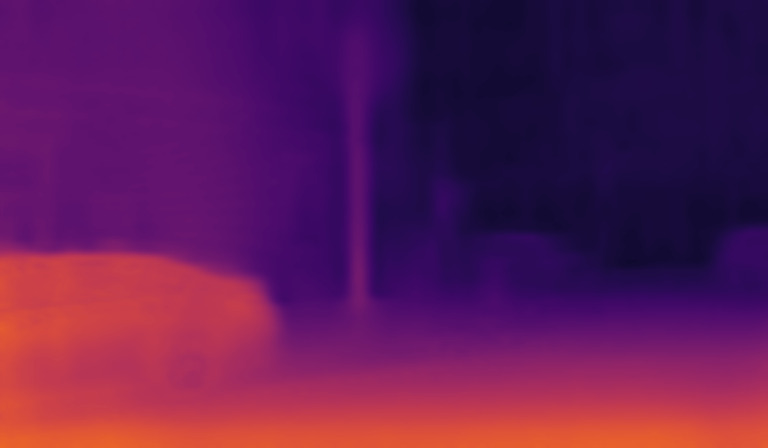}\end{tabular} &
\begin{tabular}{l}\includegraphics[width=0.15\linewidth]{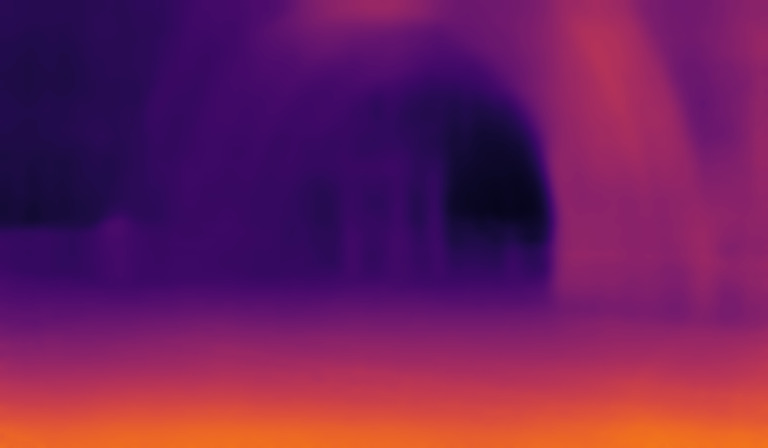}\end{tabular} &
\begin{tabular}{l}\includegraphics[width=0.15\linewidth]{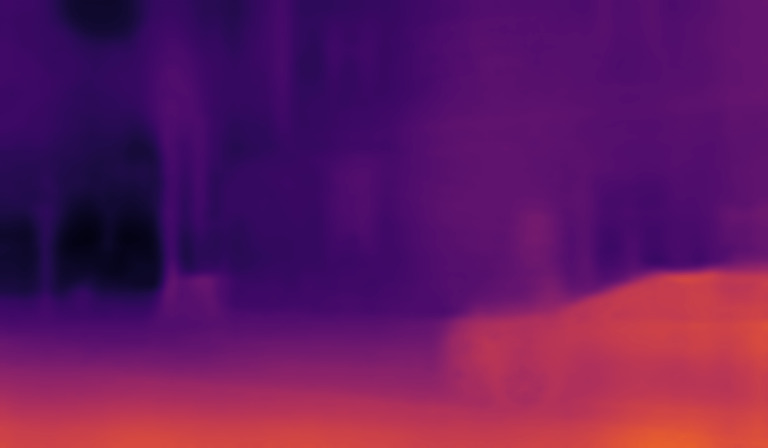}\end{tabular} &
\begin{tabular}{l}\includegraphics[width=0.15\linewidth]{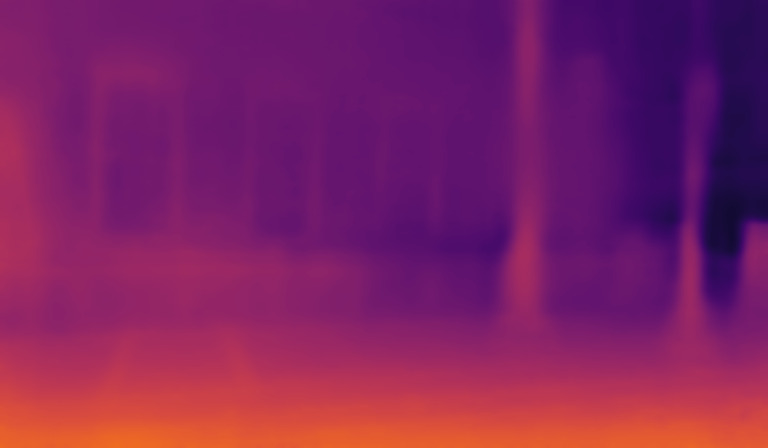}\end{tabular} &
\begin{tabular}{l}\includegraphics[width=0.15\linewidth]{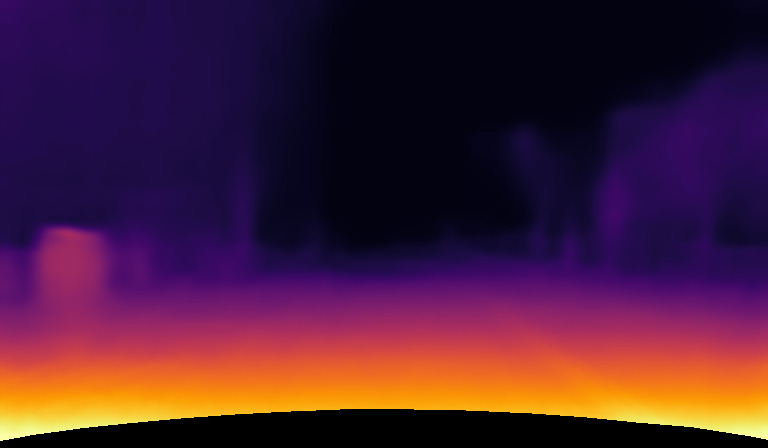}\end{tabular} \\

\rotatebox[origin=c]{90}{Ours} &
\begin{tabular}{l}\includegraphics[width=0.15\linewidth]{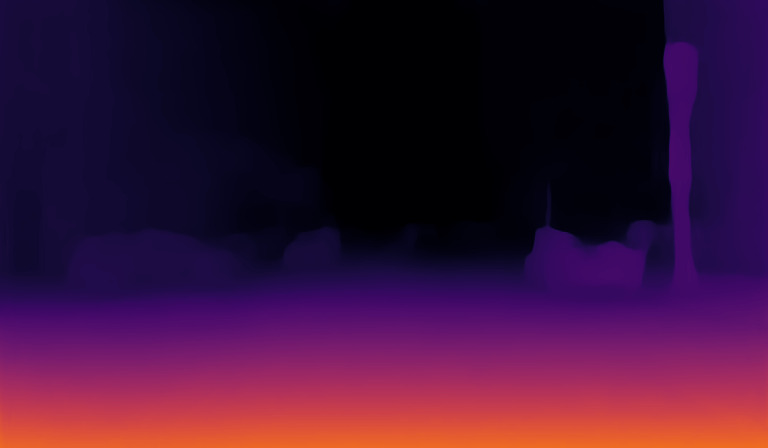}\end{tabular} &
\begin{tabular}{l}\includegraphics[width=0.15\linewidth]{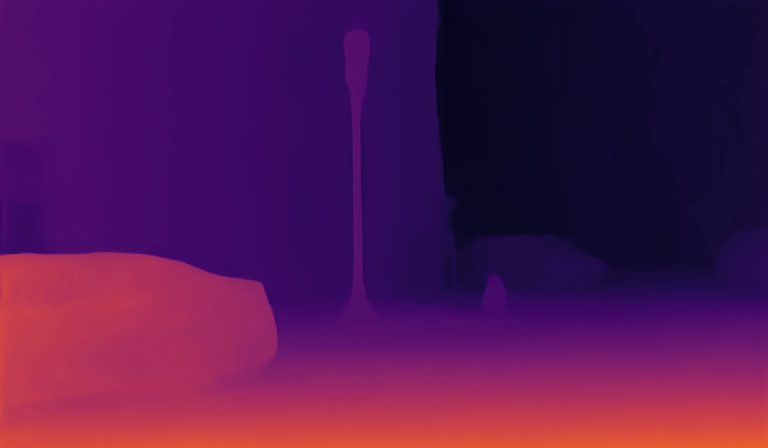}\end{tabular} &
\begin{tabular}{l}\includegraphics[width=0.15\linewidth]{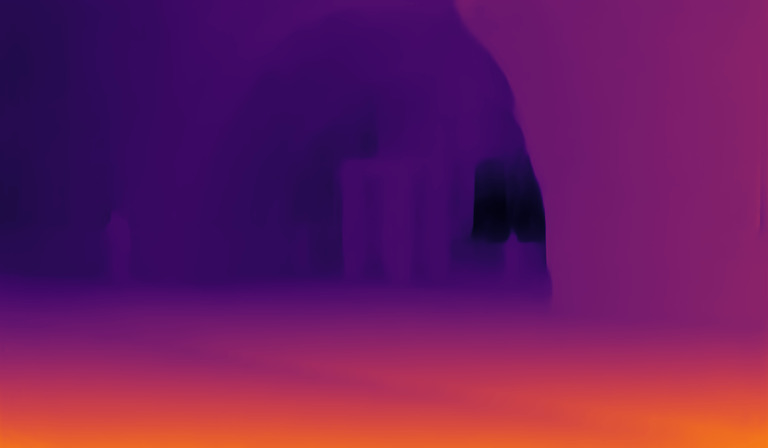}\end{tabular} &
\begin{tabular}{l}\includegraphics[width=0.15\linewidth]{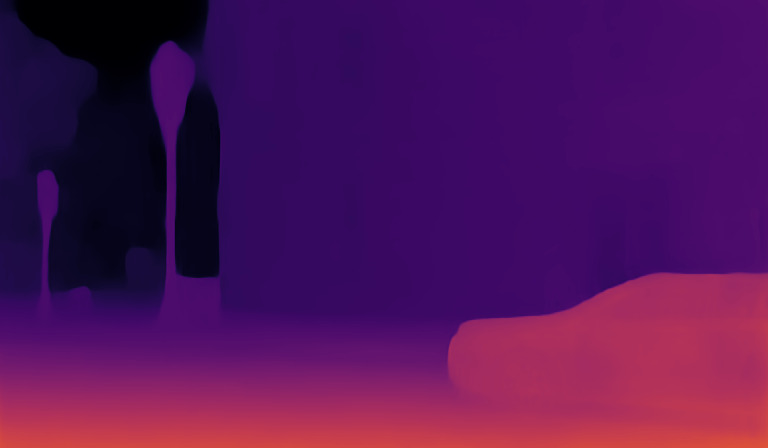}\end{tabular} &
\begin{tabular}{l}\includegraphics[width=0.15\linewidth]{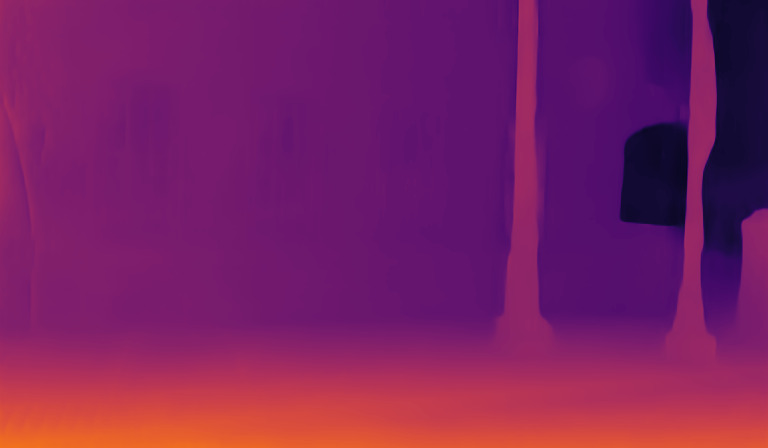}\end{tabular} &
\begin{tabular}{l}\includegraphics[width=0.15\linewidth]{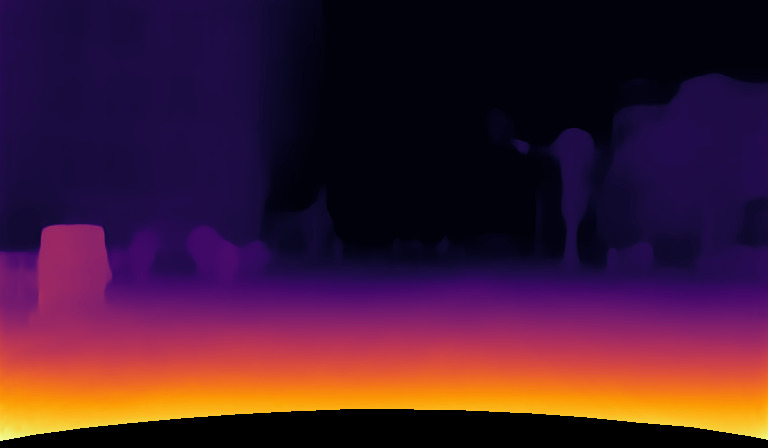}\end{tabular} \\\midrule

\rotatebox[origin=c]{90}{Input} &
\begin{tabular}{l}\includegraphics[width=0.15\linewidth]{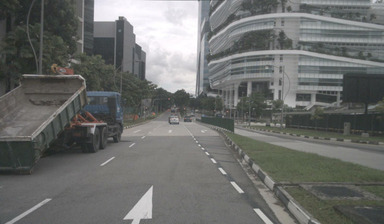}\end{tabular} &
\begin{tabular}{l}\includegraphics[width=0.15\linewidth]{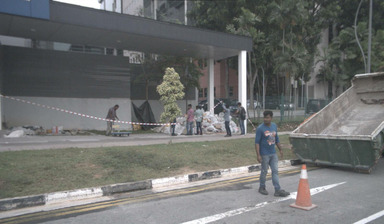}\end{tabular} &
\begin{tabular}{l}\includegraphics[width=0.15\linewidth]{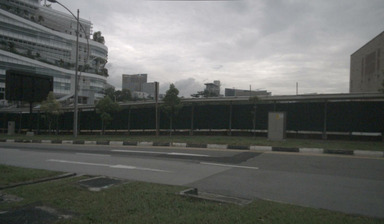}\end{tabular} &
\begin{tabular}{l}\includegraphics[width=0.15\linewidth]{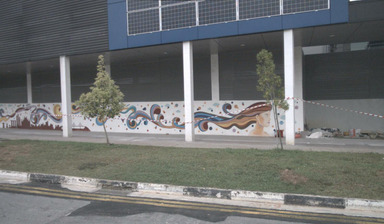}\end{tabular} &
\begin{tabular}{l}\includegraphics[width=0.15\linewidth]{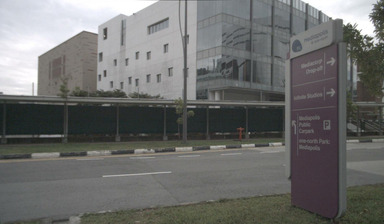}\end{tabular} &
\begin{tabular}{l}\includegraphics[width=0.15\linewidth]{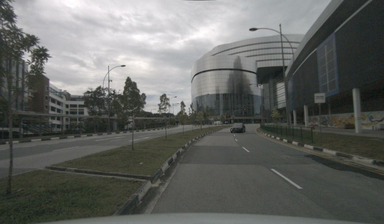}\end{tabular} \\

\rotatebox[origin=c]{90}{SD~\cite{SurroundDepth}} &
\begin{tabular}{l}\includegraphics[width=0.15\linewidth]{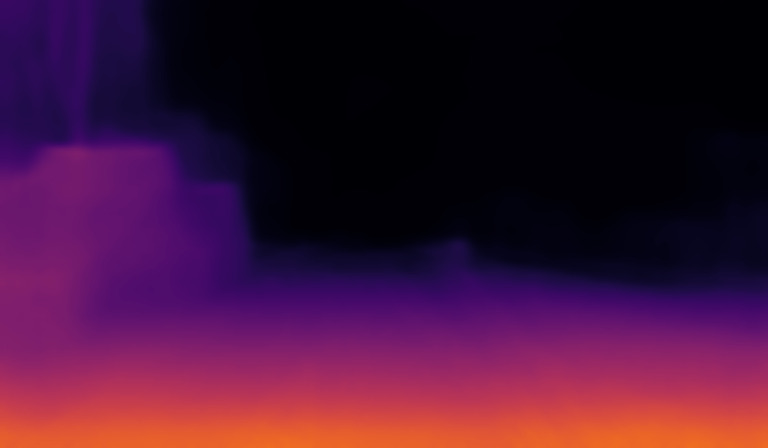}\end{tabular} &
\begin{tabular}{l}\includegraphics[width=0.15\linewidth]{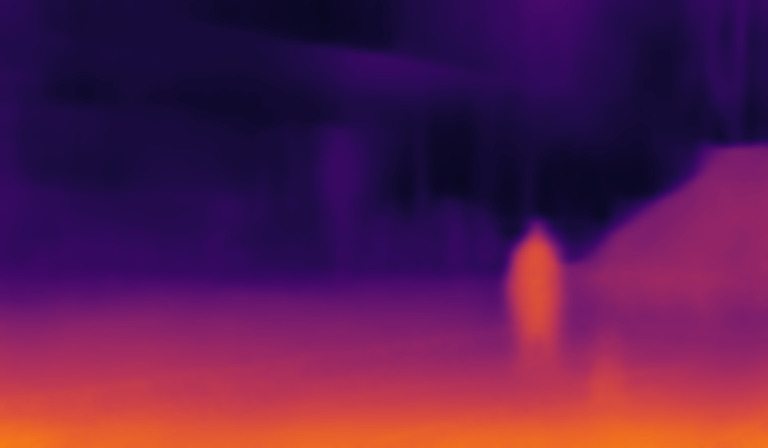}\end{tabular} &
\begin{tabular}{l}\includegraphics[width=0.15\linewidth]{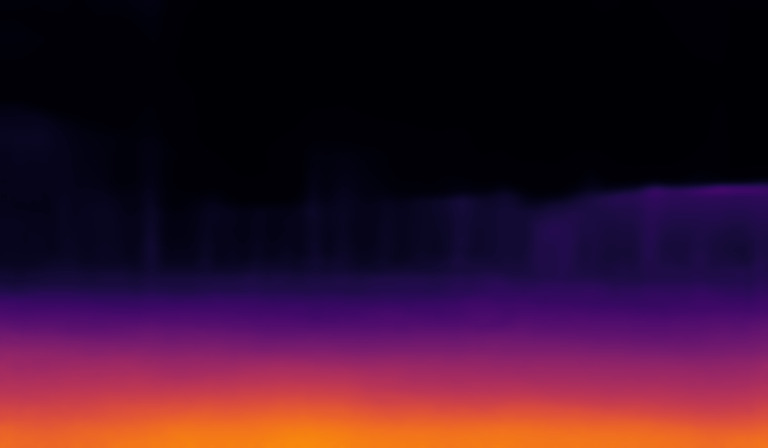}\end{tabular} &
\begin{tabular}{l}\includegraphics[width=0.15\linewidth]{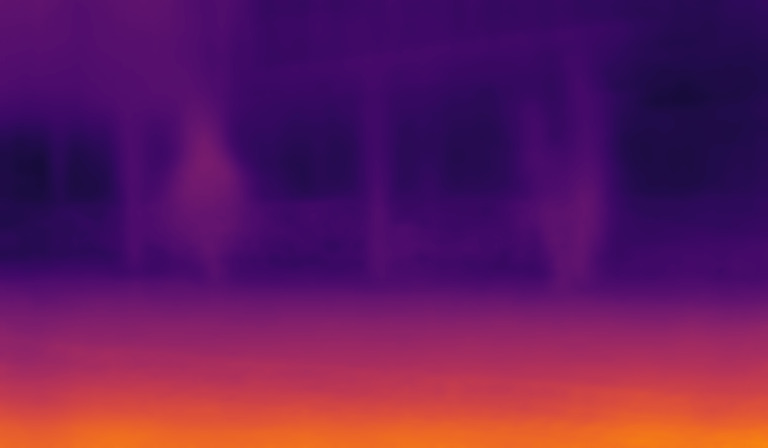}\end{tabular} &
\begin{tabular}{l}\includegraphics[width=0.15\linewidth]{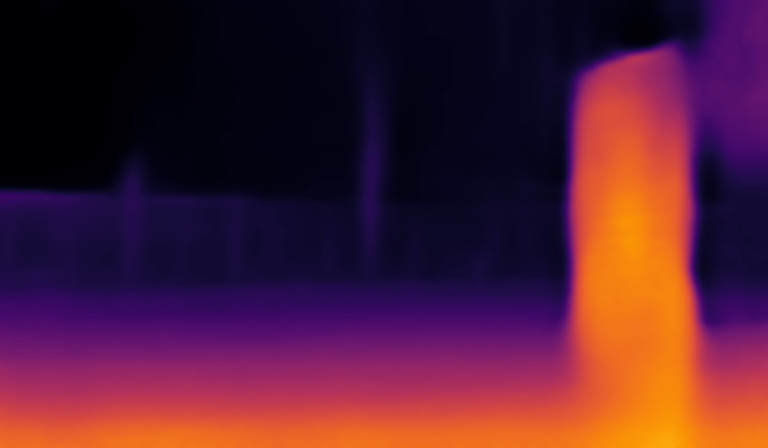}\end{tabular} &
\begin{tabular}{l}\includegraphics[width=0.15\linewidth]{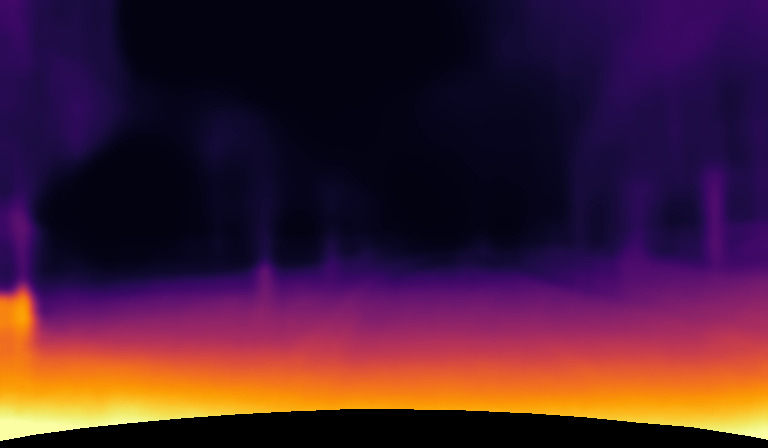}\end{tabular} \\

\rotatebox[origin=c]{90}{Ours} &
\begin{tabular}{l}\includegraphics[width=0.15\linewidth]{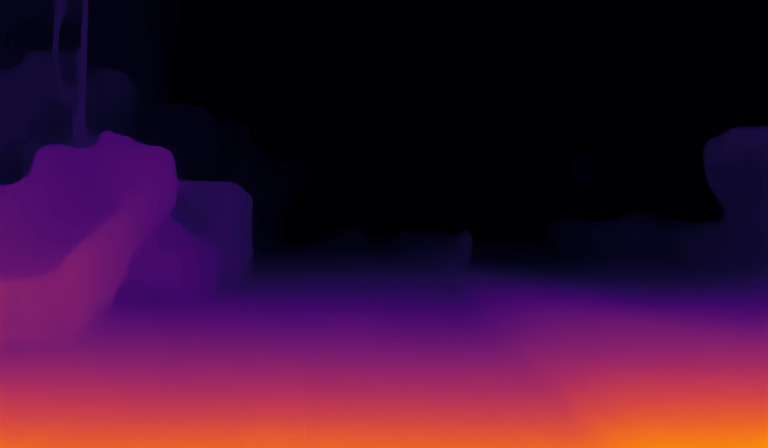}\end{tabular} &
\begin{tabular}{l}\includegraphics[width=0.15\linewidth]{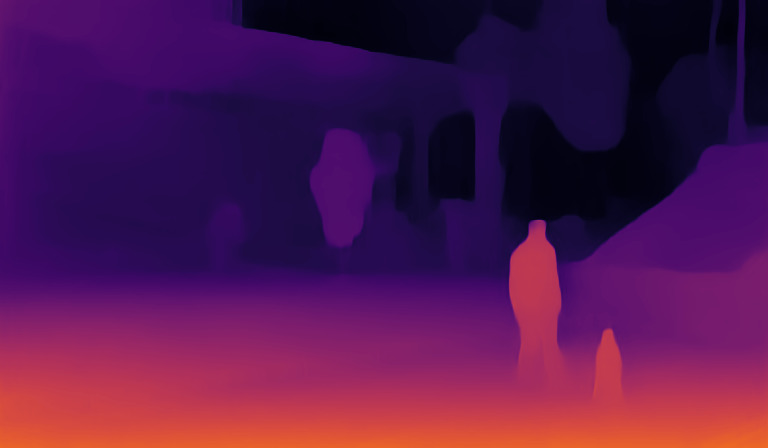}\end{tabular} &
\begin{tabular}{l}\includegraphics[width=0.15\linewidth]{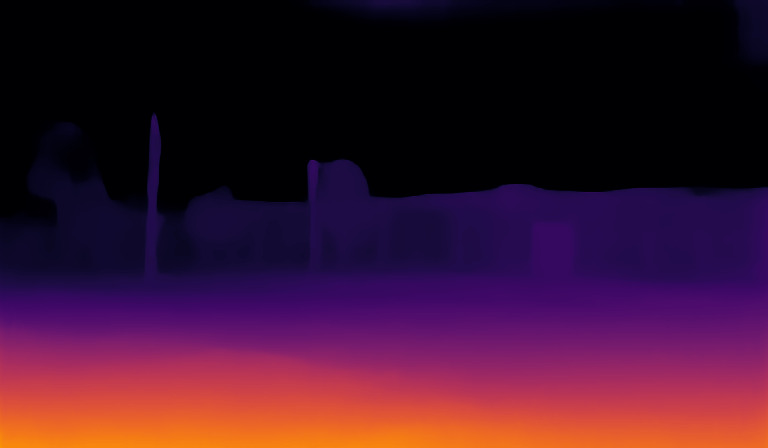}\end{tabular} &
\begin{tabular}{l}\includegraphics[width=0.15\linewidth]{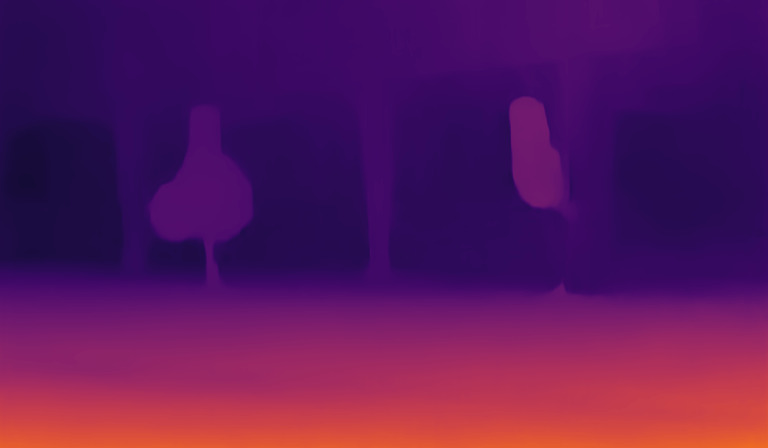}\end{tabular} &
\begin{tabular}{l}\includegraphics[width=0.15\linewidth]{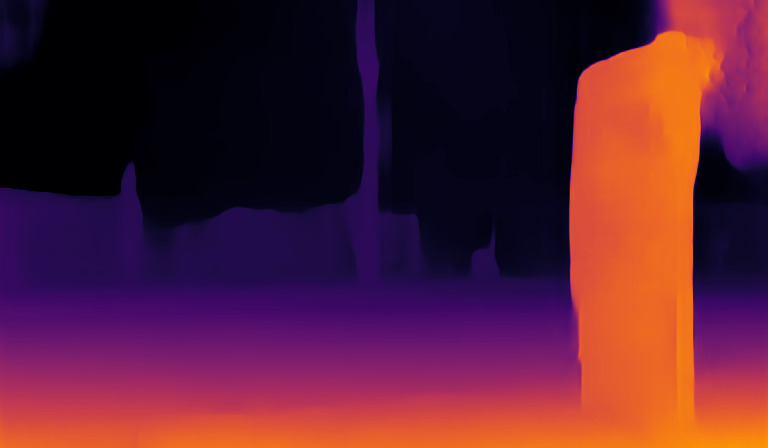}\end{tabular} &
\begin{tabular}{l}\includegraphics[width=0.15\linewidth]{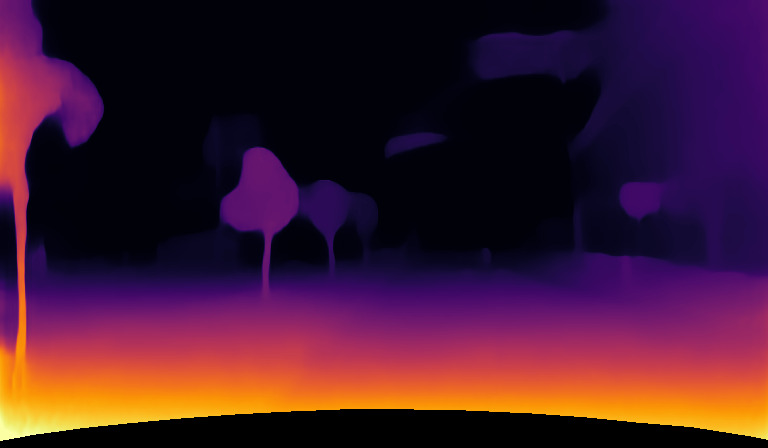}\end{tabular} \\ \midrule

\rotatebox[origin=c]{90}{Input} &
\begin{tabular}{l}\includegraphics[width=0.15\linewidth]{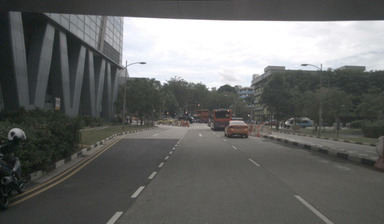}\end{tabular} &
\begin{tabular}{l}\includegraphics[width=0.15\linewidth]{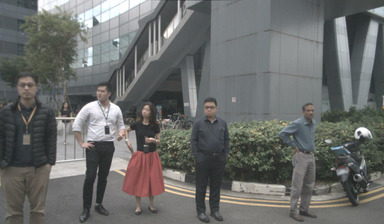}\end{tabular} &
\begin{tabular}{l}\includegraphics[width=0.15\linewidth]{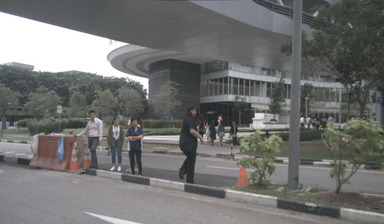}\end{tabular} &
\begin{tabular}{l}\includegraphics[width=0.15\linewidth]{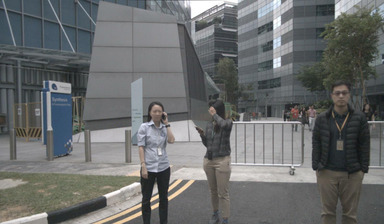}\end{tabular} &
\begin{tabular}{l}\includegraphics[width=0.15\linewidth]{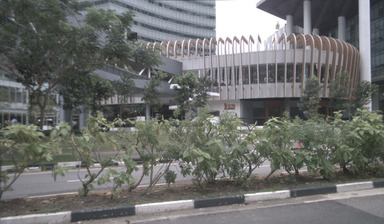}\end{tabular} &
\begin{tabular}{l}\includegraphics[width=0.15\linewidth]{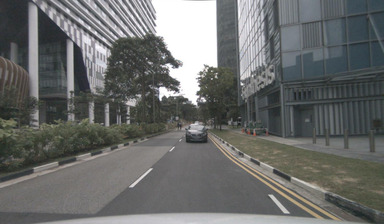}\end{tabular} \\

\rotatebox[origin=c]{90}{SD~\cite{SurroundDepth}} &
\begin{tabular}{l}\includegraphics[width=0.15\linewidth]{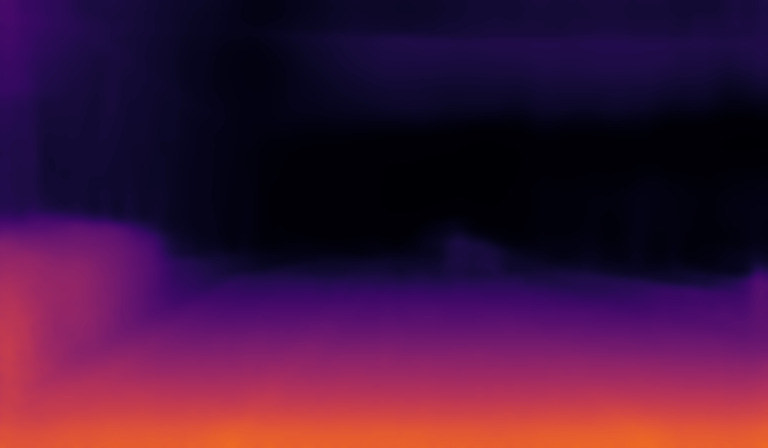}\end{tabular} &
\begin{tabular}{l}\includegraphics[width=0.15\linewidth]{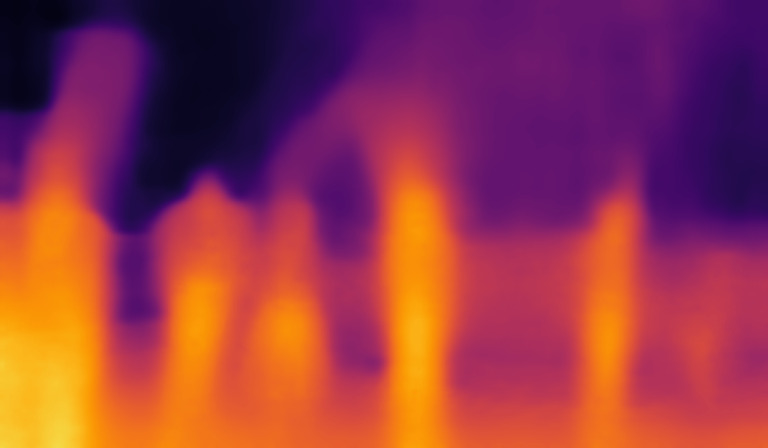}\end{tabular} &
\begin{tabular}{l}\includegraphics[width=0.15\linewidth]{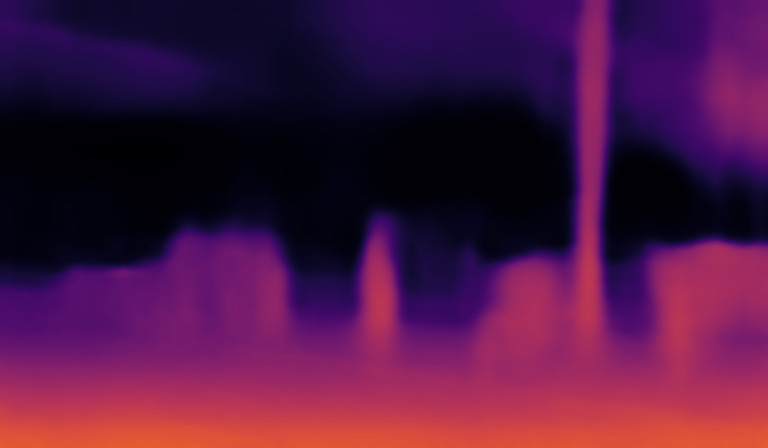}\end{tabular} &
\begin{tabular}{l}\includegraphics[width=0.15\linewidth]{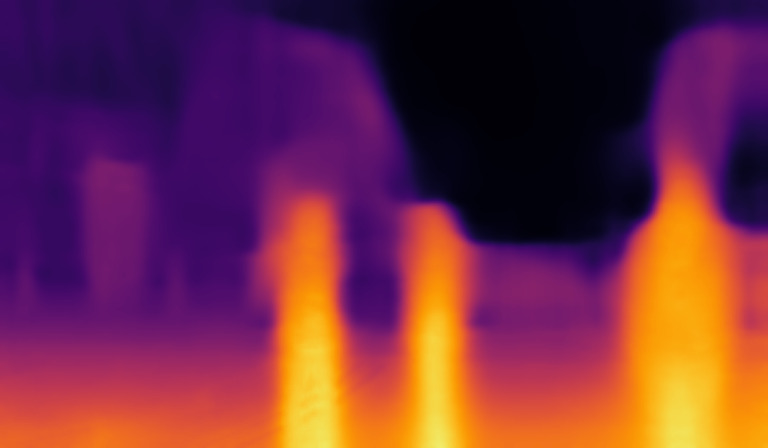}\end{tabular} &
\begin{tabular}{l}\includegraphics[width=0.15\linewidth]{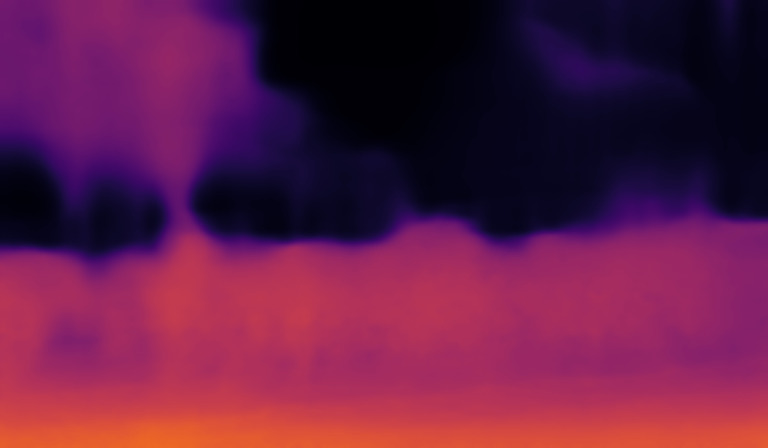}\end{tabular} &
\begin{tabular}{l}\includegraphics[width=0.15\linewidth]{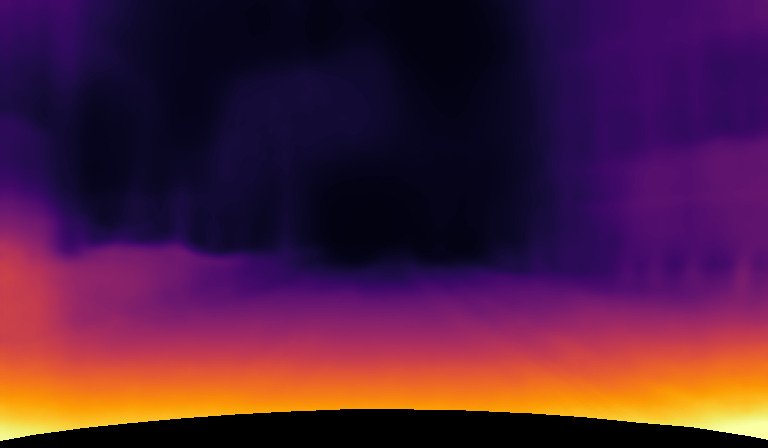}\end{tabular} \\

\rotatebox[origin=c]{90}{Ours} &
\begin{tabular}{l}\includegraphics[width=0.15\linewidth]{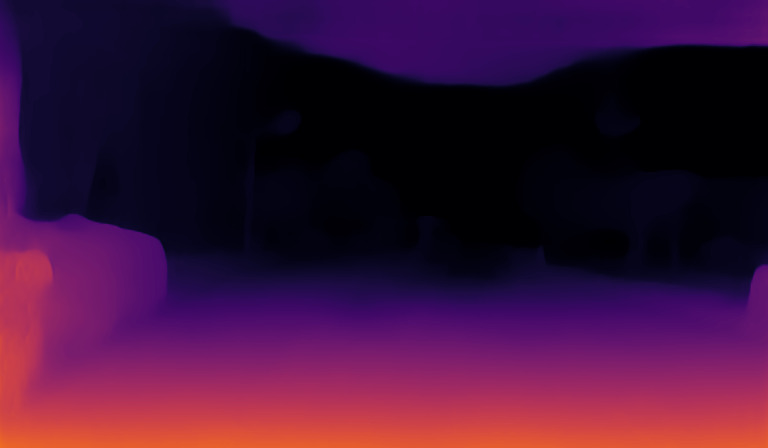}\end{tabular} &
\begin{tabular}{l}\includegraphics[width=0.15\linewidth]{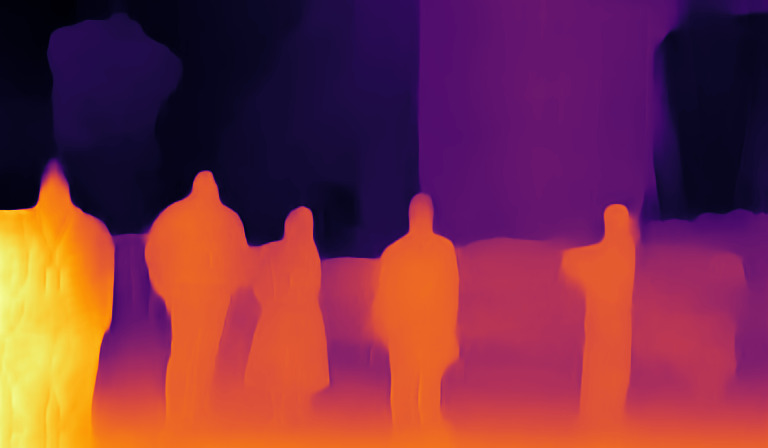}\end{tabular} &
\begin{tabular}{l}\includegraphics[width=0.15\linewidth]{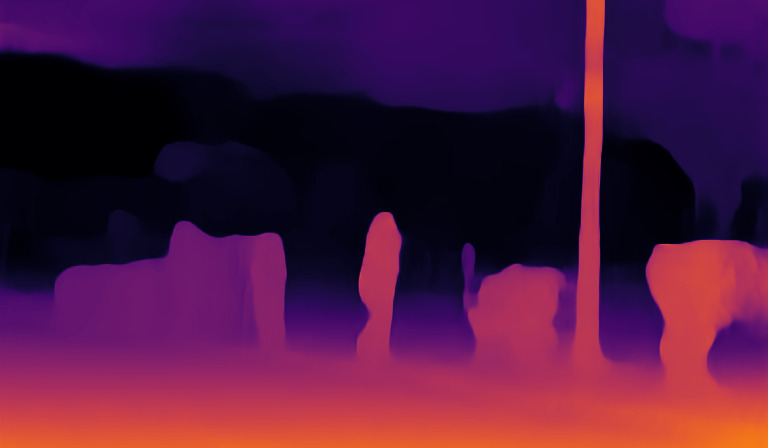}\end{tabular} &
\begin{tabular}{l}\includegraphics[width=0.15\linewidth]{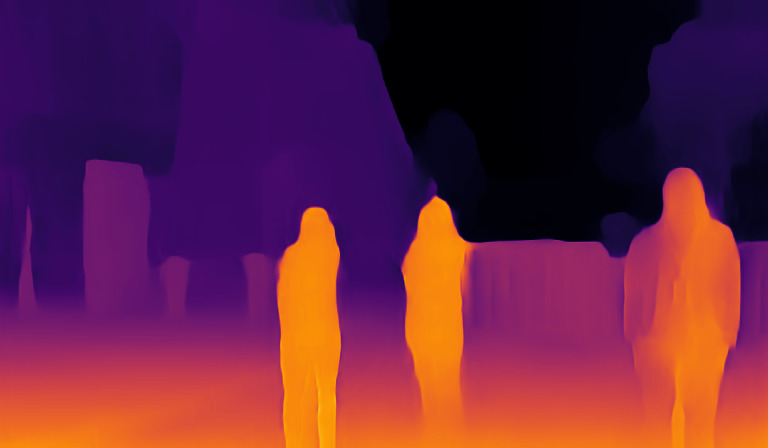}\end{tabular} &
\begin{tabular}{l}\includegraphics[width=0.15\linewidth]{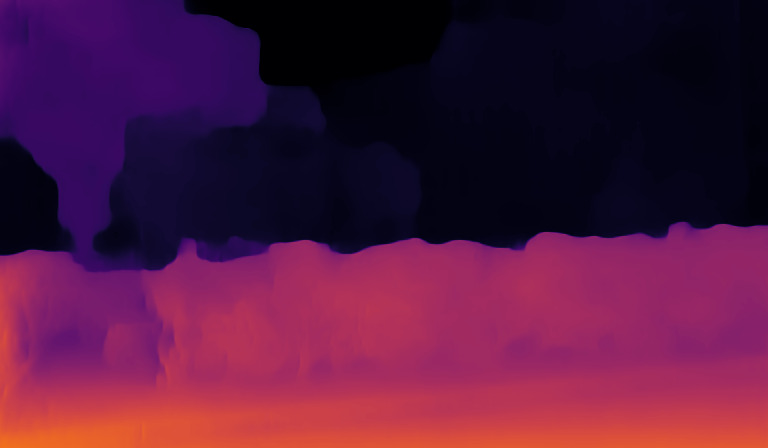}\end{tabular} &
\begin{tabular}{l}\includegraphics[width=0.15\linewidth]{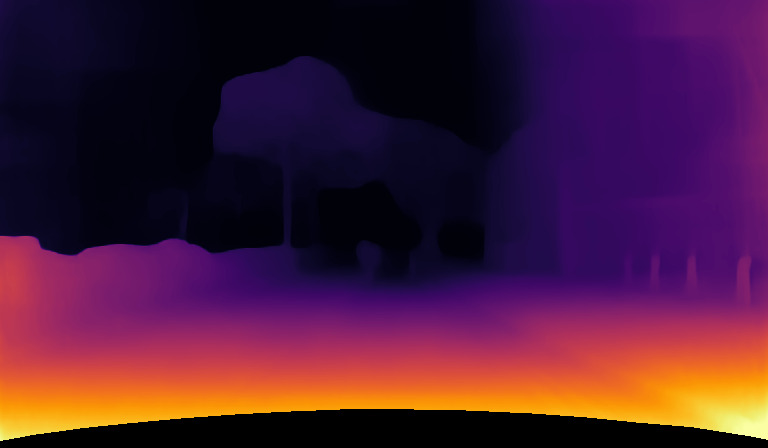}\end{tabular} \\

\end{tabular}

\caption{\textbf{Qualitative comparison on NuScenes}. We show a comparison of depth maps from our method to the depth maps of the state-of-the-art approach SurroundDepth~\cite{SurroundDepth}. We observe that our approach produces significantly sharper and more accurate depth predictions.}
\label{fig:nuscenes_examples}
\end{figure*}

\begin{figure*}[h]
\centering
\setlength\tabcolsep{0.5 pt}

\begin{tabular}{lcc}
& DDAD 000194 & DDAD 000188 \\
\rotatebox[origin=c]{90}{Ground-Truth} &
\begin{tabular}{l}\adjincludegraphics[width=.45\linewidth,trim={0 {.15\height} 0 {.25\height}},clip]{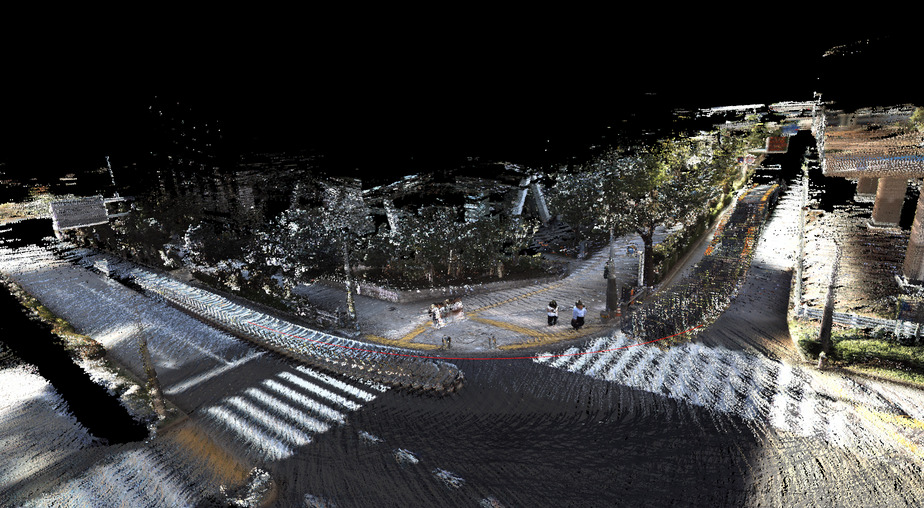}\end{tabular} &
\begin{tabular}{l}\adjincludegraphics[width=.45\linewidth,trim={0 {.15\height} 0 {.25\height}},clip]{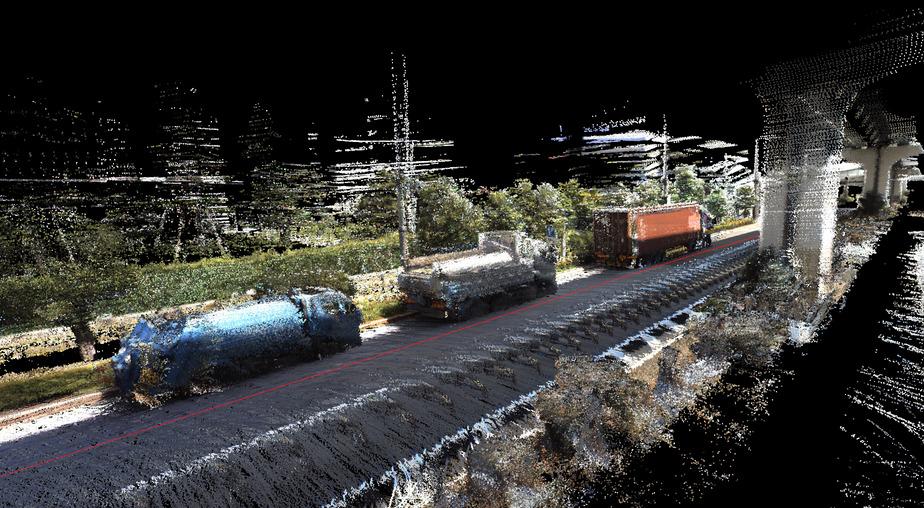}\end{tabular} \\
\rotatebox[origin=c]{90}{FSM \cite{FSM}} &
\begin{tabular}{l}\adjincludegraphics[width=.45\linewidth,trim={0 {.15\height} 0 {.25\height}},clip]{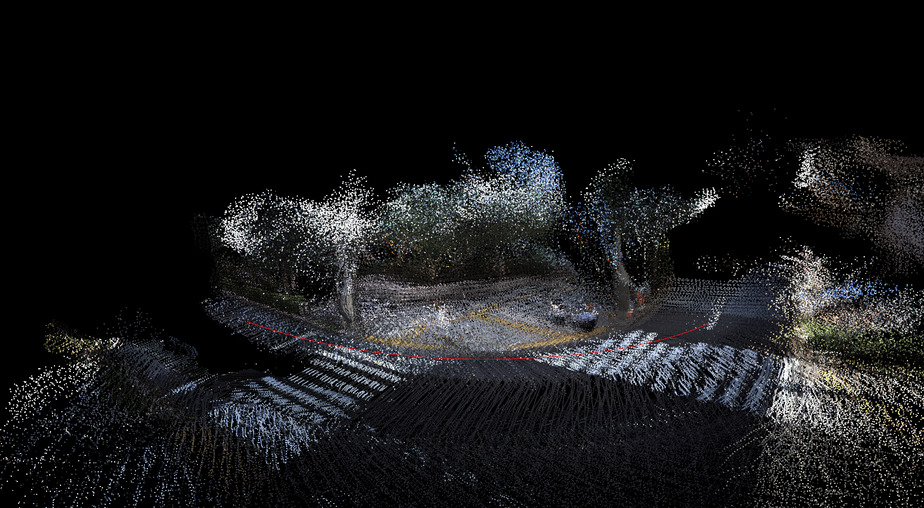}\end{tabular} &
\begin{tabular}{l}\adjincludegraphics[width=.45\linewidth,trim={0 {.15\height} 0 {.25\height}},clip]{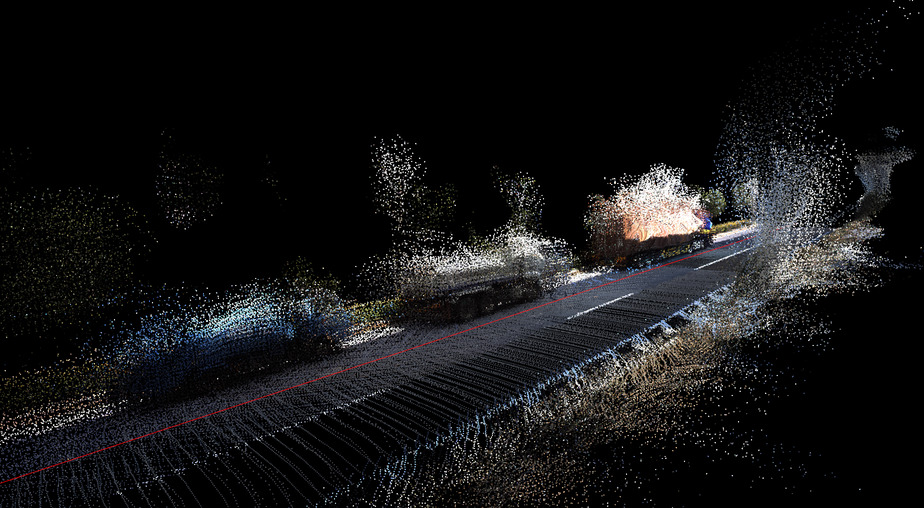}\end{tabular} \\
\rotatebox[origin=c]{90}{SD \cite{SurroundDepth}} &
\begin{tabular}{l}\adjincludegraphics[width=.45\linewidth,trim={0 {.15\height} 0 {.25\height}},clip]{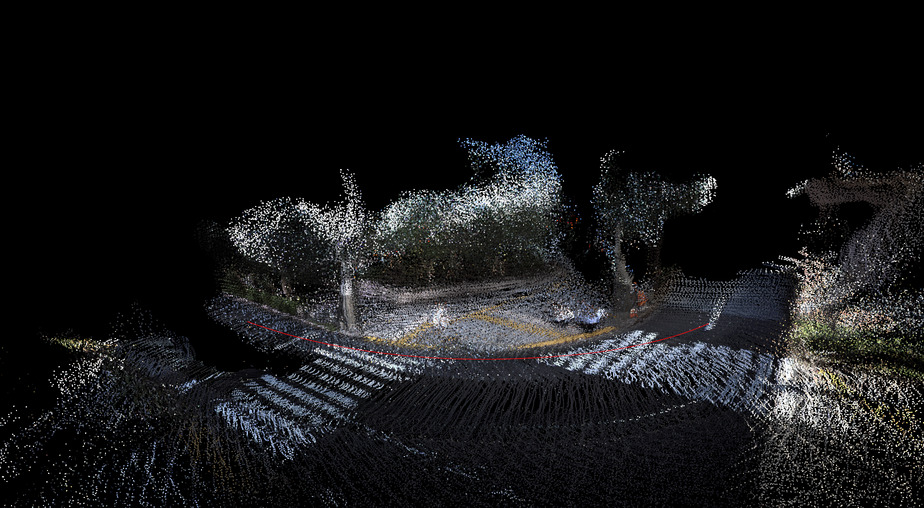}\end{tabular} &
\begin{tabular}{l}\adjincludegraphics[width=.45\linewidth,trim={0 {.15\height} 0 {.25\height}},clip]{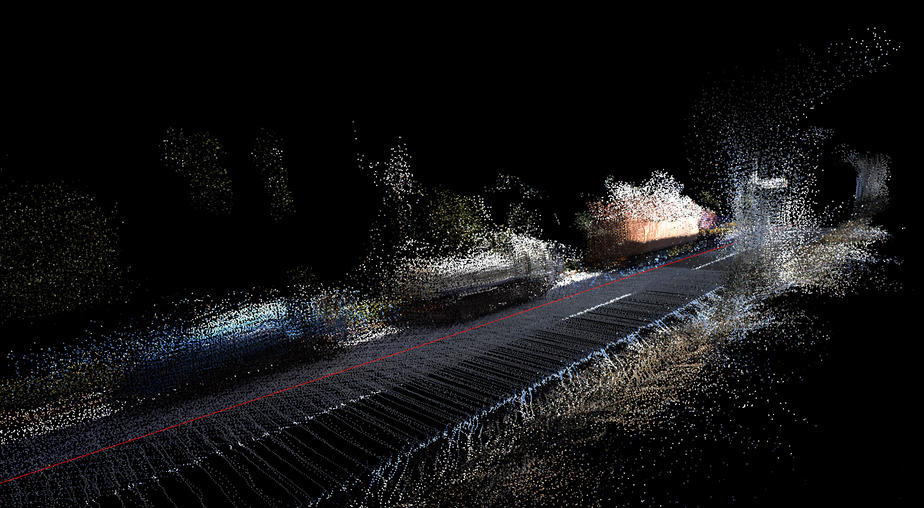}\end{tabular} \\
\rotatebox[origin=c]{90}{Ours} &
\begin{tabular}{l}\adjincludegraphics[width=.45\linewidth,trim={0 {.15\height} 0 {.25\height}},clip]{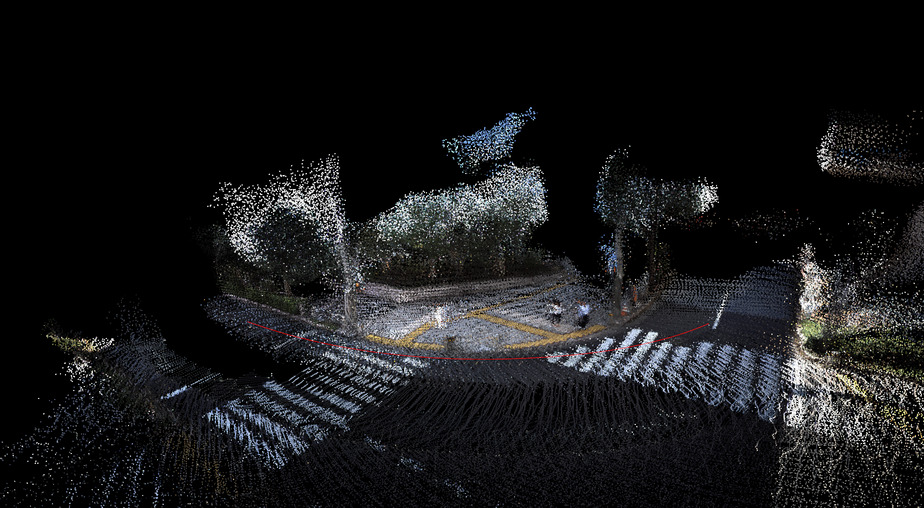}\end{tabular} &
\begin{tabular}{l}\adjincludegraphics[width=.45\linewidth,trim={0 {.15\height} 0 {.25\height}},clip]{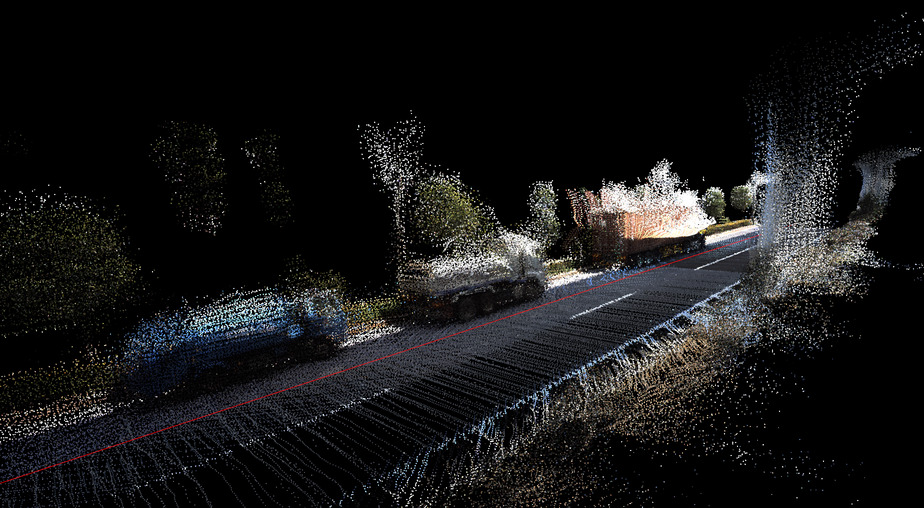}\end{tabular} \\

\end{tabular}
\caption{\textbf{Qualitative comparison of 3D reconstructions on DDAD.} We show the 3D reconstructions of our method compared to SurroundDepth~\cite{SurroundDepth} and FSM~\cite{FSM}. Additionally, we plot the ground-truh LiDAR 3D reconstruction at the top. The ego-vehicle trajectory is marked in red. We observe that our method produces significantly more consistent and accurate 3D reconstructions than competing methods, as can be seen when focusing on the trucks (right) and the street markings and pedestrians (left).}
\label{fig:ddad_pointclouds}
\end{figure*}

\begin{figure*}[h]
\centering
\setlength\tabcolsep{0.5 pt}

\begin{tabular}{lcc}
& NuScenes 0016 & NuScenes 0268 \\

\rotatebox[origin=c]{90}{Ground-Truth} &
\begin{tabular}{l}\adjincludegraphics[width=.45\linewidth,trim={0 {.15\height} 0 {.25\height}},clip]{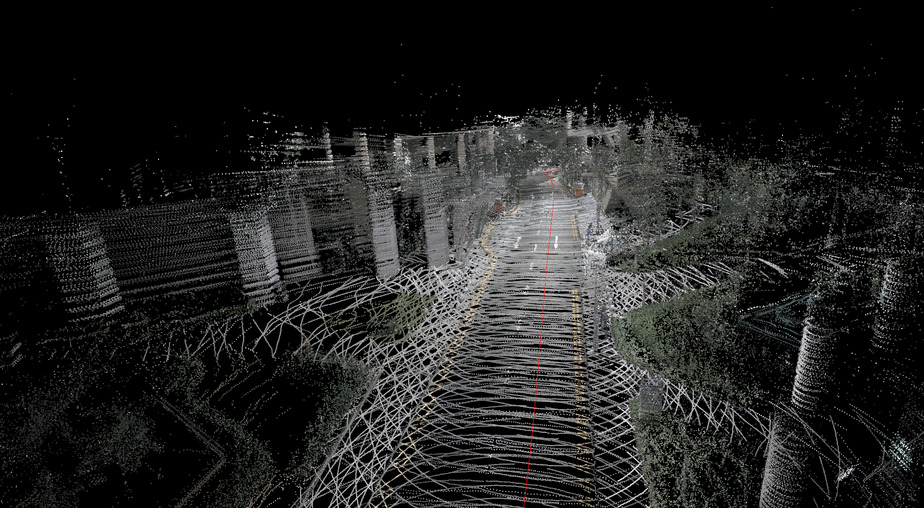}\end{tabular} &
\begin{tabular}{l}\adjincludegraphics[width=.45\linewidth,trim={0 {.15\height} 0 {.25\height}},clip]{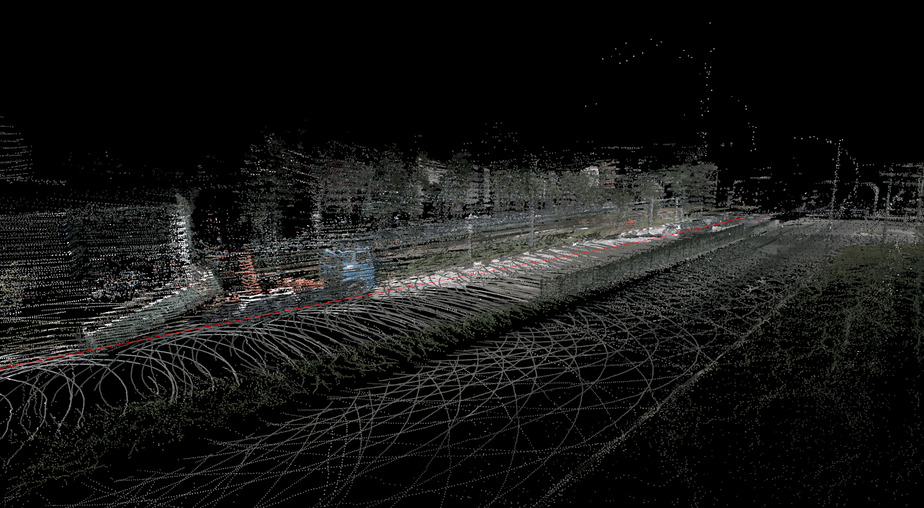}\end{tabular} \\

\rotatebox[origin=c]{90}{Ours} &
\begin{tabular}{l}\adjincludegraphics[width=.45\linewidth,trim={0 {.15\height} 0 {.25\height}},clip]{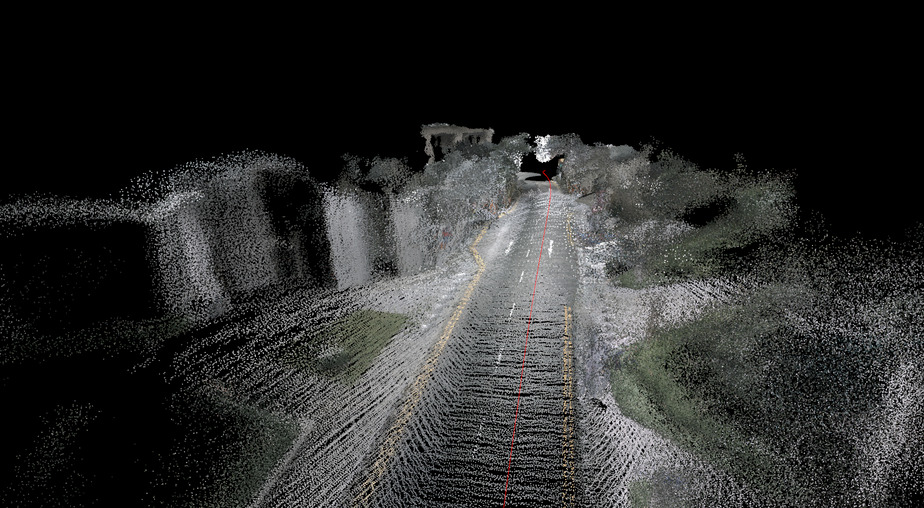}\end{tabular} &
\begin{tabular}{l}\adjincludegraphics[width=.45\linewidth,trim={0 {.15\height} 0 {.25\height}},clip]{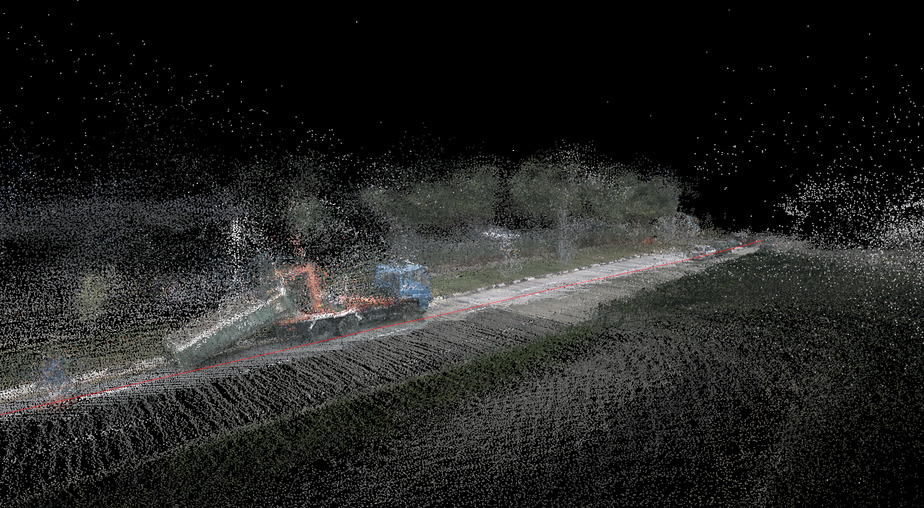}\end{tabular} \\

\end{tabular}
\caption{\textbf{Qualitative 3D reconstruction results on NuScenes.} We show our 3D reconstruction results alongside the LiDAR ground-truth 3D reconstruction. The ego-vehicle trajectory is marked in red. We observe that our method yields similarly consistent 3D reconstruction results as the LiDAR ground-truth while being denser.}
\label{fig:nuscenes_pointclouds}
\end{figure*}

\end{document}

\typeout{get arXiv to do 4 passes: Label(s) may have changed. Rerun}